%% file: main.tex
\documentclass[table]{gtech}
\PassOptionsToPackage{table, usenames, dvipsnames}{xcolor}
\usepackage{xcolor}
\usepackage{amssymb}
\usepackage{multirow}
\usepackage{bigdelim}
\usepackage{longtable}
\usepackage{tabularray}
\usepackage{wrapfig}
\usepackage[most]{tcolorbox}
\usepackage{url}
\usepackage{float}
\usepackage{datatool}
\usepackage{enumitem}
\usepackage{subcaption} %
\usepackage[justification=centering]{caption}

\RequirePackage{tgpagella} %
\RequirePackage{mathpazo}  %
\RequirePackage{inconsolata} %
\usepackage{makecell}

\usepackage{booktabs} %
\usepackage{array}    %
\newcolumntype{C}[1]{>{\centering\arraybackslash}p{#1}}

\usepackage[utf8]{inputenc} %
\usepackage[T1]{fontenc}    %
\usepackage{hyperref}       %
\usepackage{url}            %
\usepackage{booktabs}       %
\usepackage{amsfonts}       %
\usepackage{nicefrac}       %
\usepackage{microtype}      %
\usepackage{xcolor}         %
\usepackage{xspace}
\usepackage{amsthm}   %
\usepackage{amsmath}  %
\usepackage{amssymb}  %
\usepackage{bm}       %

\theoremstyle{definition}
\newtheorem{definition}{Definition}[section]

\renewcommand{\title}[1]{\newcommand{\titlelist}{{\huge\fontfamily{optimistic}\selectfont #1}}}

\newcommand{\ie}{\emph{i.e.,}\xspace}

\newcommand{\eg}{\emph{e.g.,}\xspace}

\newcommand{\ignore}[1]{}

\newcommand{\mini}{\texttt{Ling-mini-beta}\xspace}
\newcommand{\dense}{\texttt{Dense-6.1B}\xspace}

\usepackage{arydshln}
\definecolor{CQColor}{rgb}{0.0,0.0,1.0} %

\usepackage{colortbl}
\usepackage{amssymb}
\usepackage{pifont}
\usepackage{booktabs,multirow}
\usepackage{makecell}
\usepackage{tabulary}
\usepackage{fontawesome5}
\usepackage{bbding}
\usepackage{multicol}

\newlength\savewidth

\title{Towards Greater Leverage: Scaling Laws \\for Eff{}icient Mixture-of-Experts Language Models}

\author[]{Changxin Tian, Kunlong Chen, Jia Liu, Ziqi Liu}
\author[*]{Zhiqiang Zhang}
\author[]{Jun Zhou}

\affiliation[]{Ling Team, Ant Group\\[0.5em]}
\contribution[*]{Corresponding author}

\abstract{
Mixture-of-Experts (MoE) has become a dominant architecture for scaling Large Language Models (LLMs) efficiently by decoupling total parameters from computational cost. However, this decoupling creates a critical challenge: predicting the model capacity of a given MoE configurations (\eg expert activation ratio and granularity) remains an unresolved problem.
To address this gap, we introduce \textit{Efficiency Leverage (EL)}, a metric quantifying the computational advantage of an MoE model over a dense equivalent. We conduct a large-scale empirical study, training over 300 models up to 28B parameters, to systematically investigate the relationship between MoE architectural configurations and EL. 
Our findings reveal that EL is primarily driven by the expert activation ratio and the total compute budget, both following predictable power laws, while expert granularity acts as a non-linear modulator with a clear optimal range. We integrate these discoveries into a unified scaling law that accurately predicts the EL of an MoE architecture based on its configuration. 
To validate our derived scaling laws, we designed and trained \texttt{Ling-mini-beta}, a pilot model for Ling-2.0 series with only 0.85B active parameters, alongside a 6.1B dense model for comparison. 
When trained on an identical 1T high-quality token dataset, \texttt{Ling-mini-beta} matched the performance of the 6.1B dense model while consuming over 7x fewer computational resources, thereby confirming the accuracy of our scaling laws.
This work provides a principled and empirically-grounded foundation for the scaling of efficient MoE models. 
}
\date{July 20, 2025}
\gtechdata[Correspondence]{\email{tianchangxin.tcx@antgroup.com},  
\email{lingyao.zzq@antgroup.com}}

\begin{document}
\maketitle

\input{sections/1-intro}

\input{sections/2-pre}

\input{sections/3-lever}

\input{sections/4-scaling}

\input{sections/5-validation}
\input{sections/6-discussion}

\input{sections/7-related}

\input{sections/8-conclusion}

\bibliographystyle{assets/plainnat}
\bibliography{main}

\input{sections/9-appendix}

\end{document}

%% file: sections/1-intro.tex
\begin{figure}[b]
    \centering
    \hfill 
    \begin{subfigure}[b]{0.4\textwidth} 
        \includegraphics[width=\textwidth]{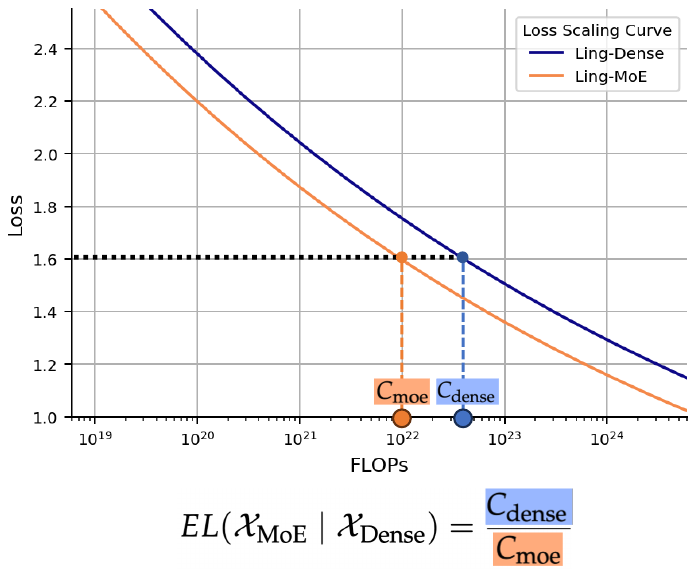}
        \caption{Definition of Efficiency Leverage (EL)}
        \label{fig:lever-1e20}
    \end{subfigure}
    \hfill 
    \begin{subfigure}[b]{0.49\textwidth} 
        \includegraphics[width=\textwidth]{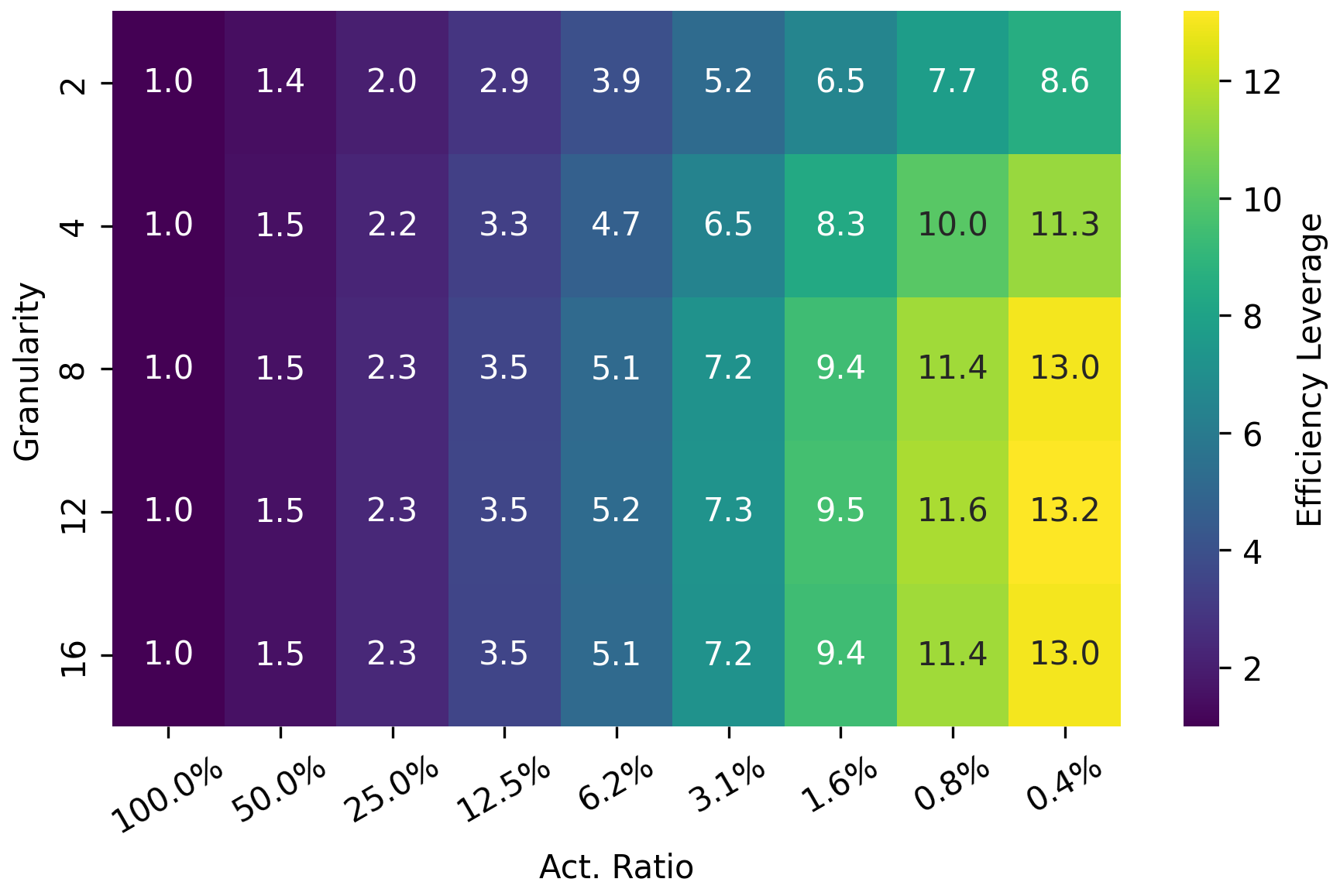}
        \caption{Estimated EL at $1e22$ FLOPs}
        \label{fig:lever-1e22}
    \end{subfigure}
    \hfill 
    \caption{\textbf{Illustration of definition of Efficiency Leverage (EL) and its estimated values using Eq.~\eqref{eq:all} for $1e22$ FLOPs.} 
    }
    \label{fig:hotmap-lever}
\end{figure}

\section{Introduction}
Recently, Mixture-of-Experts (MoE) models~\citep{shazeer2017outrageously,jiang2024mixtral,deepseekai2024deepseekv3technicalreport} have emerged as a leading paradigm for constructing large language models (LLMs)~\citep{zhao2023survey}, primarily due to its remarkable computational efficiency~\citep{clark2022unified}. By leveraging sparse activation, MoE models can dramatically increase their total parameter count without proportionally increasing the computational cost (FLOPs). For instance, DeepSeekMoE~\citep{dai2024deepseekmoe}, with 16 billion total parameters, activates only 2.8 billion per token, yet achieves performance comparable to a 7-billion-parameter dense model, showcasing a parameter efficiency gain of approximately 2.5x.
However, the decoupling of computational cost from the total parameter count in MoE architectures introduces a new challenge in assessing a model's capacity. While capacity in dense models is traditionally correlated with the parameter count, for MoE models, neither the total nor the activated parameter count alone serves as a reliable proxy for performance.
Consequently, predicting the effective capacity of a specific MoE architecture and setting realistic performance expectations before pre-training remains a critical and unresolved problem.

To understand the relationship between MoE architecture and performance, recent studies~\citep{clark2022unified, ludziejewski2024scaling, abnar2025parameters, ludziejewski2025joint} have begun to investigate their scaling laws. As a cornerstone to LLM research, scaling laws~\citep{kaplan2020scaling, hestness2017deep} reveal a core principle: model performance improves predictably with increases in compute, model size, and data scale. This fundamental property allows us to infer the performance of vastly larger models by training a series of smaller ones, offering crucial insights into architectural scalability. However, existing research on MoE scaling laws has predominantly focused on isolated dimensions, such as parameter sparsity~\citep{clark2022unified, abnar2025parameters} or expert granularity~\citep{ ludziejewski2024scaling}. Given that MoE performance is governed by the complex interplay of multiple interdependent factors, we are still unable to intuitively determine the ``equivalent capacity'' of a given MoE architecture. 

To address this challenge, we introduce \textbf{Efficiency Leverage (EL)}, a metric designed to quantify the computational efficiency of a MoE architecture relative to a dense counterpart. 
Specifically, we define the EL of a target MoE architecture $\mathcal{X}_{\text{MoE}}$ with respect to a dense baseline $\mathcal{X}_{\text{Dense}}$ as the ratio of their computational costs $C$ required to achieve the same performance level (\eg identical loss): 
\begin{equation}
\begin{aligned}
EL(\mathcal{X}_{\text{MoE}} \mid \mathcal{X}_{\text{Dense}} ) = \frac{C_{\text{dense}}}{C_{\text{moe}}},
\end{aligned}
\end{equation}
This definition provides a standardized benchmark for architectural comparison: a higher EL value signifies greater efficiency. 
For instance, an EL of 2 indicates the MoE model requires only half the computational cost to reach the same performance as the dense baseline. 
Consequently, for a fixed compute budget, an MoE architecture with higher EL enables larger effective parameter scaling or more comprehensive training, thereby improving efficiency.

To investigate EL and its relationship with MoE architectures, our study employs a three-stage methodology. 
First, we establish scaling laws for optimal hyper-parameters and model-data allocation in dense/MoE models, ensuring all experimental models are evaluated in their ``well-trained'' conditions. 
Second, we deconstruct the MoE architecture into its core design dimensions, including the expert activation ratio, expert granularity, the shared expert ratio, and others configurations. We then systematically analyze the impact of each dimension on EL. 
Finally, we integrate our empirical findings into a unified scaling law to models the relationship between MoE configurations and the resulting EL, providing a predictive framework for designing efficient MoE models.

Our large-scale empirical study, encompassing over 300 trained models with up to 28B parameters, reveals several core principles governing the efficiency of MoE architectures. We distill our findings into the following key insights: 
\begin{enumerate}
    \item \textbf{Activation ratio as the primary driver of efficiency.} 
    The expert activation ratio emerges as the primary determinant of EL. We observe a stable power-law relationship: EL increases as the activation ratio decreases (\ie as sparsity increases). This reveals that sparsely activated pathways yield consistent and predictable gains in computational efficiency. 
    \item \textbf{Expert granularity as a non-linear modulator.} 
    Superimposed on this primary trend, expert granularity introduces a log-polynomial adjustment to EL. This effect is independent of the total compute budget and implies an optimal range for expert size. Our experiments, which utilize a standard load-balancing loss, identify this optimum to be between 8 and 12. 
    \item \textbf{Amplifying effect of the compute budget.} 
    Crucially, the EL of a given MoE architecture is not static; it scales with the training compute budget, also following a power law. This finding underscores the advantage of MoE models in large-scale pre-training scenario, where their efficiency gains become increasingly significant as computational resources expand. 
    \item \textbf{Secondary impact of other architectural factors.} 
    Other design choices, such as the use of shared experts or the specific arrangement of MoE and dense layers, exert only a secondary influence on EL. These factors typically have broadly applicable, near-optimal settings that require minimal tuning.
\end{enumerate}
Building upon these observations, we derive a unified scaling law for efficiency leverage of MoE. This formula integrates the combined effects of the compute budget, activation ratio, and expert granularity. It enables us to directly predict an MoE architecture's EL for a given configuration, providing principled guidance for efficient MoE architectural design. 
As a practical demonstration, Figure~\ref{fig:lever-1e22} illustrates the estimated EL under compute budgets of $1e22$ FLOPs.

According to our derived scaling law for EL, we predict that an MoE model with a 3.1\% activation ratio and an expert granularity of 12 will achieve over 7x computational efficiency under a $1e22$ FLOPs compute budget. 
To empirically validate this, we designed and trained ``Ling-mini-beta,'' a pilot model for the Ling-2.0 series, with 0.85B activated parameters and 17.5B total parameters. The model was trained on a 1T-token high-quality dataset and benchmarked against its dense counterpart, a 6.1B parameter model.
Experimental results show that when trained on the same 1T-token dataset, Ling-mini-beta achieves a lower final training loss and exhibits a slight performance advantage across a suite of downstream tasks.
This outcome confirms our theoretical prediction, validating that this MoE architecture yields an efficiency gain of over 7x. These findings provide a solid theoretical and empirical foundation for the design of future large-scale, efficient MoE models.

%% file: sections/2-pre.tex
\section{Preliminary}
\label{sec:Preliminary}

\subsection{Mixture-of-Expert Transformers.}
\label{sec:pre-arch}

MoE architecture modifies the standard Transformer by replacing each Feed-Forward Network (FFN) block with an MoE layer. This layer consists of multiple expert networks (or simply "experts") and a gating mechanism. For each input token, the gating mechanism dynamically routes it to a small subset of these experts. This selective activation of experts for each token significantly reduces the computational cost per forward pass compared to a dense model of equivalent parameter count. 

\paragraph{Total and Active Parameters.} 
In MoE models, we distinguish between two parameter counts. The \textit{total parameters} ($N$) encompass all weights in the model, including those of every expert. In contrast, the \textit{active parameters} ($N_a$) for a given input consist only of the non-expert components and the specific experts selected by the top-$k$ gating mechanism.

\paragraph{Routable and Shared Experts.} 
An MoE layer typically contains two types of experts. First, there are $E$ \textit{routable experts}, from which the gating network selects a subset of $E_a$ (the number of activated experts) for each token. Additionally, many modern MoE architectures incorporate $E_s$ \textit{shared experts}, which are activated for every token to process and consolidate knowledge common to all inputs.

\paragraph{Activation Ratio and Sharing Ratio.}
We introduce two metrics to characterize the expert configuration. \textit{Activation ratio}, $A$, is the ratio of activated experts to the total number of experts. \textit{Sharing ratio}, $S$, is the ratio of shared experts to activated experts. Assuming all experts have identical dimensions, these rates are defined as $A = (E_a + E_s)/ (E + E_s)$ and $S = E_s / (E_a+E_s)$. These metrics quantify the sparsity within the MoE layer, offering an intuitive measure of expert utilization.

\paragraph{Granularity of Experts.}
In conventional MoE architectures, the intermediate dimension of each expert, $d_\text{expert}$, is typically equals the feed-forward network (FFN) dimension, which is conventionally set to $4d_{\text{model}}$.
However, recent works~\citep{deepseekai2024deepseekv3technicalreport} have diverged from this practice by decoupling the expert dimension from the model's hidden size and the FFN's intermediate dimension. To systematically analyze this design choice, we define \textit{expert granularity} as $G = 2d_\text{model} / d_\text{expert}$. A higher value of $G$ corresponds to having a larger number of smaller experts for a fixed total parameter count within the MoE layers. 
It is important to note that, to align with recent leading MoE models~\citep{deepseekai2024deepseekv3technicalreport,kimiK2}, we adopted a different definition of ``granularity'' from that of \cite{ludziejewski2024scaling}. They define granularity as $4d_\text{model} / d_\text{expert}$, whereas our definition results in each expert being half the size for the same granularity value, which consequently leads to different observed phenomena. 

\paragraph{Defining Model Scale via Computation.} 
We quantify the computational cost using Floating Point Operations (FLOPs). Consistent with prior work~\citep{bi2024deepseek}, we define a model's scale in terms of computation, denoted as $M$, representing the number of non-embedding FLOPs per token in a single forward pass. For MoE models, this is particularly important as $M$ accounts only for the sparsely activated components (\ie the selected experts). We exclude the embedding layer from this calculation because its contribution to both overall computation and model capacity is minimal. To ensure our analysis is grounded in accurate figures, we employ an exact calculation for $M$, avoiding error accumulation found in common approximations (details in Appendix~\ref{app:flops}). The total training compute $C$ is thus a function of $M$ and the number of training tokens $D$:
\begin{equation}
\label{eq:M}
C = M \cdot D
\end{equation}
This formulation provides a consistent basis for comparing dense and MoE architectures.

\subsection{Scaling Laws for MoE Optimal Hyper-parameters}
\label{sec:scaling-hyper}
The performance of a MoE model is sensitive to its hyperparameters. To ensure that our subsequent architectural comparisons are reliable, it is crucial to evaluate each configuration under its optimal hyperparameter settings. Therefore, we first conduct a preliminary study to establish the scaling laws for optimal MoE hyperparameters.
Previous research \citep{bi2024deepseek} has established that the optimal hyperparameters are primarily a function of the total computational budget. Accordingly, we performed a hyperparameter search across a compute range of $3e17$ to $3 e20$ FLOPs, using a Warmup-Stable-Decay (WSD) learning rate schedule~\citep{hu2024minicpm}. We trained multiple models, varying both learning rate and batch size, which were sampled from a log-base-2 grid. Specifically, the exponents for the learning rate ranged from -11 to -9.0, and for the batch size, from 18 to 21.
To make this analysis tractable, we initially fixed the MoE configuration to one with 64 experts, of which 4 are activated per token, plus an additional shared expert (resulting in an activation ratio $A=7.8\%$ and a granularity $G=2$). Detailed settings of the experimental models are available in the Appendix~\ref{app:setup}. We then verified that the conclusions from this configuration generalize across different activation ratios. 

\begin{figure}[htbp]
    \centering
    \begin{subfigure}[b]{0.46\textwidth}
        \includegraphics[width=\textwidth]{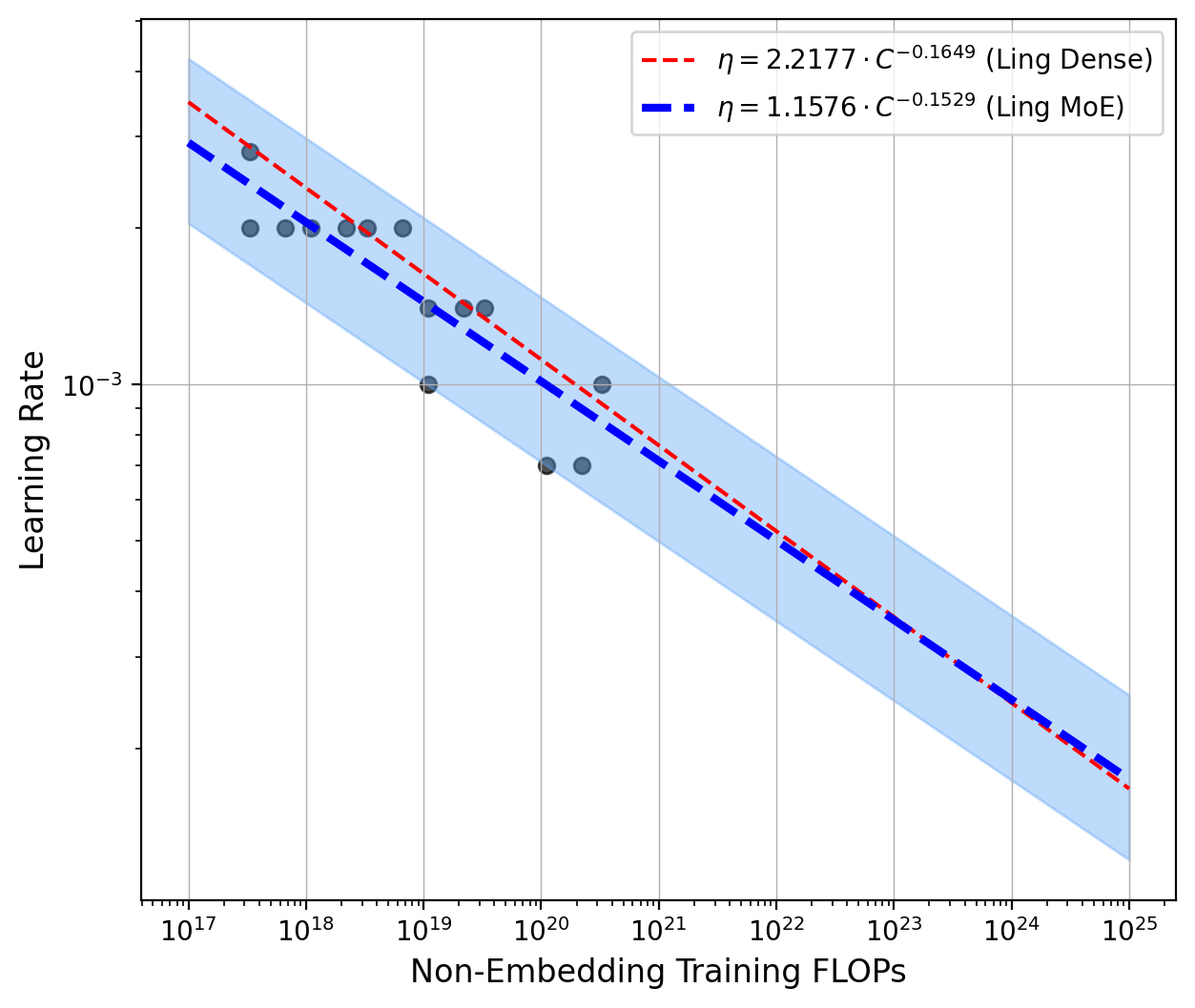}
        \caption{Learning rate scaling curve}
    \end{subfigure}
    \hfill 
    \begin{subfigure}[b]{0.45\textwidth} 
        \includegraphics[width=\textwidth]{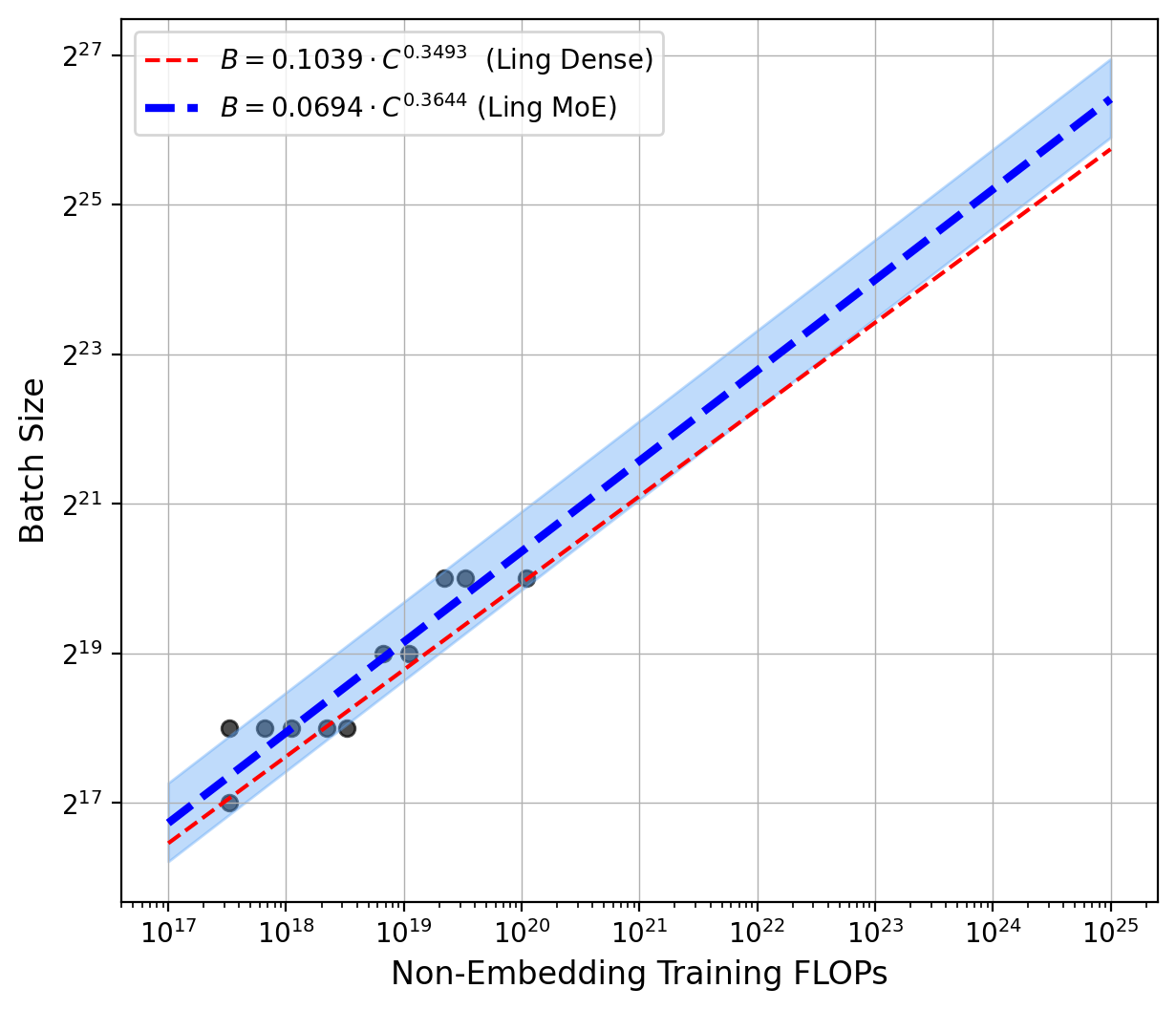}
        \caption{Batch size scaling curve}
    \end{subfigure}
    \caption{\textbf{Scaling laws for optimal hyperparameters.} Blue and red lines represent the fitted laws for MoE and dense models, respectively, derived on the same training dataset. Gray circles are the experimental data points used for fitting.}
    \label{fig:hyper-scaling}
\end{figure}

Figure~\ref{fig:hyper-scaling} illustrates the fitting process. To ensure robustness, we identify ``near-optimal'' configurations as those achieving a loss within 0.25\% of the minimum for a given compute budget. After removing outliers, we fitted the optimal batch size, $B^{\text{opt}}$, and learning rate, $\eta^{\text{opt}}$, against the compute budget $C$. The resulting scaling laws reveal clear trends: $B^{\text{opt}}$ increases and $\eta^{\text{opt}}$ decreases with larger $C$. The final formulas obtained from the fitting process are as follows:
\begin{equation}
\label{eq:hyper-scaling}
\begin{aligned}
    \eta^{\text{opt}} &= 1.1576 \cdot C^{-0.1529} \\
    B^{\text{opt}} &= 0.0694 \cdot C^{0.3644}
\end{aligned}
\end{equation}
A key finding emerges when comparing these laws to those of dense models. As shown in Figure~\ref{fig:hyper-scaling}, MoE models favor a significantly larger batch size and a slightly lower learning rate at large compute scales. 
This phenomenon is attributable to MoE's sparsity: during backpropagation, each expert's parameters are updated using only a subset of the tokens in a batch, whereas dense parameters receive gradients from the entire batch~\citep{sun2024hunyuan}.

\begin{figure}[htbp]
    \centering
    \begin{subfigure}[b]{0.32\textwidth} 
        \includegraphics[width=\textwidth]{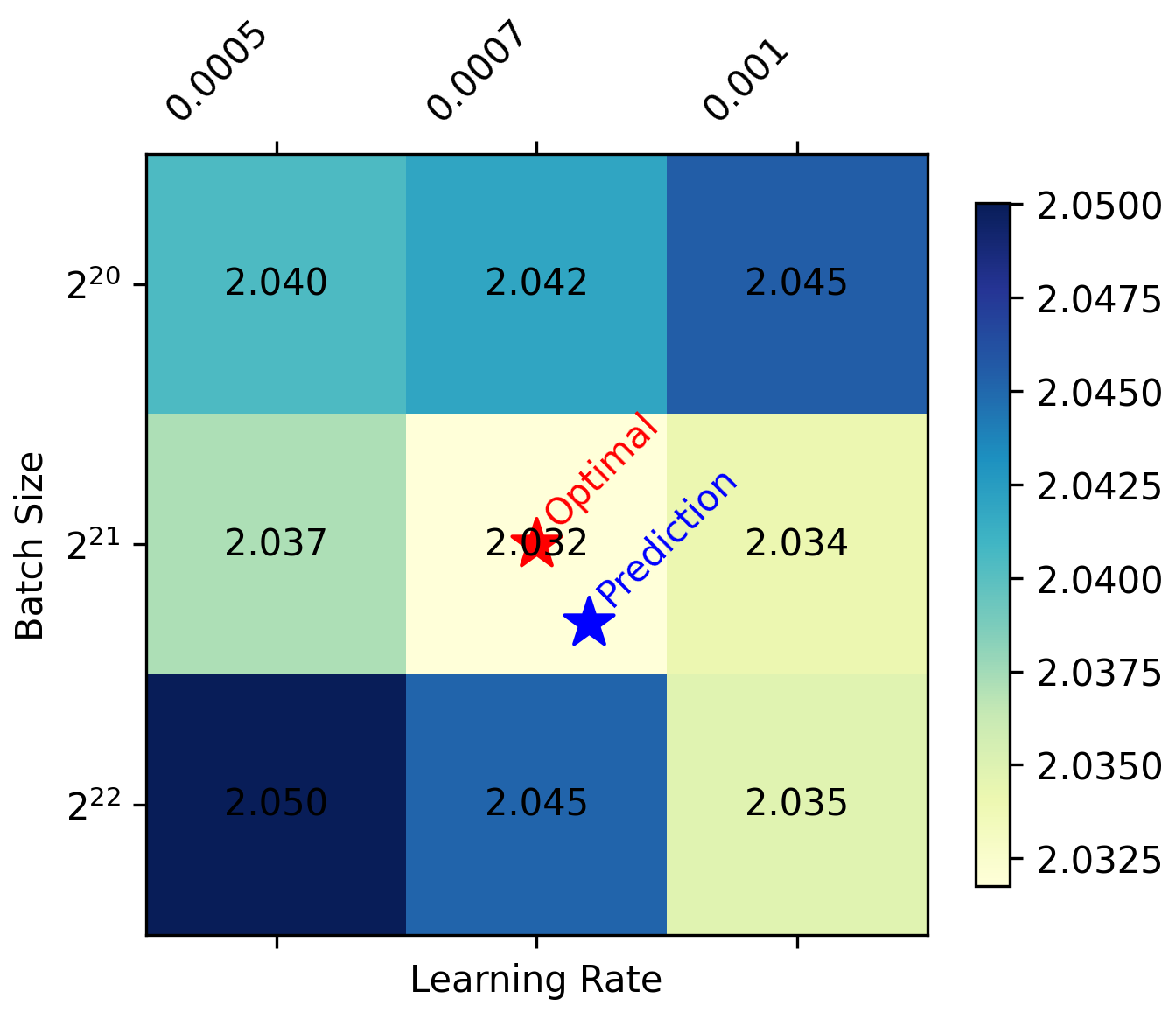}
        \caption{$A$ = 10.9\%}
        \label{fig:lite}
    \end{subfigure}
    \hfill 
    \begin{subfigure}[b]{0.32\textwidth}
        \includegraphics[width=\textwidth]{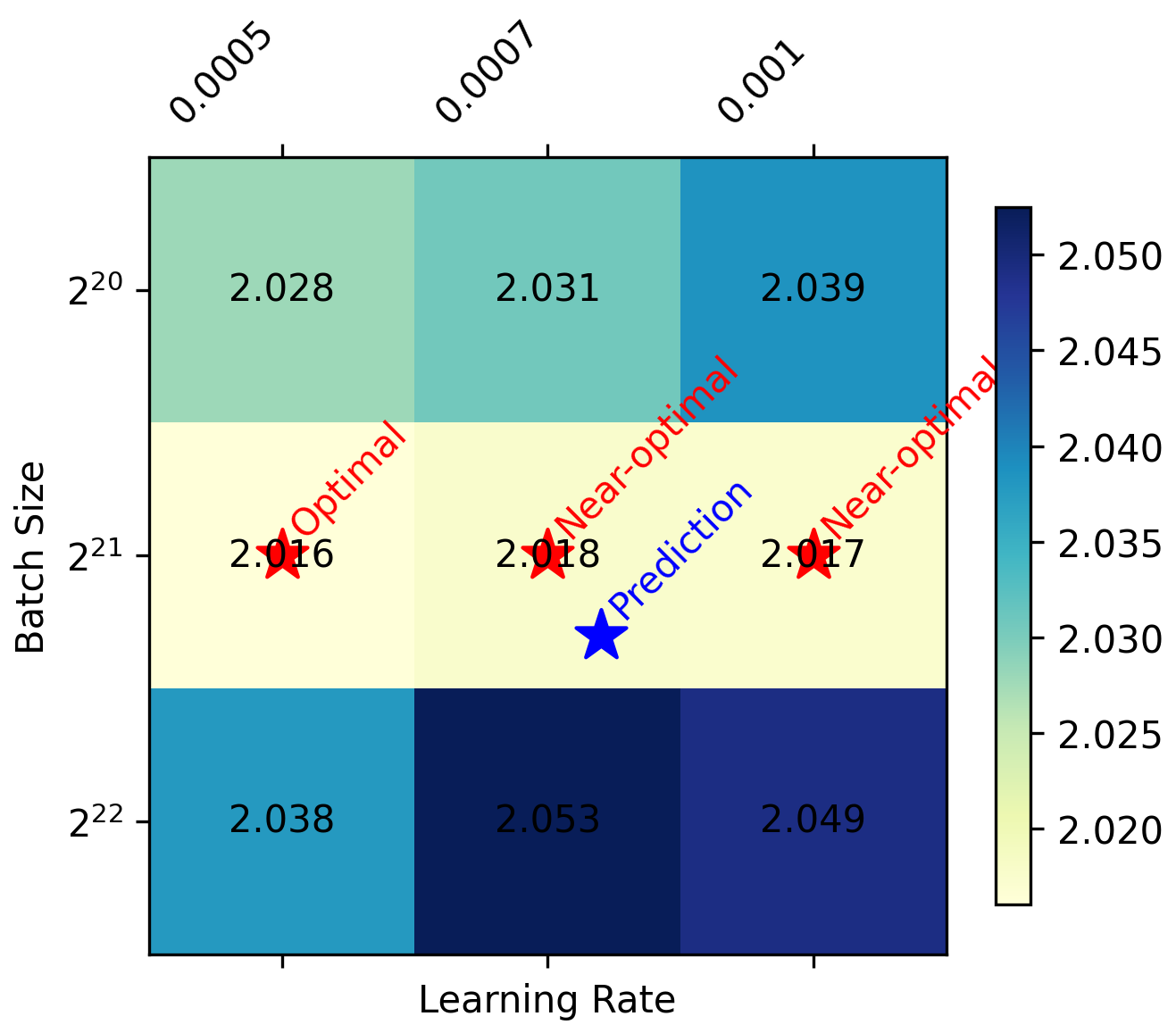}
        \caption{$A$ = 7.8\%}
        \label{fig:plus}
    \end{subfigure}
    \hfill 
    \begin{subfigure}[b]{0.32\textwidth}
        \includegraphics[width=\textwidth]{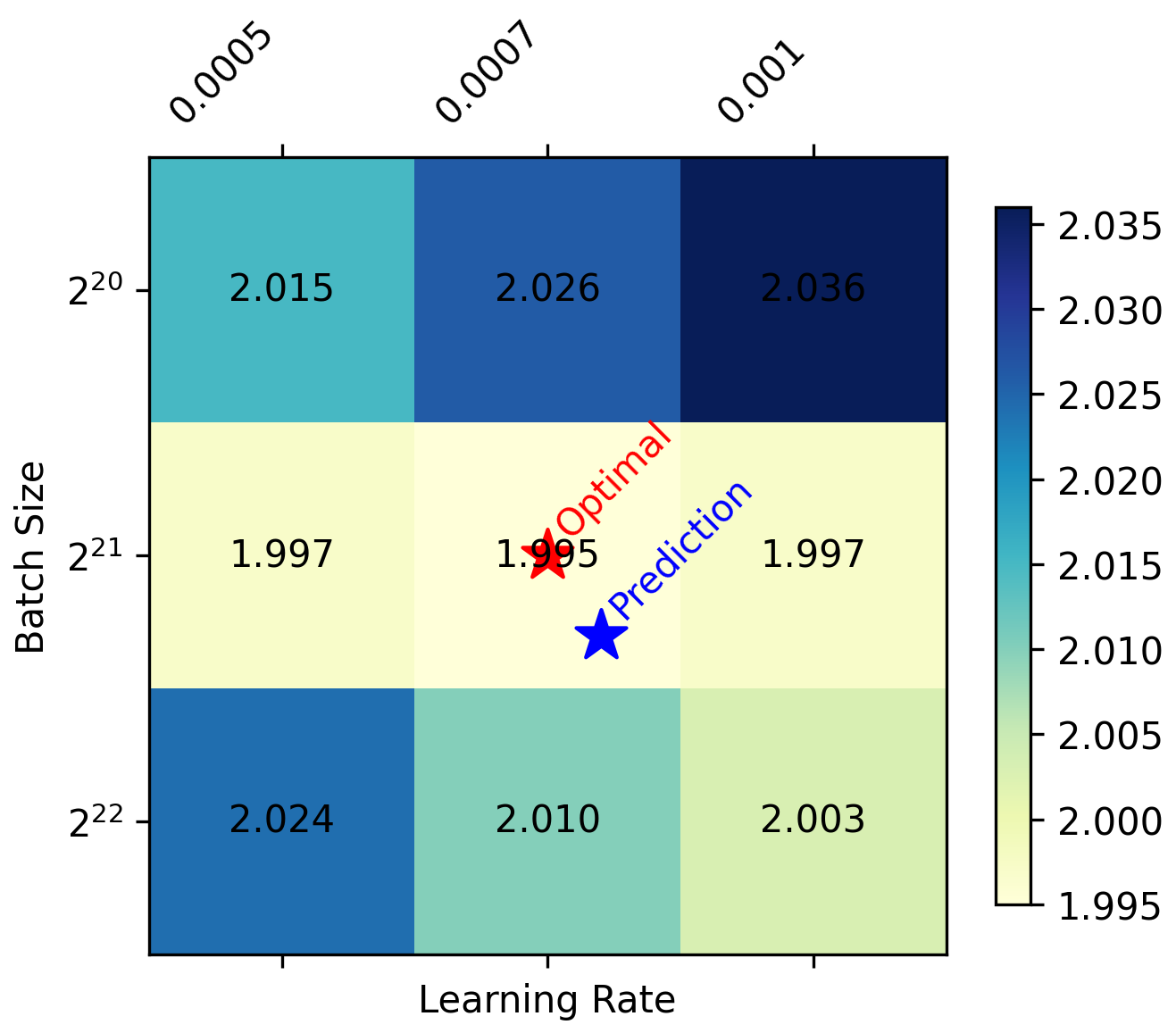}
        \caption{$A$ = 4.7\%}
        \label{fig:max}
    \end{subfigure}
    \caption{\textbf{Validation of MoE hyperparameters scaling laws across different activation ratios ($A$).} ``\textit{Near-optimal}'' refers to hyperparameters achieving a loss within 0.25\% of the optimal ones.}
    \label{fig:hyper-val}
\end{figure}

To validate the generalizability of these laws, we conduct experiments on MoE models with varying activation ratios. We used the derived laws to predict optimal hyperparameters at a compute budget of $3 e20$ FLOPs, after fitting them on data up to $1e20$ FLOPs. As shown in Figure~\ref{fig:hyper-val}, the predicted optimal regions effectively capture the best-performing hyperparameters for activation ratios from 4.7\% to 10.9\%, demonstrating that the laws can be applied to MoE models within this range of activation rates.
This confirms that our hyperparameter scaling laws provide a reliable foundation for exploring diverse MoE architectures under fair and near-optimal training conditions.

\subsection{Scaling laws for MoE Optimal Model-Data Allocation}
\label{sec:scaling-allocation}
\begin{figure}[htbp]
    \centering
    \begin{subfigure}[b]{0.46\textwidth}
        \includegraphics[width=\textwidth]{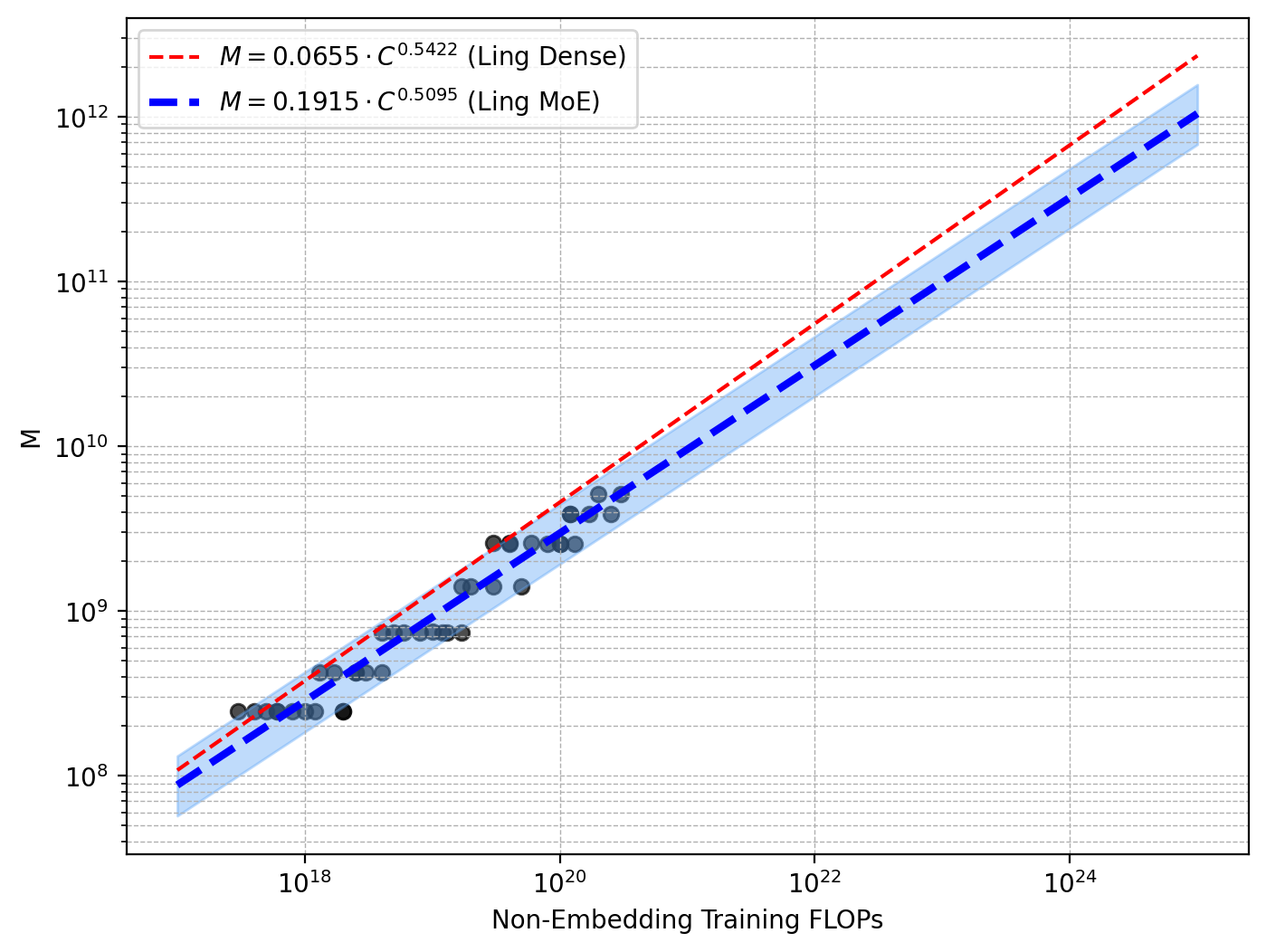}
        \caption{Optimal Model Scale ($M^{\text{opt}}$) Scaling}
    \end{subfigure}
    \hfill 
    \begin{subfigure}[b]{0.46\textwidth} 
        \includegraphics[width=\textwidth]{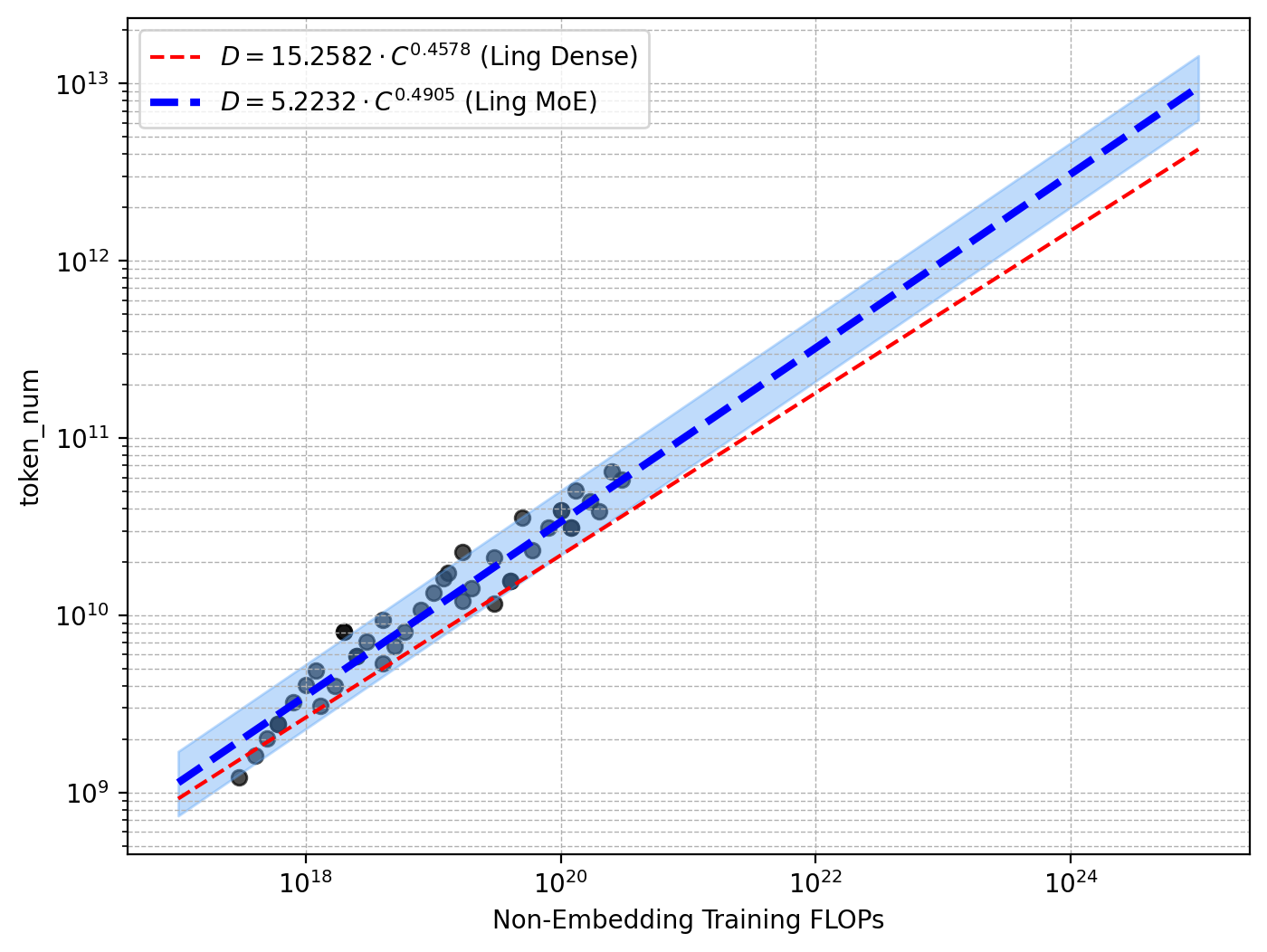}
        \caption{Optimal Data Size ($D^{\text{opt}}$) Scaling}
    \end{subfigure}
    \caption{\textbf{Scaling laws for optimal model scale ($M^{\text{opt}}$) and data size ($D^{\text{opt}}$) on identical datasets.} For a given budget, MoE models (blue) optimally allocate more resources to data and fewer to model size compared to dense models (red).}
    \label{fig:allocation}
\end{figure}

To determine optimal allocation between model size and data size, we analyze loss trajectories across FLOPs budgets from hyperparameter scaling experiments. By identifying the $(M, D)$ combination that yields the minimum loss for a fixed FLOP budget, we derive optimal allocation strategies for specific MoE configurations activating 4 of 64 experts and an additional shared expert ($A=7.8\%$, $G=2$). Crucially, MoE capacity exhibits strong dependence on activation ratio. Thus, this analysis aims to deepen our understanding of MoE architectures and to provide general guidance for model selection in subsequent experiments. The problem can be formally defined as:
\begin{equation}
\begin{aligned}
    (M^{\text{opt}}, D^{\text{opt}}) = \arg\min_{M,D} \mathcal{L}(M,D;C,A,G,S)  \quad \text{s.t.} \quad C=M \cdot D
\end{aligned}
\end{equation}
The resulting scaling laws for the optimal model size ($M^{\text{opt}}$) and data size ($D^{\text{opt}}$) are presented in Figure~\ref{fig:allocation} and summarized in Table~\ref{tab:allocation}. For comparison, we derive the same laws for dense models. Our analysis yields two key insights:
\begin{enumerate}
    \item 
    The optimal allocation coefficients for different architectures are similar and close to 0.5. This aligns with findings from previous studies \citep{bi2024deepseek,hoffmann2022training}, indicating that for compute-optimal training, the budget should be split roughly equally between increasing model size and data volume.
    \item 
    Crucially, at any given compute budget, the optimal MoE model is computationally smaller (lower $M^{\text{opt}}$) but trained on more data (larger $D^{\text{opt}}$) than its optimal dense counterpart. This suggests that MoEs possess greater capacity, enabling them to support larger training datasets with smaller model sizes. In real-world scenarios where data is abundant but computational resources are limited, this is significant for improving efficiency.
\end{enumerate}
While practical training strategies may deviate from this compute-optimal allocation, these scaling laws provide a crucial reference. They offer a principled basis for determining the necessary amount of training data for a given model to approach convergence, designing informative ablation studies, and ultimately, developing more efficient MoE architectures.

\begin{table}[!hbt]
    \caption{Scaling law parameters for compute-optimal allocation of model scale ($M^{\text{opt}}$) and data size ($D^{\text{opt}}$) for MoE and dense models on identical datasets.}
    \label{tab:allocation}
    \centering
    \small
    \begin{tabular}{p{3cm}cc}
        \toprule
        ~ & Optimal Model Scale ($M^{\text{opt}}$)  & Optimal Data Size ($D^{\text{opt}}$) \\ \midrule
        Dense & $M^{\text{opt}} = 0.0655 \cdot C^{0.5422}$ & $D^{\text{opt}} = 15.2582 \cdot C^{0.4578}$ \\ 
        MoE & $M^{\text{opt}} = 0.1915 \cdot C^{0.5095}$ & $D^{\text{opt}} = 5.2232 \cdot C^{0.4905}$ \\ 
        \bottomrule
    \end{tabular}
\end{table}

%% file: sections/3-lever.tex
\section{Efficiency Leverage: Metric for Quantifying MoE Compute-Efficiency}

Next, we define the efficiency leverage and use it to outline our objectives and roadmap.

\paragraph{Definition of Efficiency Leverage. }

To quantify the computational efficiency gain of MoE compared to dense models, we introduce core metric of \textbf{Efficiency Leverage (EL)}. 
Let $\mathcal{X}_{\text{Dense}}$ denote a standard dense architecture and $\mathcal{X}_{\text{MoE}}$ represent a MoE architectures. Within $\mathcal{X}_{\text{MoE}}$, models share identical core configurations (attention mechanisms, expert count, granularity, shared experts), scaled solely through hidden dimensions ($d_{\text{model}}$, $d_{\text{ffn}}$, $d_{\text{expert}}$) and layer count $n_{\text{layer}}$. 
Formally, we define the EL of $\mathcal{X}_{\text{MoE}}$ as the ratio of compute budgets required for the dense and MoE models to achieve the same performance level. While model performance can be quantified through loss values, benchmark scores, or task-specific metrics, this study adopts loss as the primary metric. 
\begin{definition}[Efficiency Leverage]
For $\mathcal{X}_{\text{MoE}}$ achieving minimal loss $\mathcal{L}(C_{\text{moe}}; \mathcal{X}_{\text{MoE}})$ at compute budget $C_{\text{moe}}$, assuming there exists a compute budget $C_{\text{dense}}$ such that $\mathcal{X}_{\text{Dense}}$ attains comparable minimal loss $\mathcal{L}(C_{\text{dense}}; \mathcal{X}_{\text{Dense}})$, we define the efficiency leverage as: 
\begin{equation}
\begin{aligned}
EL(\mathcal{X}_{\text{MoE}} \mid \mathcal{X}_{\text{Dense}}\;;\;C_\text{target}) = \frac{C_{\text{dense}}}{C_{\text{moe}}}, \\
\text{s.t.} \quad \left| \mathcal{L}(C_{\text{moe}};\mathcal{X}_{\text{MoE}}) - \mathcal{L}(C_{\text{dense}};\mathcal{X}_{\text{Dense}}) \right| &\leq \epsilon \quad (\epsilon \to 0)
\end{aligned}
\end{equation}
\end{definition}
Here, the minimal loss achievable by an architecture under specific computational constraints represents its performance ceiling at that scale.  
An EL greater than 1 signifies that the MoE architecture is more computationally efficient than the dense model, achieving the same performance with less compute. Conversely, an EL less than 1 indicates inferior efficiency. 

Following established practice \citep{kaplan2020scaling}, we model the relationship between compute ($C$) and loss ($\mathcal{L}$) with a power law: $\mathcal{L}(C;\mathcal{X}) = \alpha_{\mathcal{X}} C^{-\beta_{\mathcal{X}}}$. This allows us to simplify the EL definition in the compute-optimal training regime~\citep{hoffmann2022training} or similar over-training regime~\citep{gadre2024language}. 
Given the computational cost $C = M \cdot D$, the efficiency leverage simplifies under fixed data size $D$ to the ratio of model scales: $ EL(\mathcal{X}_{\text{MoE}} \mid \mathcal{X}_{\text{Dense}}) \approx {M_{\text{Dense}}}/{M_{\text{MoE}}}$. 
This formulation demonstrates that $\mathrm{EL}$ quantifies the relative model scale of $\mathcal{X}_{\text{MoE}}$ compared to $\mathcal{X}_{\text{Dense}}$ in achieving equivalent performance. 
\textit{In other words, given the model scale of an MoE and its corresponding efficiency leverage, we can directly determine the equivalent dense model scale required to achieve the same performance. }

\paragraph{Objective and Roadmap. }
Existing studies~\citep{ludziejewski2024scaling,abnar2025parameters,clark2022unified} indicate that the model capacity of MoE is significantly influenced by architectural configurations. 
The primary objective of this work is to understand and quantify how MoE architectural choices influence Efficiency Leverage. Our central research question is:
\begin{center}
    \textit{How do the architectural configurations of an MoE model affect its Efficiency Leverage, \\and how does this relationship scale with the computational budget? }
\end{center}
Specifically, our investigation focuses on three critical architectural dimensions\footnote{Other architectural configurations, such as the arrangement of MoE and dense layers, have been verified to have a secondary impact on the efficiency leverage of MoE. See Appendix~\ref{app:setup} and Appendix~\ref{app:add_exp} for details.}: the \textit{Activation Ratio} ($A$), \textit{Expert Granularity} ($G$), and \textit{Shared Expert Ratio} ($S$). They jointly determine the effective capacity of MoE models, and can be used to derive other MoE configurations (\eg the number of experts, the number of actived experts) based on the definitions in Section~\ref{sec:pre-arch}. Our goal is to find the configuration $(A^{\text{opt}}, G^{\text{opt}}, S^{\text{opt}})$ that maximizes EL for a given compute budget $C$:
\begin{equation}
\begin{aligned}
    (A^{\text{opt}}, G^{\text{opt}}, S^{\text{opt}}) = \arg \max_{(A, G, S)\in\mathcal{X}_{\text{MoE}}} EL(\mathcal{X}_{\text{MoE}} \mid \mathcal{X}_{\text{Dense}}\; ; \; C)
\end{aligned}
\end{equation}

To make the analysis tractable, we assume the effects of these dimensions are largely independent and conduct systematic ablation studies. We start with a baseline MoE architecture (2 of 64 experts activated, plus one shared expert) and vary one dimension at a time across a range of compute budgets (from $3e18$ to $3e20$ FLOPs).
To ensure a fair and robust comparison, we leverage the findings from our preliminary studies (Sections~\ref{sec:scaling-hyper} and \ref{sec:scaling-allocation}). For each architecture and compute budget, we determine the reasonable model size ($M$) and data size ($D$) using our derived allocation laws and configure training with optimal hyperparameters from our hyperparameter scaling laws. This rigorous protocol ensures that each architecture is evaluated at or near its peak potential for a given budget, yielding reliable and cost-effective conclusions. Further details on the experimental setup are provided in Appendix~\ref{app:setup}. 
Next, we first empirically analyze the impact of each dimension on EL, and then integrate our empirical findings into a unified scaling law to models the relationship between MoE configurations and the resulting EL.

%% file: sections/4-scaling.tex
\section{Scaling Laws for Efficient MoE Architecture}
\label{sec:scaling}

To achieve greater leverage, we first conduct an extensive empirical study on the architectural configurations of MoE and derive scaling laws for efficient MoE architectures.

\subsection{Empirical Study on the Interplay between Efficiency Leverage and MoE Architecture}
\label{sec:Interplay}

To identify the MoE architecture that maximizes Efficiency Leverage (EL) for a given compute budget, we systematically investigate the impact of several key design choices. These include the activation ratio, expert granularity, shared expert ratio, and other configurations. For each architectural dimension, we vary it systematically while holding other factors and the model scale $M$ constant. 
To ensure a fair comparison, all models are trained following the configurations derived from our scaling laws (Section~\ref{sec:Preliminary}), which specify the ideal model size ($M$), data volume ($D$), and hyperparameters for any given total compute budget. 
Guided by the scaling laws for optimal model-data allocation (defined in Section~\ref{sec:scaling-allocation}, we train each model on over three times its optimal number of tokens. This was done to simulate the overtrained state commonly observed in real-world scenarios. All of trained models can be found in Appendix~\ref{app:model_list}.
Based on the observed training dynamics, we plot the resulting loss curves and EL trends to isolate and quantify the influence of each design choice. 
To ensure robust analysis, we presuppose a standard power-law relationship between FLOPs cost and training loss, and observe the loss of experimental models after sufficient training using the theoretically optimal allocation as a reference.

\subsubsection{Optimal Expert Activation Ratio}
\begin{figure}[htbp]
    \centering
    \begin{subfigure}[b]{0.382\textwidth} 
        \includegraphics[width=\textwidth]{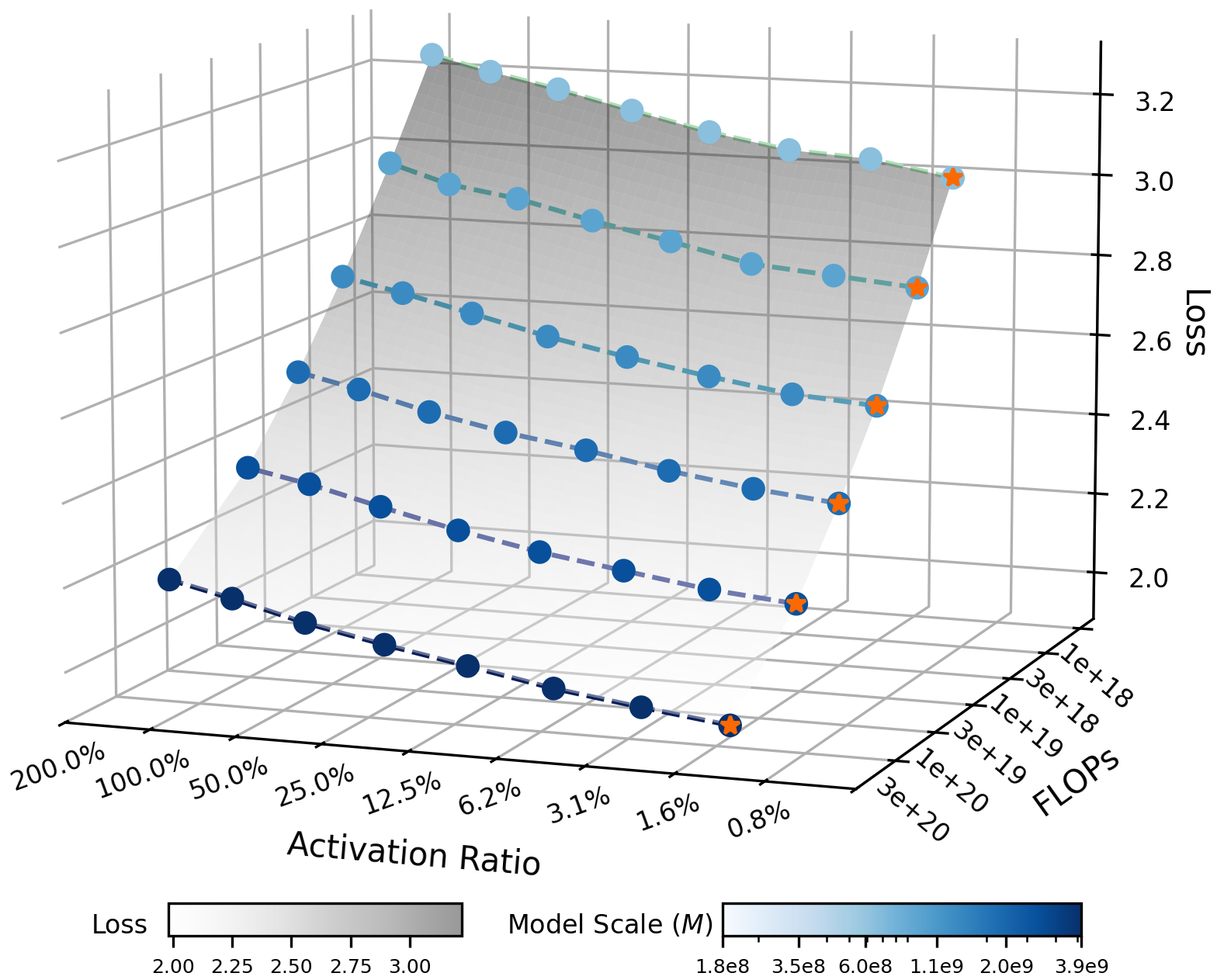}
        \caption{IsoFLOPs curves for varying $A$}
        \label{fig:active-isoflops}
    \end{subfigure}
    \hfill 
    \begin{subfigure}[b]{0.61\textwidth} 
        \includegraphics[width=\textwidth]{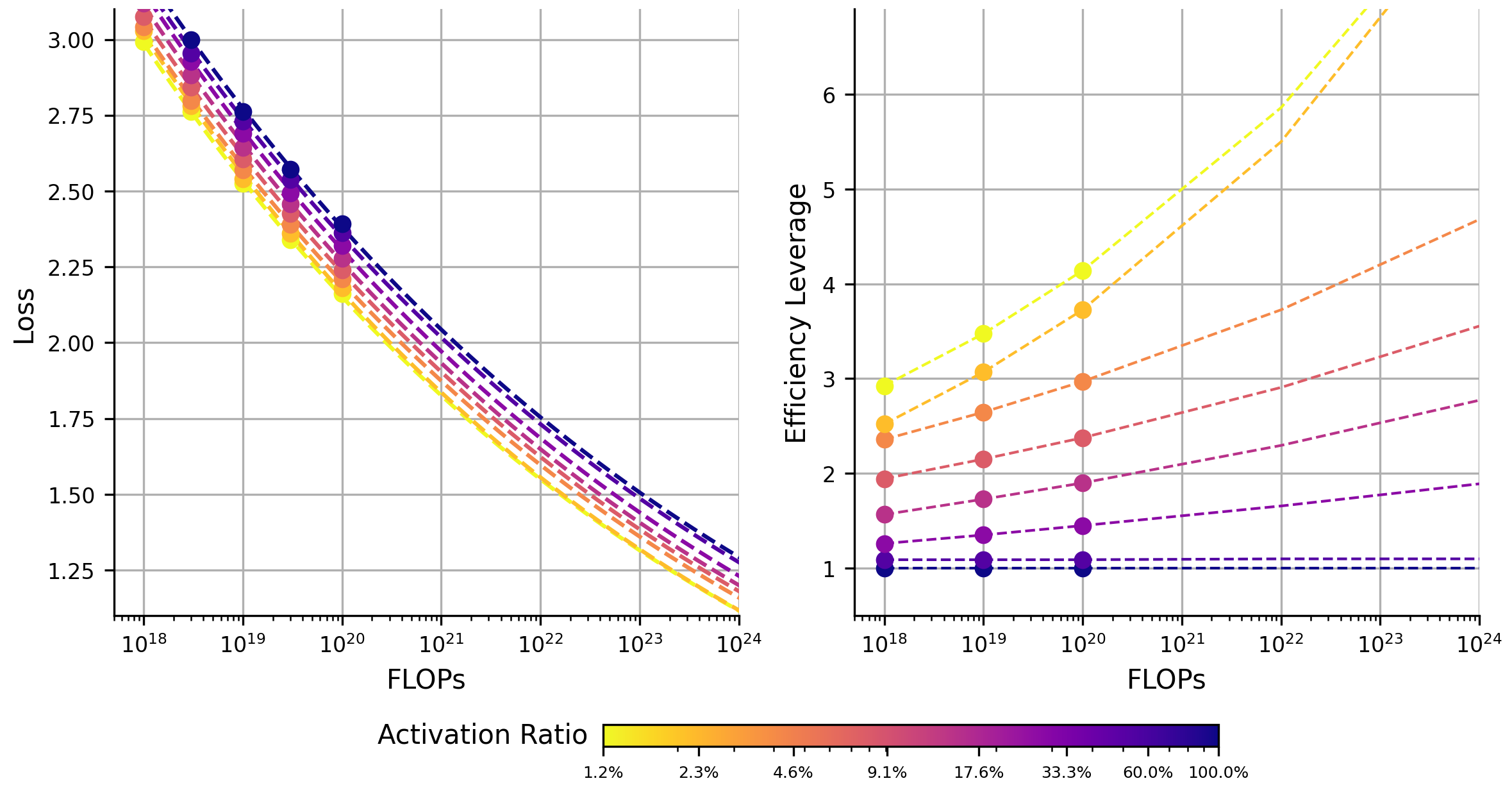}
        \caption{Loss and EL scaling curve over varying $A$.}
        \label{fig:active-loss}
    \end{subfigure}
    \caption{\textbf{Impact of the Activation Ratio $A$ on Loss and Efficiency. } 
    (a) At any fixed compute budget (each colored line), lower activation ratios yield lower loss. The orange stars mark the optimal (lowest) loss point.
    (b) Loss and EL scaling curves illustrate that EL increases with both higher compute budgets and lower activation ratios, showing that MoE advantages are magnified at scale.}
    \label{fig:active}
\end{figure}

We begin by investigating the activation ratio ($A$), a critical factor governing MoE efficiency. Our experimental design isolates the effect of $A$ by holding the computational cost per token ($M$) constant. This is achieved by fixing the number of activated experts and their granularity, while varying the total number of experts in the pool from 2 to 256. This setup allows us to explore a wide range of activation ratios (from 0.8\% to 100\%, where 100\% represents a dense model) without altering the forward pass FLOPs. The optimization problem for a given compute budget $C$ is thus:
\begin{equation}
\begin{aligned}
    A^{\text{opt}} = \arg\min_{A} \mathcal{L}(A;C,M,G,S)
\end{aligned}
\end{equation}
The IsoFLOPs curves, presented in Figure~\ref{fig:active-isoflops}, reveal a clear and consistent trend. Across all tested FLOPs budgets (from $1 e18$ to $3 e20$), loss monotonically decreases with activation ratio, following a power-law pattern. For all configurations, the lowest tested ratio of 0.8\% consistently yields the minimum loss. This finding suggests a core principle: for a fixed computational cost, greater model sparsity (\ie lower activation ratio) leads to higher parameter efficiency.

To quantify this efficiency improvement, we fit a series of loss scaling curves at different activation ratios.
Based on these curves, we compute the efficiency leverage for different activation ratios and FLOPs budgets, as illustrated in Figure~\ref{fig:active-loss}.  The results reveal two key trends.
First, for a fixed FLOPs budget, the EL consistently increases as the activation ratio decreases, indicating that sparse activation can always enhance computational efficiency.
Second, for a fixed activation ratio, the EL grows with the computational budget, demonstrating that the MoE advantage is amplified at larger scales. 
These findings confirm that reducing the activation ratio yields substantial efficiency gains, and these benefits are magnified in large-scale, high-computation regimes. 

\begin{center}
\begin{tcolorbox}[
    colback=gray!10, 
    colframe=black!70, 
    width=0.95\textwidth, %
    title={Key Takeaway 1},
    fonttitle=\bfseries, %
    top=2mm, %
    bottom=2mm, %
    left=4mm, %
    right=4mm %
]
    \begin{itemize}[leftmargin=0.5em] %
        \item \textbf{Monotonic Relationship Between Efficiency and Activation Ratio.} 
        For a fixed computational cost, model performance consistently improves as the activation ratio decreases. This indicates a direct, monotonic relationship between sparsity and efficiency.
        \item \textbf{Efficiency Gains Amplify with Scale.} 
        The efficiency advantage of MoE models (their EL) grows with the total training budget. This highlights their suitability for large-scale training, where their benefits become even more significant.
    \end{itemize}
\end{tcolorbox}
\end{center}

\subsubsection{Optimal Granularity of Experts}
\label{sec:granularity}
\begin{figure}[htbp]
    \centering
    \begin{subfigure}[b]{0.382\textwidth} 
        \includegraphics[width=\textwidth]{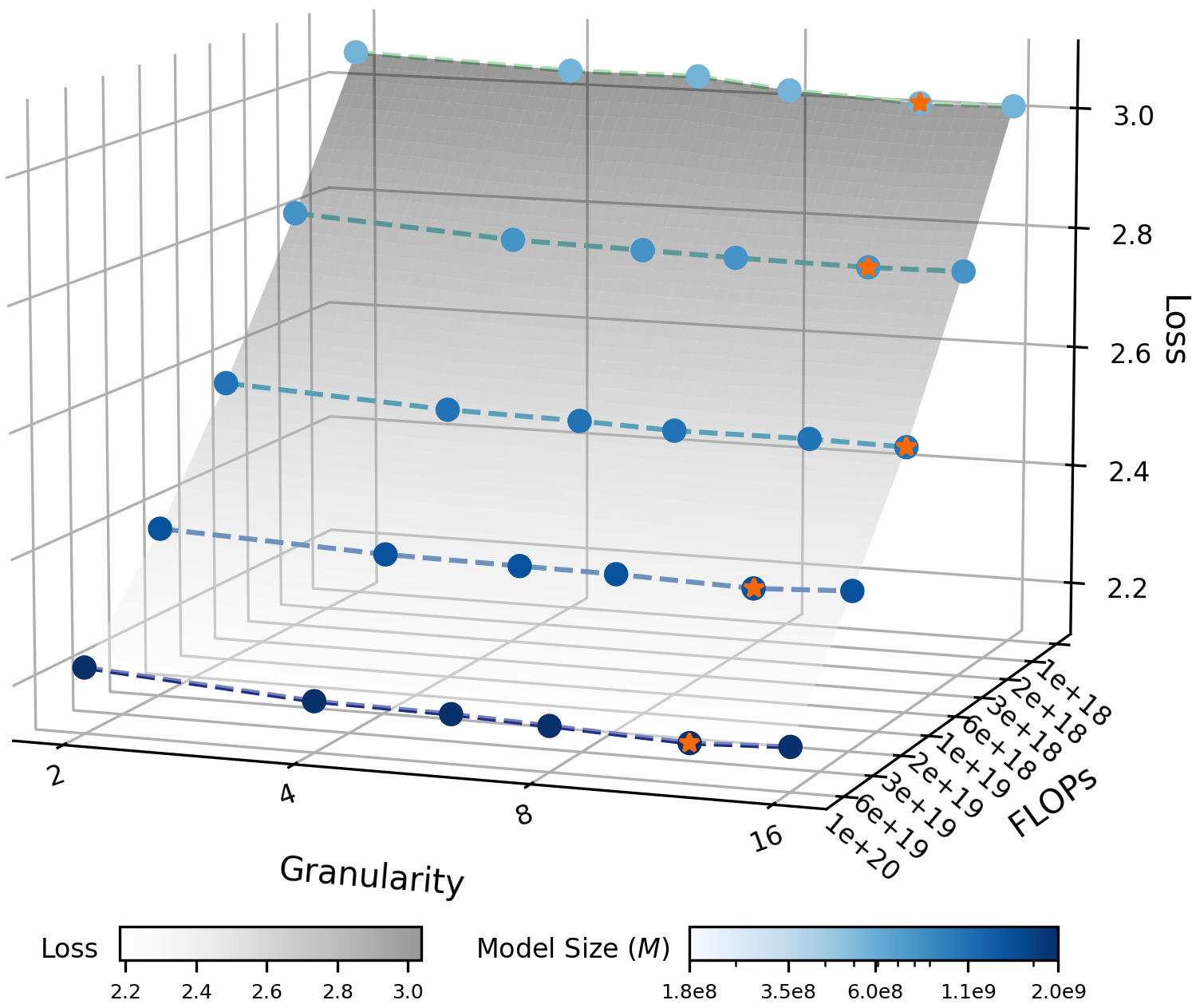}
        \caption{IsoFLOPs curves over varying $G$.}
        \label{fig:granularity}
    \end{subfigure}
    \hfill 
    \begin{subfigure}[b]{0.61\textwidth} 
        \includegraphics[width=\textwidth]{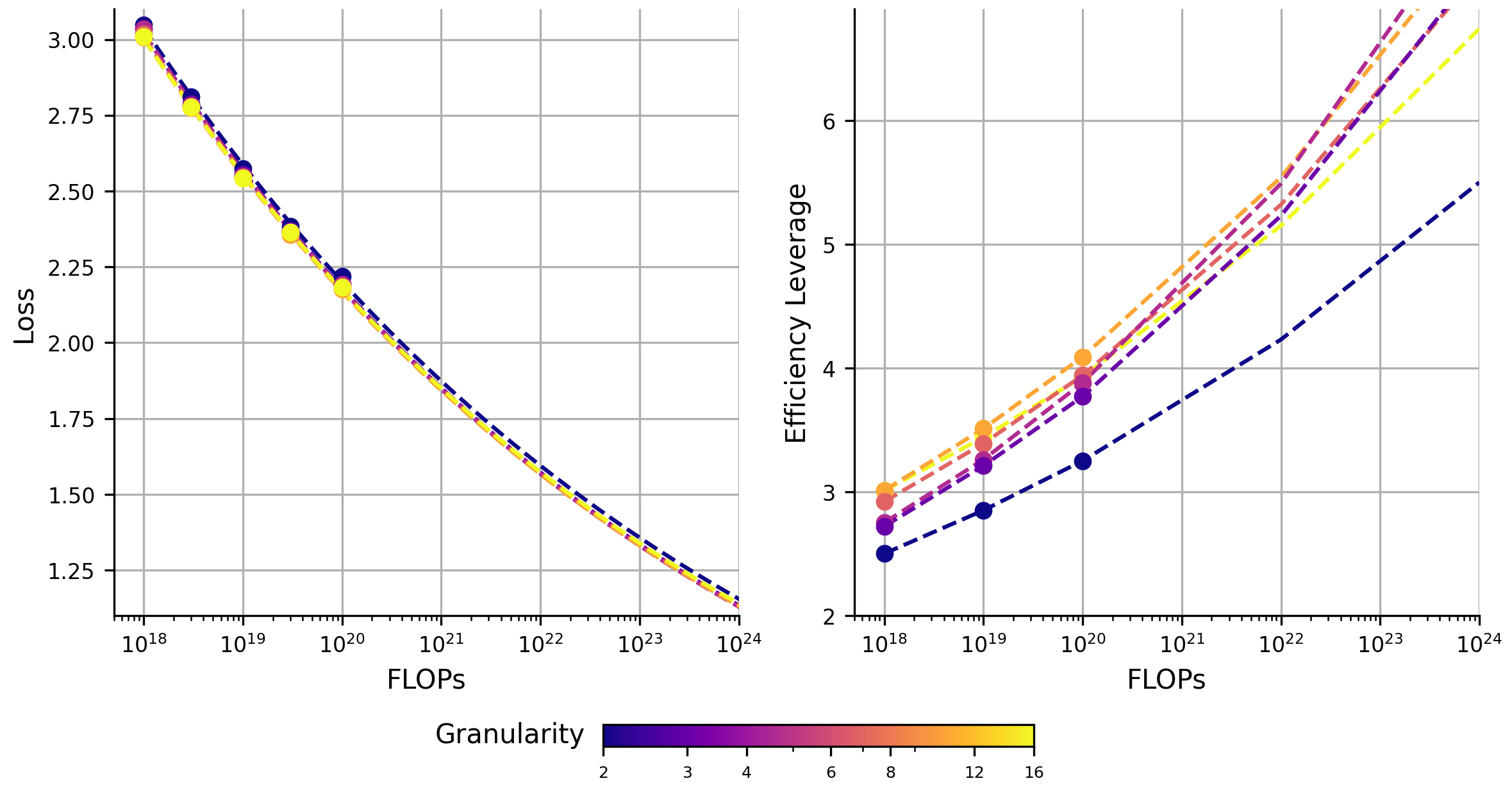}
        \caption{Loss and efficiency leverage scaling curve over varying $G$.}
        \label{fig:granularity-loss}
    \end{subfigure}
    \caption{\textbf{Impact of the Experts Granularity $G$ on Loss and Efficiency. } 
    (a) IsoFLOPs curves reveal a U-shaped (polynomial) relationship between expert granularity and training loss. Orange stars mark the optimal granularity for each FLOPs budget. 
    (b) Loss and EL scaling curves show that MoE efficiency improves as FLOPs increase and expert granularity approaches the optimal range.}
    \label{fig:granularity-all}
\end{figure}

The granularity of experts is a critical factor in the efficiency of MoE. 
While prior works~\citep{ludziejewski2024scaling,dai2024deepseekmoe} suggests that finer-grained experts improve performance, the optimal balance remains an open question. 
To investigate the influence of expert granularity on MoE efficiency, for a fixed model size $M$ and activation ratio $A$, we vary the expert granularity from 2 to 16 by increasing the total number of experts from 64 to 512 while proportionally decreasing the size of each expert to keep computational cost (FLOPs) per token constant. 
This creates a spectrum of models from coarse-grained (fewer, larger experts) to fine-grained (more, smaller experts). By training these models and comparing their final training losses, we can identify the granularity that yields the best performance for a given FLOPs budget. This problem is formalized as:
\begin{equation}
\begin{aligned}
    G^{\text{opt}} = \arg\min_{G} \mathcal{L}(G;C,M,A,S)
\end{aligned}
\end{equation}
where $G^{\text{opt}}$ is the optimal granularity that minimizes the training loss $\mathcal{L}$ under a fixed FLOPs budget $C$, model size $M$, activation ratio $A$, and shared expert ratio $S$.
As shown in Figure~\ref{fig:granularity}, our experiments across a range of FLOPs budgets ($10^{18}$ to $10^{20}$) reveal a distinct trend. For any given budget, as we increase expert granularity, the training loss first decreases and then, after reaching a minimum, begins to increase. This demonstrates the existence of an optimal expert granularity that maximizes computational efficiency of MoE. To further analyze this relationship, we fit loss scaling curves for different granularities (Figure~\ref{fig:granularity-loss}), quantifying their impact on EL.

Our study yields two primary insights:
First, for a fixed FLOPSs budget, the training loss follows a U-shaped (polynomial) relationship with respect to expert granularity, which confirms an optimal point for maximizing model performance per FLOP. 
This finding contrasts with the conclusions of \cite{ludziejewski2024scaling}, and we detail the reasons for this discrepancy in Section~\ref{sec:dis-dif}.
Second, across different FLOPSs budget, the optimal granularity remains within a stable range (around 12 in our experiments), offering a reliable heuristic for model design.
Furthermore, we find that routing balance significantly impacts the choice of optimal granularity. Poor routing balance shifts the optimal point towards coarser granularities and degrades overall model performance (see Appendix~\ref{app:add_exp} for details). This suggests that improving routing mechanisms could unlock the potential of even more fine-grained MoEs, marking a promising direction for future work.

\begin{center}
\begin{tcolorbox}[
    colback=gray!10, 
    colframe=black!70, 
    width=0.95\textwidth, %
    title={Key Takeaway 2},
    fonttitle=\bfseries, %
    top=2mm, %
    bottom=2mm, %
    left=4mm, %
    right=4mm %
]
    \begin{itemize}[leftmargin=0.5em] %
        \item \textbf{Existence of Optimal Expert Granularity.} 
        For a fixed FLOPs budget and model scale, training loss exhibits a U-shaped (polynomial) relationship with expert granularity, indicating an optimum that maximizes efficiency.
        \item \textbf{Stable Range of Optimal Expert Granularity.} 
         The optimal granularity (\eg around 12 in our experiments) is stable across a wide range of FLOPs budgets. However, poor routing balance shifts this optimum toward coarser granularity.
    \end{itemize}
\end{tcolorbox}
\end{center}

\subsubsection{Optimal Shared Expert Ratio}
\begin{figure}[htbp]
    \centering
    \begin{subfigure}[b]{0.382\textwidth} 
        \includegraphics[width=\textwidth]{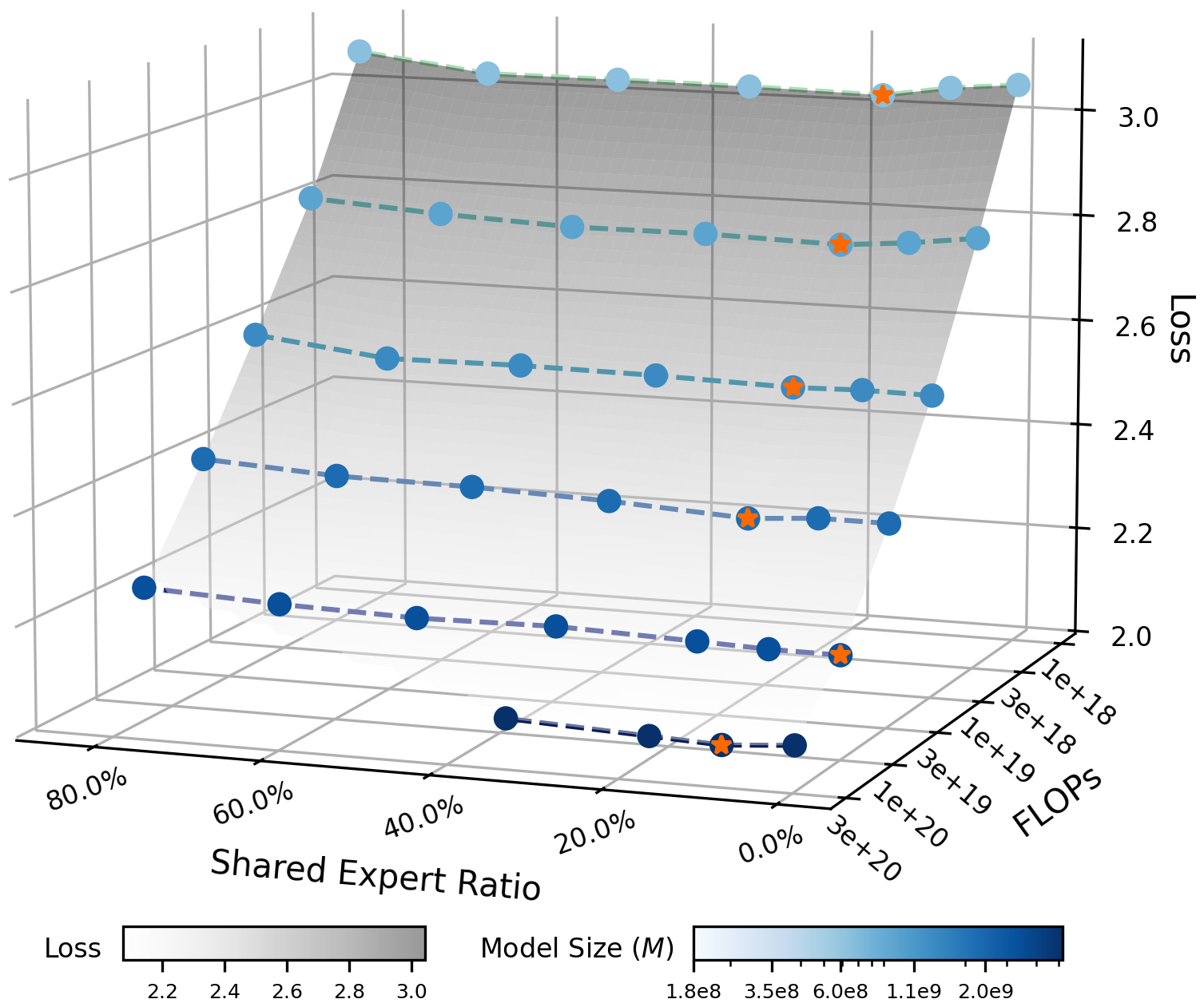}
        \caption{IsoFLOPs curves over varying $G$.}
        \label{fig:sharing}
    \end{subfigure}
    \hfill 
    \begin{subfigure}[b]{0.61\textwidth} 
        \includegraphics[width=\textwidth]{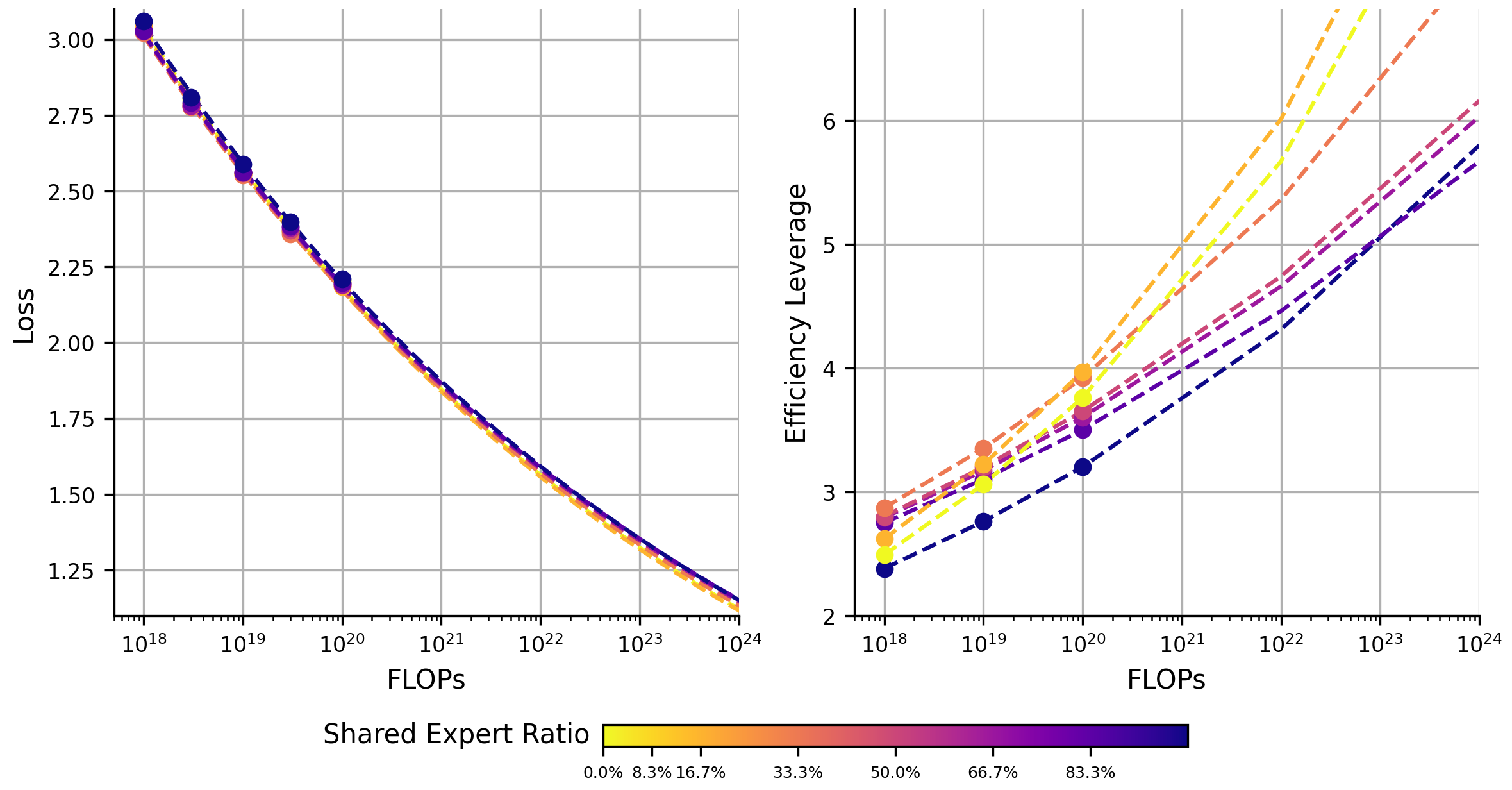}
        \caption{Loss and efficiency leverage scaling curve over varying $G$.}
        \label{fig:sharing-loss}
    \end{subfigure}
    \caption{\textbf{Impact of the Shared Ratio $S$ on Loss and Efficiency. } 
    (a) Loss curves demonstrate that a low, non-zero sharing ratio minimizes training loss, outperforming both no shared experts ($S=0\%$) and highly shared configurations..
    (b) EL analysis reveal that the optimal sharing ratio is higher ($S=16.7\%$) for smaller FLOPs ($< 10^{20}$) and decreases to $S=8.3\%$ for larger FLOPs ($> 10^{20}$).}
    \label{fig:sharing-all}
\end{figure}

Shared experts are always active to capture common knowledge~\citep{dai2024deepseekmoe}. To determine the optimal proportion of shared experts, we designed a series of experiment to isolate the impact of the shared expert ratio $S$.
We fix the total model size $M$, the activation ratio $A$, and the total number of active experts ($E_s + E_a$). We then systematically vary $S$ by substituting routed experts ($E_a$) with shared experts ($E_s$), exploring configurations from fully specialized ($S=0\%$) to highly shared ($S=83.3\%$). This allows us to identify the optimal ratio that minimizes training loss for a given computational budget. The problem is formalized as:
\begin{equation}
\begin{aligned}
    S^{\text{opt}} = \arg\min_{S} \mathcal{L}(S;C,M,A,G)
\end{aligned}
\end{equation}
where $S^{\text{opt}}$ is the optimal shared expert that minimizes the training loss $\mathcal{L}$ under a fixed FLOPs budget $C$, model size $M$, activation ratio $A$, and granularity $G$.
Our experiments, as depicted in Figure~\ref{fig:sharing}, reveal a U-shaped relationship between the shared expert ratio and training loss. The minimum loss is generally achieved at a relatively low shared expert ratio, while having no shared experts ($S=0\%$) usually results in suboptimal performance. Furthermore, we observe a subtle trend where the optimal sharing ratio appears to scale with the compute budget. This is supported by our empirical scaling law (EL) analysis in Figure~\ref{fig:sharing-loss}, which shows that lower FLOPs budgets ($\le 10^{20}$) benefit from a slightly higher sharing ratio ($S=16.7\%$), whereas larger budgets ($> 10^{20}$) achieve greater efficiency with a lower ratio ($S=8.3\%$).

Since large-scale pre-training runs typically exceed $10^{20}$ FLOPs, this suggests a practical heuristic: the optimal design choice is to use the lowest possible non-zero sharing ratio. Assuming the dimensions of shared and regular experts are equal, this can be heuristically implemented by setting the number of shared experts to one.

\begin{center}
\begin{tcolorbox}[
    colback=gray!10, 
    colframe=black!70, 
    width=0.95\textwidth, %
    title={Key Takeaway 3},
    fonttitle=\bfseries, %
    top=2mm, %
    bottom=2mm, %
    left=4mm, %
    right=4mm %
]
    \begin{itemize}[leftmargin=0.5em] %
        \item \textbf{Optimal Sharing Ratio Exhibits a Subtle Scaling Trend.} 
        We identify a subtle scaling trend between the optimal shared expert ratio and the compute budget: the ideal ratio decreases as the compute budget increases.
        \item \textbf{``One Shared Expert'' Rule for Large-Scale Training.} 
        For large-scale pre-training with uniformly sized experts, the optimal design heuristic is to employ a single shared expert. This configuration establishes the minimal non-zero sharing ratio.
    \end{itemize}
\end{tcolorbox}
\end{center}

\subsubsection{Other Configurations of MoE Architecture}

To further optimize the efficiency of MoEs , we also explore two design dimensions: arrangement of MoE and dense layers and compute resource allocation between attention and FFN. The detailed experimental results can be found in Appendix~\ref{app:add_exp}.

First, we analyze replacing the initial MoE layers with dense layers while keeping total FLOPs constant (e.g., 60-layer models with the first 1–3 layers set dense). The experimental results show that replacing the first few layers with dense layers has a minor impact on performance, with efficiency leverage close to 1 within a FLOPs budget of up to $3e{20}$ FLOPs. 
This adjustment reduces the total number of parameters and mitigates routing imbalances, making it a valuable design optimization. As FLOPs budgets increase, the optimal dense proportion also grows; for example, at $1 e18$ FLOPs, the optimal dense proportion is zero. As the compute budget increases to $3 e20$ FLOPs, the optimal dense layer proportion shifts to approximately $2/60$ or $3/60$. 

Second, we explore the impact of computational allocation between the attention mechanism and FFN on MoE efficiency. By constructing models with varying attention-FFN compute allocation and observing performance changes, we find that:  1) An attention FLOPs ratio of 30\%–40\% ensures stable performance, corresponding to default settings in MoE models. 2) Broad ratio adjustments (20\%–50\%) minimally impact performance due to attention's computational density, which increases knowledge density but may raise downstream inference costs

\begin{center}
\begin{tcolorbox}[
    colback=gray!10, 
    colframe=black!70, 
    width=0.95\textwidth, %
    title={Key Takeaway 4},
    fonttitle=\bfseries, %
    top=2mm, %
    bottom=2mm, %
    left=4mm, %
    right=4mm %
]
    \begin{itemize}[leftmargin=0.5em] %
        \item \textbf{Introducing Dense Layers is a Valuable Design Optimization.} 
        Incorporating dense layers in the early stages of MoE has minor impact on efficiency but helps mitigate routing imbalances and reduces overall parameters. The optimal proportion of dense layers increases with higher FLOPs budgets, though it offers limited efficiency gains. 
        \item \textbf{Robustness of Compute Budget Allocation between Attention and FFN} 
        Allocating 30\%-40\% of FLOPs to the attention mechanism achieves optimal or near-optimal performance, with minor impact outside this range. Increasing attention FLOPs proportion enhances knowledge density but reduces downstream inference efficiency.
    \end{itemize}
\end{tcolorbox}
\end{center}

\subsection{Scaling Laws for MoE Efficiency Leverage}
From the observations, both the dense layer and attention ratio have minimal impact on MoE's efficiency leverage. While sharing experts is broadly beneficial, employing a single shared expert typically emerges as the optimal choice in practice.
Thus, we aim to derive a parametric scaling law for predicting the efficiency leverage based on activation ratio $A$, granularity $G$, and FLOPs $C$.

\subsubsection{Separable Scaling Laws for Efficiency Leverage}

\begin{figure}[thbp]
    \centering
    \begin{subfigure}[b]{0.325\textwidth} 
        \includegraphics[width=\textwidth]{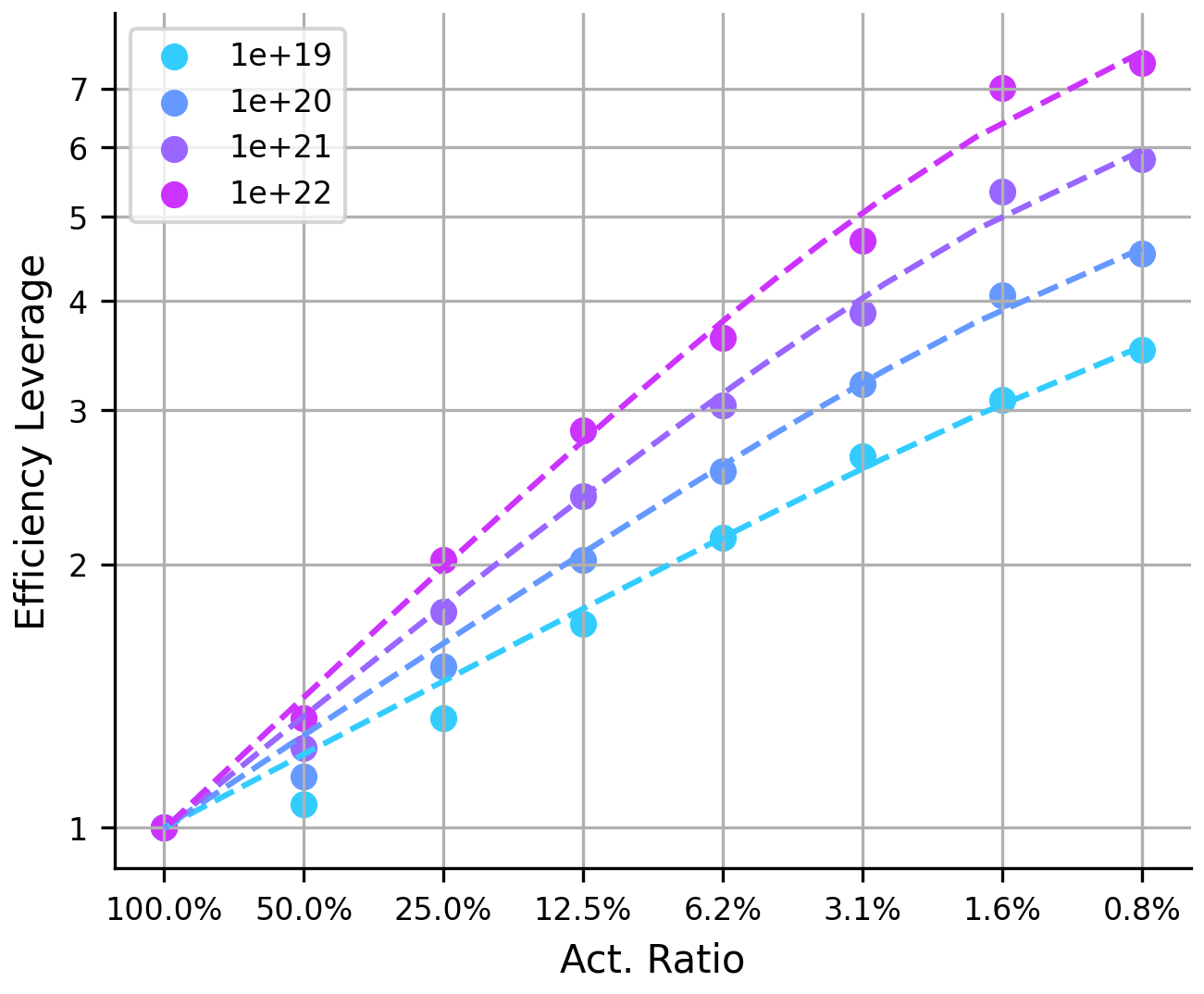}
        \caption{Scaling with Activation Ratio ($A$)}
        \label{fig:scaling-lever-act}
    \end{subfigure}
    \hfill 
    \begin{subfigure}[b]{0.325\textwidth} 
        \includegraphics[width=\textwidth]{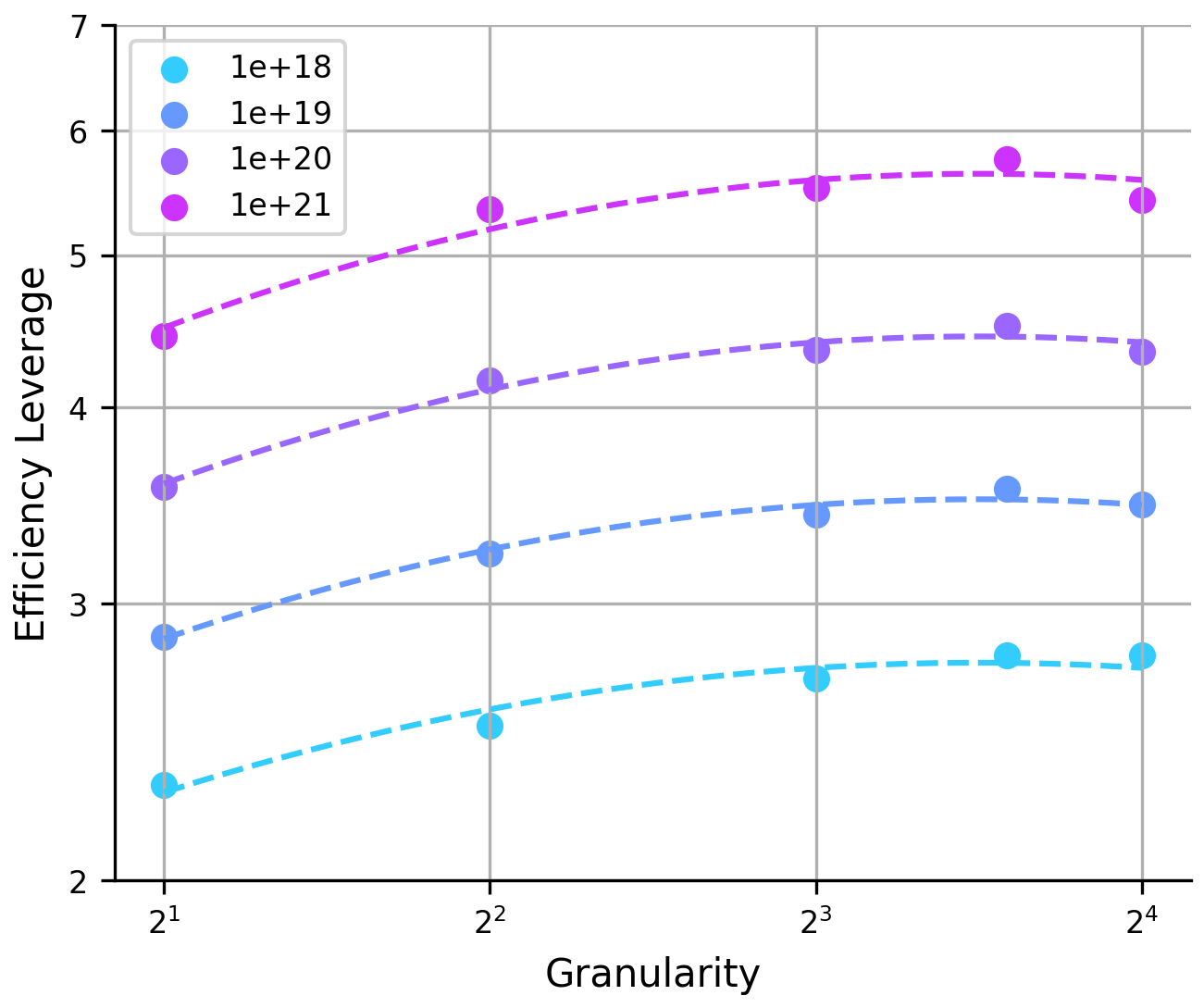}
        \caption{Scaling with Granularity ($G$)}
        \label{fig:scaling-lever-gran}
    \end{subfigure}
    \hfill 
    \begin{subfigure}[b]{0.325\textwidth} 
        \includegraphics[width=\textwidth]{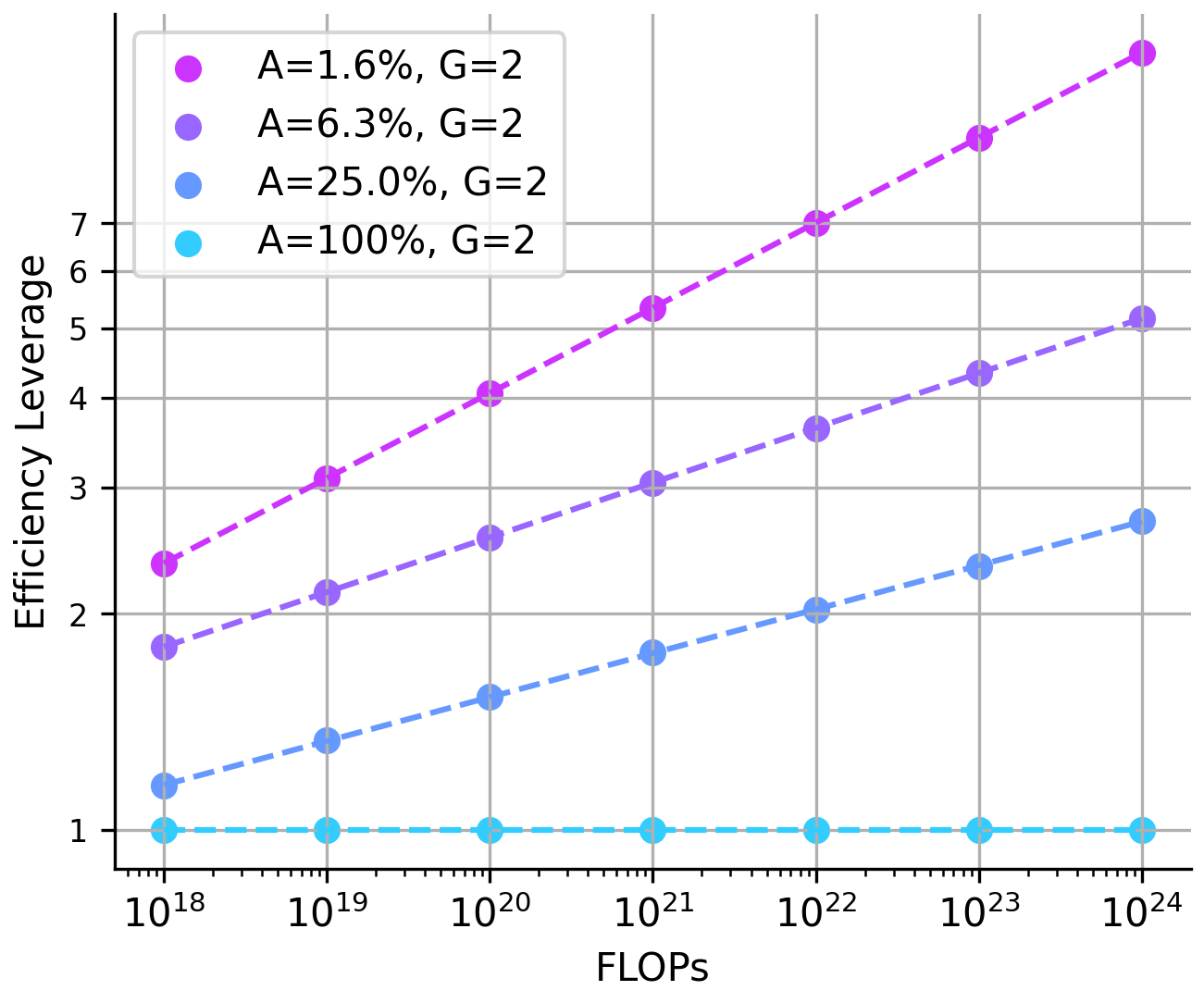}
        \caption{Scaling with Compute Budget ($C$)}
        \label{fig:scaling-lever-flops}
    \end{subfigure}
    \caption{
    \textbf{Scaling Behavior of Efficiency Leverage (EL).} 
    (a) With fixed granularity ($G=2$), EL follows a power law with respect to $A$ across all tested compute budgets ($C$).
    (b) With a fixed activation ratio ($A=3.1\%$), EL's scaling with $G$ conforms to a log-polynomial law across all compute budgets.
    (c) With both activation ratio ($A$) and granularity ($G$) held constant, EL scales with compute according to a standard power law.
    }
    \label{fig:scaling-lever}
\end{figure}

Based on a large amount of previous empirical study in Section~\ref{sec:Interplay}, we collect the MoE efficiency leverages under different settings and summarized them to be presented in Figure~\ref{fig:scaling-lever}.

\paragraph{Interaction of Efficiency Leverage and Activation Ratio.} 
Based on the preceding observations, activation ratio is identified as the primary factor influencing the computational efficiency of MoEs. As illustrated in Figure~\ref{fig:scaling-lever-act}, reducing the activation ratio (\ie increasing sparsity) consistently yields substantial efficiency gains, following a similar power-law relationship across different FLOPs budgets. Consequently, we propose the following hypothesis: for a given FLOPs budget and granularity, there exists a power-law dependence between efficiency leverage and activation ratio.
\begin{equation}
\begin{aligned}
    \log{EL}_{C,G}(\hat A ) & = a_{A} \log \hat A, \;\;\;\; \text{i.e.} \;\; {EL}_{C,G}(\hat A) = \hat A^{a_A}, \\
    \text{where} \;\; \frac{1}{\hat{A}} =& \frac{1}{A+({1/A_{start}-1/A_{max}})^{-1}} + \frac{1}{A_{\text{max}}},
\end{aligned}
\label{eq:scaling-A}
\end{equation}
where $ \hat{A} $ is a saturating transformation of $ A $, as defined in \cite{clark2022unified}, and we set the lower bound of meaningful activation ratio as 0. 
Clearly, when $A = 1$, we have $EL = 1$, indicating that the EL of the dense model is 1, which satisfies the dense equivalence. 
We fit Eq.~\eqref{eq:scaling-A} to each FLOPs budget and plot the predictions for varying activation ratios as dotted lines in the Figure~\ref{fig:scaling-lever-act}. The predictions align well with the observed data. 
Notably, we observe that $a_{A}$ increases as $A$ decreases and $C$ increases. This trend corresponds to a diminishing benefit from increased sparsity, consistent with findings from prior research~\citep{clark2022unified}. 
Additionally, $a_{A}$ also increases with $C$, suggesting a greater benefit from the bigger compute budget~\citep{ludziejewski2024scaling}. We will analyze the relationship between FLOPs and EL in the following paragraph.

\paragraph{Interaction of Efficiency Leverage and Granularity.} 
As previously mentioned, the relationship between expert granularity and EL does not exhibit an ideal power-law pattern. Instead, there exists an optimal granularity that maximizes the EL. Based on this observation, we hypothesize that under fixed FLOPs budget $C$ and activation ratio $A$, the relationship between EL and granularity $G$ follows a log-polynomial pattern:
\begin{equation}
\begin{aligned}
    \log{EL}_{C,A}(G) =& a_{G} + b_G \left( \log G \left( \log G + c_{G}\right) \right),
\end{aligned}
\label{eq:scaling-G}
\end{equation}
where $a_{G}$ is the granularity-independent base EL. It indicates the theoretical EL value of the model when the expert granularity is 1. $b_{G}$ controls the strength of the curvature in the relationship between EL and granularity, reflecting the sensitivity of the model architecture to changes in expert granularity. $c_{G}$ directly determines the location of the optimal granularity that maximizes EL. 
We fit Eq.~\eqref{eq:scaling-G} to each FLOPs budget and plot the predictions for varying granularity as dotted lines in the Figure~\ref{fig:scaling-lever-gran}.
As shown in the figure, the curves under different FLOPs are identical (i.e., with similar values of $  b_{G}  $ and $  c_{G}  $), indicating that the impact of expert granularity on MoE efficiency remains consistent across various computational budgets. 

\paragraph{Interaction of Efficiency Leverage and Compute Budget.} 
Based on the analysis presented in Section~\ref{sec:Interplay}, we observe that the efficiency advantage of MoE increases as the computational budget grows. To formalize the relationship between the FLOPs budget and Efficiency Leverage, we assume a standard power-law pattern as follows:
\begin{equation}
\begin{aligned}
    \log{EL}_{A,G}(C) & = a_{C} \log C + c_C, \;\;\;\; \text{i.e.} \;\; {EL}_{A,G}(C)  = \exp(c_C) \cdot C^{a_{C}},
\end{aligned}
\label{eq:scaling-C}
\end{equation}
where $ a_{C} $ reflects the scaling capability of MoE efficiency with respect to the computational budget under given configurations $ A $ and $G$. 
We collect the values of the EL corresponding to different model architectures under the granularity setting of 2, and fit Eq.~\eqref{eq:scaling-G} to each architectures. 
The predictions for varying granularity are plotted as dotted lines in the Figure~\ref{fig:scaling-lever-flops}. The results indicate that all tested MoE architectures show a trend of higher EL as the FLOPs budget increases, demonstrating the potential of MoE in large-scale pre-training.

\subsubsection{Joint Scaling Law for Efficiency Leverage}
Based on the preceding observations, we arrive at the following key insights:
\begin{itemize}[] %
    \item The activation ratio (or sparsity) is the core determinant of MoE efficiency, establishing its foundational power-law scaling.
    \item Building upon this power law, expert granularity adds a non-linear adjustment that operates independently of the compute budget.
    \item Furthermore, the efficiency advantage of MoE over dense models is amplified by the compute budget $C$ through the power-law term.
\end{itemize}
To unify these interconnected effects, we derive a joint scaling law for EL, formulated as follows:
\begin{equation}
\begin{aligned}
    {EL}(A,G,C) &= \hat A^{\alpha + \gamma(\log G)^2 + \beta \log G}, \\
\end{aligned}
\label{eq:all}
\end{equation}
where $\alpha = a + d \log C$  is the compute-dependent exponent that captures the primary power-law relationship between EL and activation ratio. The term $a$ represents the base scaling exponent at a reference compute budget, while $d$ is a positive constant that quantifies how the EL is amplified by a larger compute budget $C$. 
The parameters $\beta$ and $\gamma$ model the non-linear impact of granularity $G$. This quadratic form in $\log G$ directly reflects the log-polynomial pattern observed in our initial analysis, capturing the existence of an optimal granularity. 

\begin{figure}[htbp]
    \centering
    \begin{subfigure}[b]{0.6\textwidth} 
        \includegraphics[width=\textwidth]{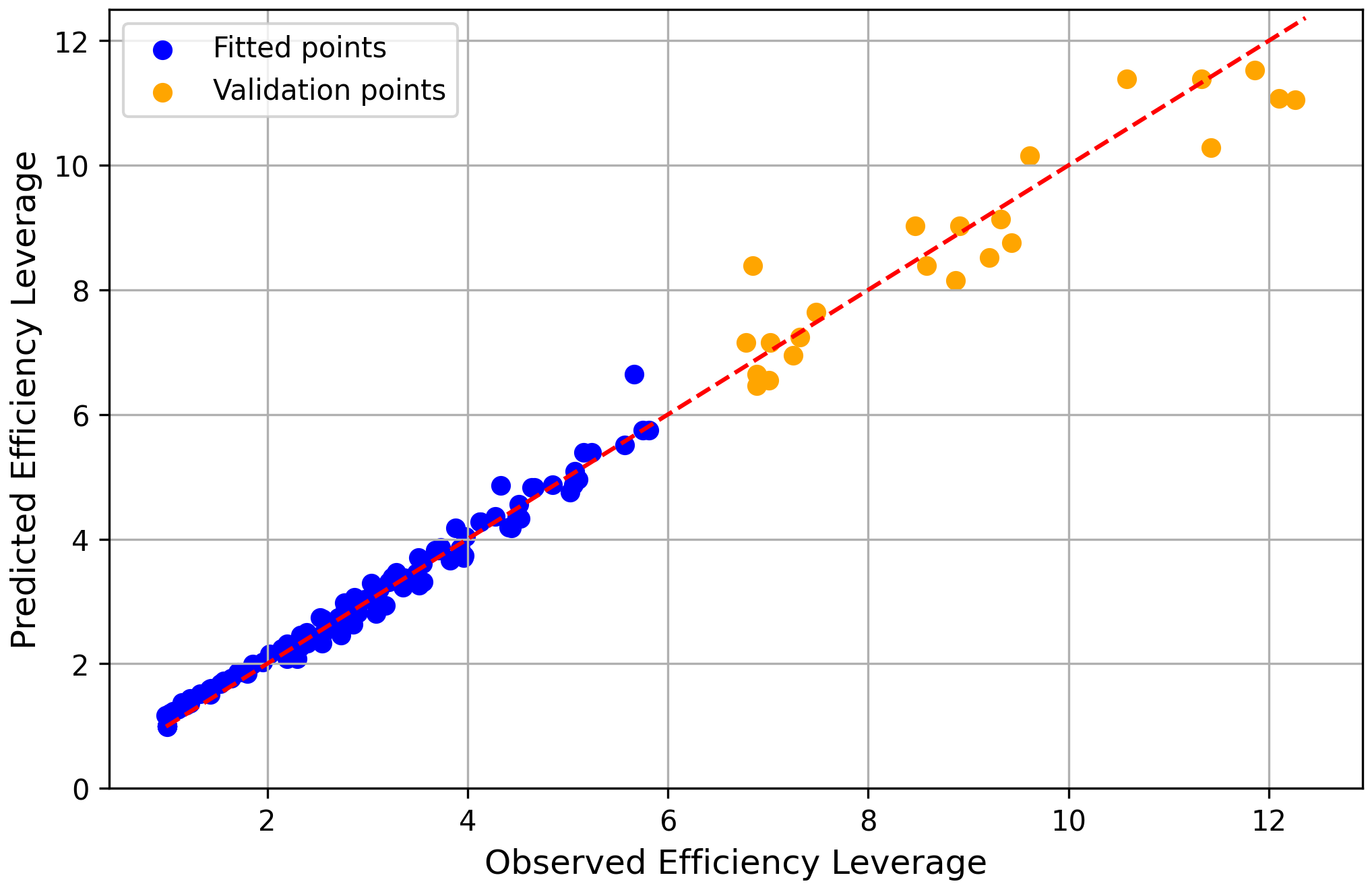}
    \end{subfigure}
    \caption{\textbf{Validation of the Scaling Laws for MoE Efficiency Leverage.} 
    We fit Eq.~\eqref{eq:all} to the data points with an efficiency leverage of less than 6, using the remaining points as the validation set.
    }
    \label{fig:Validation}
\end{figure}

To validate the proposed scaling law for EL, we fit Eq.~\eqref{eq:all} using Huber loss and the BFGS optimization algorithm~\citep{hoffmann2022training}. We use data points with an EL factor below 6 for training, while those are reserved as a validation set. We depict the results in Figure~\ref{fig:Validation}. The values are presented in Table~\ref{tab:Validation}.
The alignment between the scaling law and both the training data and validation set provides strong empirical support for the proposed relationship. More importantly, the scaling law exhibits remarkable extrapolation capabilities, as it accurately models performance trends for high-leverage validation points outside the training range.
These results confirm that Eq.~\eqref{eq:all} effectively captures the underlying interaction between MoE architecture and EL.

Furthermore, we select 1e22 FLOPs compute budget, and apply our fitted scaling laws to predict efficiency leverage across various MoE configurations. 
As shown in Figure~\ref{fig:hotmap-lever}, our analysis predicts that an efficiency leverage (EL) exceeding 7x can be achieved at a budget of 1e22 FLOPs with an activation ratio of 3.1\% and a granularity of 12. The subsequent section provides experimental validation for this specific claim.

\begin{table}[h]
\small 
  \caption{\textbf{Values of the Fitted Coefficients.}}
  \label{tab:Validation}
  \label{table:sample-table}
  \centering
  \setlength{\tabcolsep}{5pt} %
  \begin{tabular}{*{6}{>{\centering\arraybackslash}p{1.5cm}}} %
    \toprule
    $a$ & $d$ & $\gamma$ & $\beta$ & $A_{start}$ & $A_{max}$ \\
    \midrule
    1.23 & -7.61e-2 & 1.67e-2 & -1.17e-1 & 1.63e-2 & 5.28e+16 \\
    \bottomrule
  \end{tabular}
\end{table}

%% file: sections/5-validation.tex
\section{\texttt{Ling-mini-beta}: More Efficient MoE Language Model}

Based on the findings in Section~\ref{sec:scaling}, we identify the efficient architectural configuration within the current MoE framework. 
To validate the effectiveness of this configuration, we design an MoE model with 0.855 billion active parameters out of a total of 17.5 billion (referred to as ``\mini,'' a pilot model for Ling-2.0 series) and test it using 1 trillion tokens of training data. \mini is configured with a granularity $G$ of 12 and an activation ratio $A$ of just 3.4\%.
Referring to Figure~\ref{fig:hotmap-lever}, at the $1e22$ FLOPs compute budget, we hypothesize that \mini achieves \textit{\textbf{more than 7× in compute-efficiency leverage}} over a comparable dense model.
Concurrently, we train a traditional dense model with 6.1 billion parameters (named ``\textit{\dense}'') for comparison. This section presents a detailed analysis of the performance differences between \mini and the conventional dense model \dense, highlighting that the active parameter count, training costs, and downstream inference costs of \dense are more than seven times those of \mini.

\subsection{Model and Training Details}

The architectures of \mini and \dense are given in Table~\ref{table:mini}. Other settings include: 
\begin{itemize}[] %
    \item \textbf{Model Setting.} 
    \mini adopts the same GQA~\citep{ainslie2023gqa} attention architecture as \dense, with the only difference being the extension of the original FFN layers to MoE layers. Additionally, both \mini and \dense employ Rotary Position Embedding (RoPE)~\citep{su2024roformer} and supports a sequence length of 8K. 
    \item \textbf{Training Data.} 
    The training data is sourced from a large-scale multilingual corpus created by the Ling Team, primarily covering English and Chinese, while also including various other languages. This corpus encompasses web text, mathematical materials, programming scripts, published literature, and diverse textual content. To validate model performance, we extracted a 1T-token subset from this corpus for training.
    \item \textbf{Training Setting.} 
    Both \mini and \dense were trained using the AdamW optimizer~\citep{adamw} with hyperparameters set as follows: $\beta_1 = 0.9$, $\beta_2 = 0.95$, and weight decay of 0.1. Gradient clipping norm is set to 1.0. The learning rate schedule employs a WSD (warmup-stable-decay) strategy~\citep{hu2024minicpm}. According to the hyperparameter scaling laws for dense and MoE models, the maximum learning rates were set to $3.78e{-4}$ for \mini and $2.93e{-4}$ for \dense. The batch sizes were configured as 1792 and 2048, respectively.
\end{itemize}
More details about model training setting can be found in the Appendix~\ref{app:setup}.

\begin{table}[h]
\small
  \caption{\textbf{Detailed Architectures of \mini and Dense Model for Comparison.} 
  We determined the architecture of the \mini based on the findings of Section~\ref{sec:scaling}.}
  \label{table:mini}
  \centering
  \setlength{\tabcolsep}{5pt} %
  \begin{tabular}{cccccccccccc}
    \toprule
    \textbf{Model} & $n_{layers}$ & $d_{model}$ & $d_{ffn}$ & $d_{expert}$ &$n_{heads}$ & $n_{kv\_head}$ & $E$ & $E_a$ & $E_s$ & $N$ & $N_a$ \\
    \midrule
    \textbf{Dense 6.1B} & 28 & 4096 & 14336 & - & 32 & 8 & - & - & - & 6.11B & 6.11B  \\
    \textbf{\mini (A0.8B)} & 20 & 2048 & 5120 & 384 & 16 & 4 & 384 & 12 & 1 & 17.5B & 0.85B \\
    \bottomrule
  \end{tabular}
\end{table}

\subsection{Training Dynamics}

\begin{figure}[htbp]
    \centering
    \begin{subfigure}[b]{0.63\textwidth} 
        \includegraphics[width=\textwidth]{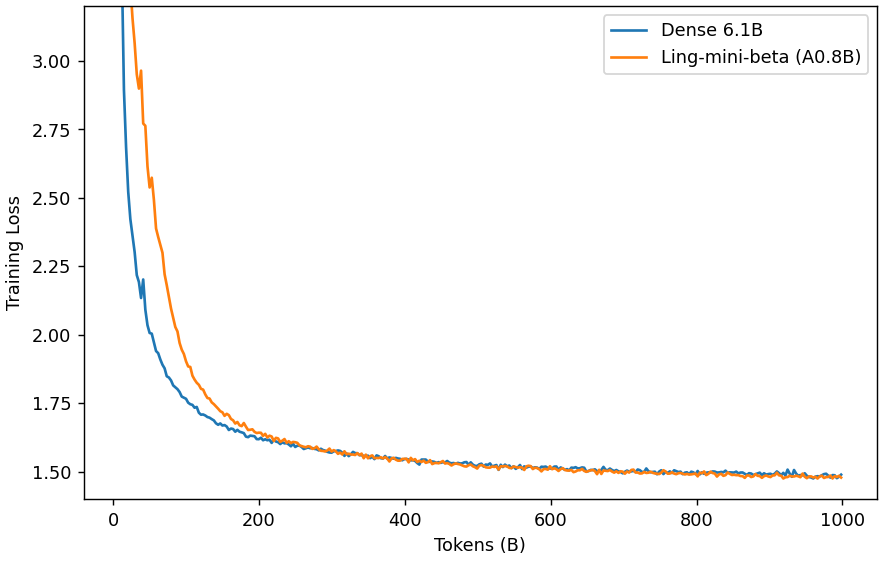}
        \caption{Overview of the loss throughout the training process.}
        \label{fig:mini-loss-all}
    \end{subfigure}
    \hfill 
    \begin{subfigure}[b]{0.35\textwidth} 
        \includegraphics[width=\textwidth]{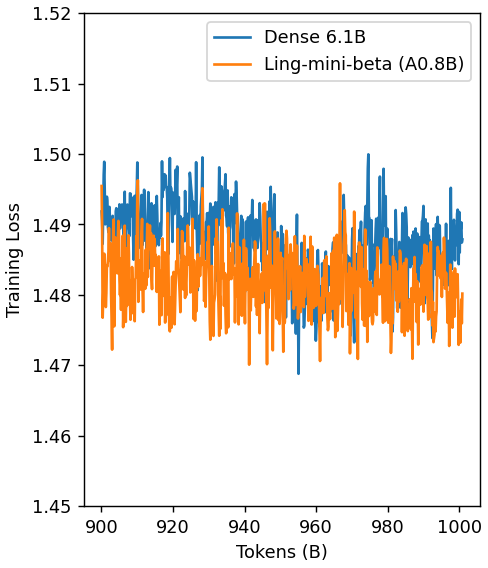}
        \caption{Zoom in the loss at the training end.}
        \label{fig:mini-loss-100B}
    \end{subfigure}
    \caption{\textbf{Dynamic of Training Loss.} 
    (a) Comparing the training processes of \mini and the dense model shows that the dense model converges faster in the early stages. However, while \mini starts slower, its training loss becomes nearly equivalent to the dense model’s after sufficient training.
    (b) Zooming in on the training loss for the final 100B tokens, the training loss difference between \mini and \dense is less than 0.01, demonstrating over 7x efficiency gains for \mini with comparable performance to the dense model.}
    \label{fig:mini-loss}
\end{figure}

\paragraph{The Dynamic of Training Loss} 
The training loss curves for \mini and \dense, shown in Figure~\ref{fig:mini-loss}, illustrate a clear difference in their convergence behavior. The dense model exhibits faster convergence during the early training phases, indicating an aptitude for rapid initial learning. In contrast, \mini's loss decreases more gradually at the start. However, over the full course of training, \mini steadily improves and ultimately achieves a performance level comparable to that of the dense model, highlighting its ability to reach high performance with sufficient training.
Focusing on the final 100 billion tokens of training provides further insight. In this concluding stage, the performance gap between \mini and \dense narrows to a negligible difference of about 0.01 in loss value. This confirms that \mini can nearly match the dense model's effectiveness while operating with significantly fewer computational resources. Crucially, this near-equal performance underscores \mini's ability to deliver over 7x gains in training efficiency, making it a highly cost-effective and powerful alternative for large-scale pre-training.

\paragraph{The Dynamic of Benchmarks}
The training process for both \mini and \dense was monitored by comparing their performance on standard benchmarks. The data reveals a clear and consistent trend: the two models improved almost synchronously. At no point during training did one model show a decisive or lasting advantage over the other. This lockstep progression continued until the end of the training cycle, where they posted nearly identical final scores on the evaluation leaderboard. This synchronous dynamic and convergent outcome suggest a fundamental parity in their learning efficiency and final performance ceiling under our experimental conditions.

\begin{figure*}[t]
    \centering
    \includegraphics[width=\textwidth]{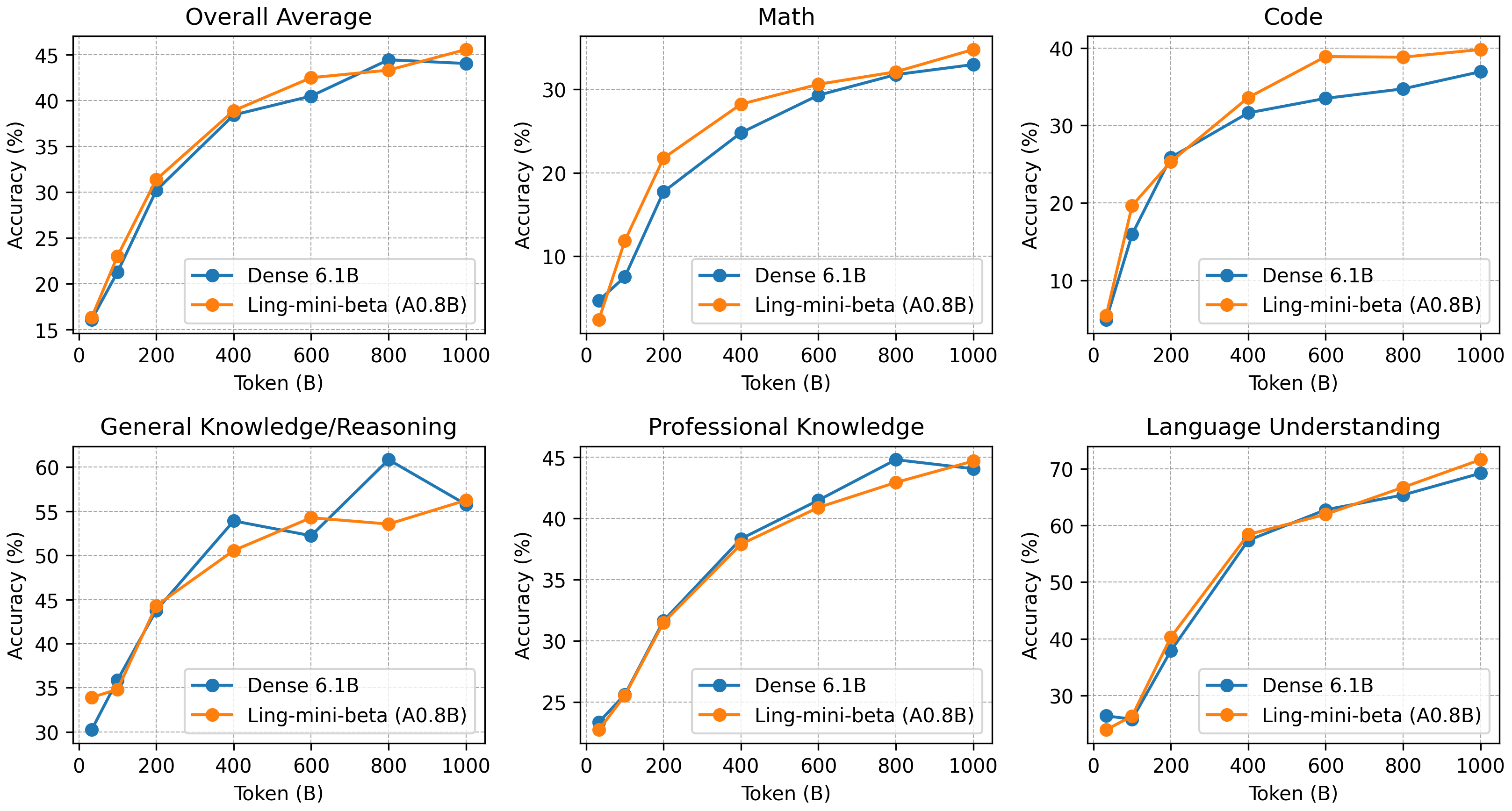}
    \caption{\textbf{Dynamic of Benchmarks.} 
    The comparison of the benchmarks changes between \mini and the \dense during training shows that their performances improved almost synchronously throughout the process, ultimately achieving similar final leaderboard results.}
    \label{fig:main}
\end{figure*}

\subsection{Evaluation}
\label{sec:mini_eval}

\paragraph{Evaluation Benchmarks}
To evaluate performance, we consider a diverse suite of downstream tasks designed to provide a holistic assessment of model capabilities. These tasks are grouped into several categories, such as: (a) General Knowledge/Reasoning (\eg ARC~\citep{arc}, AGIEval~\citep{agieval}, OpenBookQA~\citep{openbookqa}, BBH~\citep{bbh}, ProntoQA~\citep{PrOntoQA}, PIQA~\citep{piqa}, HellaSwag~\citep{hellaswag}, Multi-LogiEval~\citep{Multilogieval}) (b) Language Understanding (\eg RACE~\citep{race}) (c) Professional Knowledge (\eg MMLU~\citep{mmlu}, CMMLU~\citep{cmmlu}, MMLU-Pro~\citep{mmlu-pro}, GPQA~\citep{gpqa}, C-Eval~\citep{ceval}, CommonsenseQA~\citep{talmor2018commonsenseqa}) (d) Math (\eg GSM8K~\citep{gsm8k}, MATH~\citep{math}, GAOKAO~\citep{gaokao}, Gaokao2023-Math-En, MGSM~\citep{mgsm}, CMATH~\citep{cmath}, MathBench~\citep{mathbench}, Minerva-Math~\citep{Minerva}, CN-Middle School 24) (e) Code (\eg Humaneval~\citep{humaneval}, HumanEval-cn~\citep{humaneval_cn}, HumanEval-plus~\citep{evalplus}, HumanEval-FIM~\citep{fim}, LiveCodeBench~\citep{livecodebench}, MBPP~\citep{mbpp}, MBPP-Plus~\citep{evalplus}, CruxEval~\citep{cruxeval}).

\begin{table}[th!]
\caption{\textbf{Detailed performance comparison of \mini (17B-A0.8B) and \dense.}}
\small
  \label{table:eval}
  \centering
\begin{tabular}{clcc}
\toprule
\multicolumn{1}{l}{}                          & \textbf{Metric}                       & \textbf{\dense} & \textbf{\mini (A0.8B)} \\ \midrule
\multirow{10}{*}{\shortstack{General Knowledge\\ 
/Reasoning}} & ARC-challenge               & 59.7       & 57.0                   \\
                                              & ARC-easy                    & 78.0       & \textbf{78.7}                   \\
                                              & AGIEval                     & 33.4       & \textbf{34.9}                   \\
                                              & OpenBookQA                  & 68.6       & \textbf{75.2}                   \\ 
                                              & BBH                         & \textbf{48.0}       & 35.7                   \\
                                              & ProntoQA                    & 16.5       & \textbf{19.5}                   \\
                                              & Multi-LogiEval              & 55.6       & \textbf{61.3}                   \\
                                              & HellaSwag                   & 65.6       & \textbf{66.6}                   \\
                                              & PIQA                        & 76.6       & \textbf{77.2}                   \\ \cmidrule{2-4}
                                              & Average                     & 55.8       & \textbf{56.2}                   \\ \midrule
\multirow{7}{*}{\shortstack{Professional \\Knowledge}}       & MMLU                        & 51.1       & \textbf{53.1}                   \\ 
                                              & MMLU-Pro                    & 21.7       & \textbf{24.0 }                  \\
                                              & CMMLU                       & 50.7       & \textbf{51.9}                   \\
                                              & C-Eval                      & \textbf{52.5}       & 51.1                   \\
                                              & CommonsenseQA               & \textbf{63.6}       & 60.6                   \\
                                              & GPQA                        & 24.8       & \textbf{27.3}                   \\ \cmidrule{2-4}
                                              & Average                     & 44.0       & \textbf{44.7}                   \\ \midrule
\multirow{3}{*}{\shortstack{Language \\Understanding}}     & RACE-middle                 & 73.4       & \textbf{75.6}                   \\ 
                                              & RACE-high                   & 65.0       & \textbf{67.6}                   \\ \cmidrule{2-4}
                                              & Average                     & 69.2       & \textbf{71.6}                   \\ \midrule
\multirow{9}{*}{Code}                         & HumanEval                   & 31.7       & \textbf{35.4}                   \\ 
                                              & HumanEval-cn                & \textbf{34.2}       & 32.3                   \\
                                              & HumanEval-Plus              & 35.4       & \textbf{51.8}                  \\
                                              & HumanEval-FIM               & \textbf{62.8}       & 61.3                   \\
                                              & MBPP                        & 41.0       & \textbf{44.6}                   \\
                                              & MBPP-Plus                   & 50.0       & \textbf{51.6}                   \\
                                              & LiveCodeBench               & \textbf{7.5}        & 7.4                    \\
                                              & CruxEval                    & 32.9       & \textbf{34.1}                   \\ \cmidrule{2-4}
                                              & Average                     & 36.9       & \textbf{39.8}                   \\  \midrule
\multirow{10}{*}{Math}                        & GSM8K                       & \textbf{59.2}       & 58.0                   \\
                                              & MATH                        & 23.7       & \textbf{29.8 }                  \\
                                              & CMATH                       & 60.5       & \textbf{62.9}                   \\
                                              & MGSM-zh                     & 35.6       & \textbf{36.8}                   \\
                                              & CN-Middle School 24         & 41.6       & \textbf{42.6}                   \\
                                              & Minerva-Math                & \textbf{3.3}        & 2.9                    \\
                                              & MathBench                   & 27.5       & \textbf{28.6}                   \\
                                              & Gaokao2023-Math-En          & 33.1       & \textbf{33.5}                   \\
                                              & GAOKAO-Math24               & 12.1       & \textbf{17.6}                   \\ \cmidrule{2-4}
                                              & Average                     & 32.9       & \textbf{34.7}                   \\ \midrule
\multicolumn{2}{c}{{Overall Average}}     & 44.0       & \textbf{45.5}                   \\ \bottomrule
\end{tabular}
\end{table}

\paragraph{Evaluation Results}
The comparative evaluation in Table~\ref{table:eval} reveals that {\mini} achieves an average score of 45.5, surpassing \dense's 44.0. This result compellingly demonstrates that \mini accomplishes a "small yet powerful" feat with significantly lower inference costs, its activated parameters amount to only about \textit{13\%} of its competitor's, striking an exceptional balance between performance and efficiency.  

Upon closer examination of performance across specific dimensions, {\mini}'s advantages are both comprehensive and focused. 
In general knowledge and reasoning tasks, it exhibits notable advantages in open-ended question answering tasks such as OpenBookQA and complex logical reasoning benchmarks like Multi-LogiEval. This trend continues in specialized knowledge domains, where {{\mini}} delivers better results on comprehensive academic benchmarks like MMLU and MMLU-Pro. 
Its superiority is particularly evident in language understanding tasks, as it consistently outperforms its competitor in the RACE series of reading comprehension tests, showcasing stronger contextual understanding capabilities. 
In tasks requiring high coding proficiency, {\mini} stands out significantly, especially in the HumanEval-Plus benchmark, which measures code robustness, achieving an impressive lead of over \textit{16 points}. Similarly, in mathematical reasoning, while slightly lagging in basic arithmetic tasks like GSM8K, it excels in challenging benchmarks such as MATH and GAOKAO-Math24, demonstrating strong potential in solving complex problems. 
Collectively, {\mini} achieves a 1.5-point overall advantage, validating its parameter-efficient MoE design.  It not only drastically reduces inference costs through sparse activation but, more critically, its "expert networks" seem to enable higher performance ceilings in key areas such as language understanding, code generation, and advanced reasoning.

\begin{center}
\begin{tcolorbox}[
    colback=gray!10, 
    colframe=black!70, 
    width=0.95\textwidth, %
    title={Conclusion on \mini (17B-A0.8B)},
    fonttitle=\bfseries, %
    top=2mm, %
    bottom=2mm, %
    left=4mm, %
    right=4mm %
]
    Based on the scaling laws for efficiency leverage in Section~\ref{sec:scaling}, we design the \mini, a pilot model for the Ling-2.0 series, which has 17.5 B total parameters but only active 0.8 B parameters. Experimental results demonstrate that \mini achieves over a 7× efficiency leverage while maintaining comparable performance  to dense models with 6.1B, more than 7× the number of active parameters.
\end{tcolorbox}
\end{center}

%% file: sections/6-discussion.tex
\section{Discussion and Limitations}

\subsection{Comparison with Previous Works.}
\label{sec:dis-dif}

\paragraph{Comparison with \cite{clark2022unified}.}
In their study, \cite{clark2022unified} used a fixed dataset and concluded that the efficiency of MoE models over dense models diminishes beyond a certain scale. Contrary to the findings of them, our results in Figure~\ref{fig:active} demonstrate that MoE models are consistently more compute-efficient than their dense counterparts across all evaluated model sizes. 
The apparent contradiction can be reconciled by examining the experimental design. Our preliminary studies reveal that for a fixed compute budget, the optimal resource allocation for MoE and dense models differs significantly: MoE models favor fewer parameters and more training tokens, as shown in Section~\ref{sec:scaling-allocation}. Consequently, evaluating all models on a fixed dataset can lead to an unfair comparison, where MoE models are likely under-trained relative to dense models, yielding potentially misleading conclusions. This hypothesis is further corroborated by the convergence dynamics in Figure~\ref{fig:mini-loss-all}, which show that MoE models, despite a slower start, eventually outperform dense models as training progresses, which also be verified by \cite{ludziejewski2024scaling}.
Therefore, diverging from prior work, our experiments is guided by the scaling laws for optimal compute-allocation. We dynamically scale the number of training tokens with compute budget, ensuring that experimental models achieve a comparable and sufficient degree of training. This approach ensures the fairness and reliability of our comparison.

\paragraph{Comparison with \cite{ludziejewski2024scaling}.} 
Our findings regarding the impact of expert granularity differ from those of \cite{ludziejewski2024scaling} in two main aspects. First, we observe a log-polynomial relationship between performance and granularity, indicating an optimal granularity range, whereas \cite{ludziejewski2024scaling} reported a monotonic trend where finer granularity consistently reduces loss. Second, our experiments show that the EL of the MoE is usually within a 10x factor at the tested compute scales—significantly lower than their >10× ``Relative FLOPs to train equivalent Transformer'' (Figure 1(b) in their paper).
We attribute these differences to three key variations in experimental setups:
(1) Granularity definition: While \cite{ludziejewski2024scaling} uses $G = d_{\text{ffn}}/d_{\text{expert}} = 4d_{\text{model}}/d_{\text{expert}}$, our experiments, aligned with leading models~\citep{deepseekai2024deepseekv3technicalreport,kimiK2}, adopts stricter $G = 2d_{\text{model}}/d_{\text{expert}}$. As a result, our experts are effectively half-sized at equivalent nominal granularity (\eg $  G = 16  $), enabling exploration of finer actual granularity.
(2) Hyperparameter strategies: Unlike their use of uniform hyperparameters across experiments, we optimize these settings for each compute budget, as our preliminary studies in Section~\ref{sec:scaling-hyper} confirm that optimal configurations vary significantly with compute scale. This avoids inequitable comparisons that may arise from a fixed hyperparameter set.
(3) Base MoE architectures: Our experiments utilize a base MoE architecture with a 1/32 routable expert activation ratio, whereas their architecture employs a sparser 1/64 ratio. This sparser activation inherently provides higher baseline efficiency, potentially amplifying its measured advantage. 
In summary, our distinct conclusions arise from investigating a finer granularity spectrum under a different definition and ensuring appropriate training conditions for all models.

\paragraph{Comparison with \cite{abnar2025parameters}.}
Our findings on the optimal activation ratio align with those of \cite{abnar2025parameters}, confirming that under a fixed compute budget, larger and sparser models yield better performance. However, our research extends beyond this conclusion in both methodology and scope.
However, our research substantially extends this direction. First, we determine training hyperparameters through extensive preliminary experiments. Second, we systematically investigate how architectural factors—particularly expert granularity and shared expert ratios—affect model performance. This reveals that beyond the primary activation ratio trend, expert granularity introduces log-polynomial adjustments to performance.
Ultimately, our primary contribution is the direct derivation of scaling laws for the efficiency leverage of MoE models relative to their dense counterparts, rather than conventional scaling laws for loss. The key advantage is its independence from specific training datasets. It directly establishes a quantitative relationship between MoE architectural configurations and their relative performance efficiency, offering more generalizable and actionable principles for model design.

\paragraph{Comparison with \cite{ludziejewski2025joint}.}
Our research and the work of \cite{ludziejewski2025joint} are complementary, with each study addressing a distinct facet of the scaling laws for MoE models. 
Our work addresses the question: given a fixed compute budget and a specific model scale (\ie FLOPs per token), how should one configure the architectural parameters (\ie expert granularity, activation ratio) to maximize performance? In contrast, their study concentrates on a different optimization problem: under the dual constraints of a compute budget and memory limitations, what is the optimal allocation of resources between model scale and data size? 
While our preliminary experiments did touch upon the model-data allocation for MoE models, this exploration was intentionally limited. It was conducted under a single compute budget constraint and for a specific MoE architecture. Its primary purpose was not to derive a comprehensive allocation strategy, but rather to establish the fundamental differences in optimal resource allocation between MoE and dense models. This foundational understanding was crucial for our main experiments, as it enabled us to provision a sufficient training budget to ensure all models were compared fairly under conditions of adequate, near-optimal training.

\subsection{Limitations}
Consistent with standard practice in scaling law research \citep{kaplan2020scaling, hoffmann2022training,clark2022unified,ludziejewski2024scaling,abnar2025parameters}, our analysis quantifies computational cost exclusively in terms of theoretical FLOPs. While FLOPs provide a valuable, hardware-agnostic metric for comparing model architectures, we acknowledge that this abstraction does not capture the full spectrum of real-world costs. Factors such as hardware specifications, system infrastructure, and implementation details can introduce discrepancies between theoretical FLOPs and actual wall-clock time. 
Furthermore, due to significant resource constraints, our methodology relies on the simplifying assumption that the effects of different MoE architectural factors are largely independent. Based on this premise, we conducted a series of individual ablation studies to quantify the impact of each factor in isolation, subsequently synthesizing these effects into a unified scaling law.
We acknowledge that a primary limitation of this approach is its potential to overlook interaction effects between architectural components. Nevertheless,it remains the most pragmatic and feasible pathway within the scope of our available resources.
Despite these limitations, our findings underscore the immense potential of MoE models. By enabling a massive increase in model capacity with a minimal increase in per-token computation, they offer a clear path toward improving both model performance and efficiency.

%% file: sections/7-related.tex
\section{Related Work}

\subsection{Scaling Laws for Language Models}
Scaling laws provide a framework for understanding and predicting the performance of language models under varying conditions. \cite{kaplan2020scaling} laid the foundation by demonstrating that model performance adheres to predictable power-law relationships involving model size, dataset size, and compute budget. Building on this, \cite{hoffmann2022training} introduced the Chinchilla scaling laws, highlighting the importance of balancing model size and training data volume for compute-optimal training. They showed that scaling model size without a corresponding increase in data leads to diminishing performance gains. \cite{sardana2023beyond} advanced this understanding by incorporating inference costs into compute-optimal frameworks, proposing strategies for optimizing performance under fixed inference constraints. Additionally, \cite{bi2024deepseek} emphasized the critical role of data quality, demonstrating that higher-quality datasets enable more efficient scaling, particularly with larger models.
Recent advancements have applied these scaling laws to various specialized areas. For example, hyperparameter optimization has been explored in the context of scaling laws~\citep{bi2024deepseek,li2025predictable}, while \cite{gadre2024language} investigated the phenomena of over-training and its implications on model performance. Furthermore, scaling laws have been analyzed for their impact on downstream task performance across a range of applications~\citep{chen2024scaling,ruan2024observational,isik2025scaling,hu2023predicting,grattafiori2024llama,li2025predictable}, underscoring their adaptability and relevance in addressing both theoretical and practical challenges in language modeling. 

\subsection{Scaling Laws for Mixture-of-Experts (MoE)}
Mixture-of-Experts (MoE) models~\citep{shazeer2017outrageously,lepikhin2020gshard} have emerged as a powerful architecture for language modeling, primarily due to their ability to decouple computational cost from parameter count. Recent research has further explored optimizations within the MoE paradigm. For instance, DeepSeekMoE~\citep{dai2024deepseekmoe} investigated the impact of fine-grained expert settings on model performance, proposing a novel design that incorporates shared experts and a hybrid structure combining dense layers with MoE layers. Complementing this, \cite{zoph2022st} highlighted that the performance gains from increased sparsity diminish significantly once the number of experts exceeds 256, suggesting a practical limit for highly sparse models.
With the widespread adoption of the MoE architecture, the scaling laws governing MoE models have been extensively studied. Early work by \cite{clark2022unified} examined scaling by varying model size and the number of experts on a fixed dataset, concluding that routed models offer efficiency advantages only up to a certain scale. This analysis was subsequently extended by \cite{ludziejewski2024scaling}, who incorporated variable dataset sizes and explored the effects of expert granularity. Additionally, \cite{wang2024scaling} investigated the transferability and discrepancies of scaling laws between dense models and MoE models. \cite{abnar2025parameters} advanced this line of inquiry by deriving scaling laws for optimal sparsity, explicitly considering the interplay between training FLOPs and model size. They also analyzed the relationship between pretraining loss and downstream task performance, noting distinct behaviors between MoE and dense models on certain tasks. More recently, \cite{ludziejewski2025joint} derived joint scaling laws applicable to both dense Transformers and MoE models, demonstrating that MoE architectures can outperform dense counterparts even under constraints of memory usage or total parameter count.

%% file: sections/8-conclusion.tex
\section{Conclusion}

In this work, we introduced Efficiency Leverage (EL), a metric that measures the computational advantage of an MoE model relative to a dense counterpart, to quantify the scaling behavior of MoE performance with architectural factors. 
Our large-scale empirical study, based on over 300 trained models, systematically deconstructed the relationship between MoE design choices and EL. We found that the efficiency of an MoE architecture is governed by a set of predictable principles. Specifically, EL scales as a power law with both the activation ratio and the total compute budget, while expert granularity acts as a non-linear modulator with a stable optimal range. Other factors, such as shared experts, were found to have only a secondary impact. 
We distilled these insights into a unified scaling law that accurately predicts the EL of any MoE configuration. The predictive power of our framework was empirically validated through the successful design and training of a 17.5B parameter MoE model, which, as predicted, achieved an efficiency leverage of over 7x compared to its dense equivalent. 

For future work, our framework can be extended in several key directions: (1) Incorporating memory constraints and communication overhead into the EL framework, particularly for distributed training scenarios where these factors dominate practical efficiency. (2) Developing a unified metric that balances training compute leverage with inference latency requirements, enabling end-to-end efficient architecture co-design.
We hope this work inspires continued innovation in MoE architectures, ultimately propelling the field {toward greater leverage}.

%% file: sections/9-appendix.tex
\appendix
\section{Notation}
To aid readability, we provide a list of key symbols used throughout this paper.

\begin{table}[h]
  \caption{Notation.}
  \small
  \label{tab:notation}
  \centering
  \begin{tabular}{lll}
    \toprule
    Symbol     & Description       \\
    \midrule
    $E$     & Number of routable experts.      \\
    $E_a$   & Number of activated experts.      \\
    $E_s$   & Number of shared experts.      \\
    $N$     & Number of non-vocabulary parameters.      \\
    $N_a$   & Number of activated parameters.      \\
    $d_{model}$    & Model hidden dimension.      \\
    $d_{expert}$   & Expert hidden dimension.      \\
    $C$     & Total training compute in FLOPs    \\
    $M$     & Compute (w/o embedding) per token in FLOPs.    \\
    $D$     & Dataset size in tokens. \\
    $A$     & Activation ratio, \ie $(E_a + E_s)/(E+E_s)$.  \\
    $G$     & Granularity of experts, \ie $2d_{mode;}/d_{expert}$    \\
    $S$     & Shared expert ratio, \ie $E_s/(E_a+E_s)$      \\
    \bottomrule
  \end{tabular}
\end{table}

\section{Experimental Setup}
\label{app:setup}

Our experiments primarily follow the architecture and training configurations of the Ling series models~\citep{team2025every}.

\paragraph{Architecture and Tokenizer}
We adopt a Grouped Query Attention (GQA)~\citep{ainslie2023gqa} architecture based on the standard decoder-only Transformer, consisting of an embedding layer, multiple alternating layers of attention mechanisms and feed-forward networks, and a final de-embedding layer. Additionally, we use the BPE (Byte-Pair Encoding) algorithm~\citep{sennrich2015neural} and RoPE (Rotary Positional Embedding)~\citep{su2024roformer} to handle positional information. The vocabulary size is 126,464, and the sequence length is 4,096.

\paragraph{Expert Routing Strategy} 
In our MoE layers, a routing network assigns each token's hidden state $h_t$ to the top-$N_a$ experts. This is achieved by generating gating scores $g_t = \text{Softmax}(W_g \cdot h_t)$, where $W_g$ is a learnable matrix. The final output is a weighted sum of the selected experts' outputs: $o_t = \sum_{i \in \text{TopK}(g_t)} g_{t,i} \cdot E_i(h_t)$, where $E_i$ is the $i$-th expert in total $N$ experts. To ensure balanced expert utilization and stable training, we incorporate two standard auxiliary losses: a load balancing loss~\citep{lepikhin2020gshard} (coefficient of 0.01) to encourage uniform token distribution, and a router z-loss~\citep{zoph2022st} (coefficient of 0.001) to regularize the magnitude of the gating logits.

\paragraph{Optimizer and Scheduler}
The parameters of experimental models are initialized from a distribution with a standard deviation of $0.006$ and optimized using the AdamW optimizer~\citep{adamw}. The optimizer's hyperparameters are set to $\beta_1 = 0.9$ and $\beta_2 = 0.95$, with 0.1 weight decay applied. The learning rate schedule employs a WSD (warmup-stable-decay) strategy~\citep{hu2024minicpm}: the first $1\%$ of training steps use linear warm-up, followed by exponential decay that reduces the learning rate to $10\%$ of its peak value.

\paragraph{Pre-training Data}
The training data is sourced from a large-scale multilingual corpus created by the Ling Team, primarily covering English and Chinese, while also including various other languages. This corpus encompasses web text, mathematical materials, programming scripts, published literature, and diverse textual content. To validate model performance, we extracted a 2T-token subset from this corpus for training.
In Table~\ref{table:training-data}, we present the composition of the training datasets for all experiments. Unless otherwise specified, this configuration is used throughout.

\begin{table}[h]
\small
  \caption{Pre-training data composition.}
  \label{table:training-data}
  \centering
  \setlength{\tabcolsep}{5pt} %
  \begin{tabular}{cccccccccccc}
    \toprule
    \textbf{Type} & Web & Books & Wiki & Academic & Code & News & Social & Domain & SFT & Math & Exam \\
    \midrule
    \textbf{Ratio} & 46.0\% & 5.0\% & 4.0\% & 6.0\% & 25.0\% & 0.1\% & 1.9\% & 1.0\% & 4.0\% & 6.0\% & 1.0\% \\
    \bottomrule
  \end{tabular}
\end{table}

\section{Estimating FLOPs}
\label{app:flops}

To analyze the efficiency of our models, we quantify the computational cost in terms of total training Floating Point Operations (FLOPs). Following standard practice \citep{kaplan2020scaling}, we estimate the total training FLOPs as approximately three times the cost of a single forward pass ($C_{\text{train}} \approx 3 \cdot C_{\text{fwd}}$). The forward pass FLOPs are the sum of computations from the attention and feed-forward network (FFN) layers, plus a final logit projection.

For a model with hidden size $d_{\text{model}}$, batch size $B$, and sequence length $s$, the cost of the attention block per layer, $C_{\text{attn}}$, which includes Grouped-Query Attention (GQA)~\citep{ainslie2023gqa} and all projections, is approximately:
\begin{equation}
C_{\text{attn}} \approx B s d_{\text{model}}^2 \left(2 + \frac{2}{n_h / n_{kv}}\right) + 4 B s^2 d_{\text{model}}
\end{equation}
where $n_h$ and $n_{kv}$ are the number of attention and key-value heads, respectively.
The FFN cost varies by layer type. A dense layer with intermediate size $d_{\text{ffn}}$ requires $C_{\text{dense\_ffn}} = 6 B s d_{\text{model}} d_{\text{ffn}}$ FLOPs. A MoE layer activating $E_a$ experts, each with size $d_{\text{expert}}$, requires:
\begin{equation}
C_{\text{moe\_ffn}} \approx 6 B s d_{\text{model}} (E_a \cdot d_{\text{expert}})
\end{equation}
If a shared expert of size $d_{\text{shared}}$ is used, its cost, $6 B s d_{\text{model}} d_{\text{shared}}$, is added.
For a model with $L$ layers (of which the first $L_{\text{dense}}$ are dense) and a vocabulary of size $V$, the total forward FLOPs are:
\begin{equation}
C_{\text{fwd}} = \sum_{i=1}^{L} (C_{\text{attn}} + C_{\text{ffn}, i}) + 2 B s d_{\text{model}} V
\end{equation}
where $C_{\text{ffn}, i}$ is the FFN cost for the $i$-th layer, which can be either $C_{\text{dense\_ffn}}$ or $C_{\text{moe\_ffn}}$.

\section{Additional Experiments}
\label{app:add_exp}

\begin{figure}[htbp]
    \centering
    \begin{subfigure}[b]{0.382\textwidth} 
   \includegraphics[width=\textwidth]{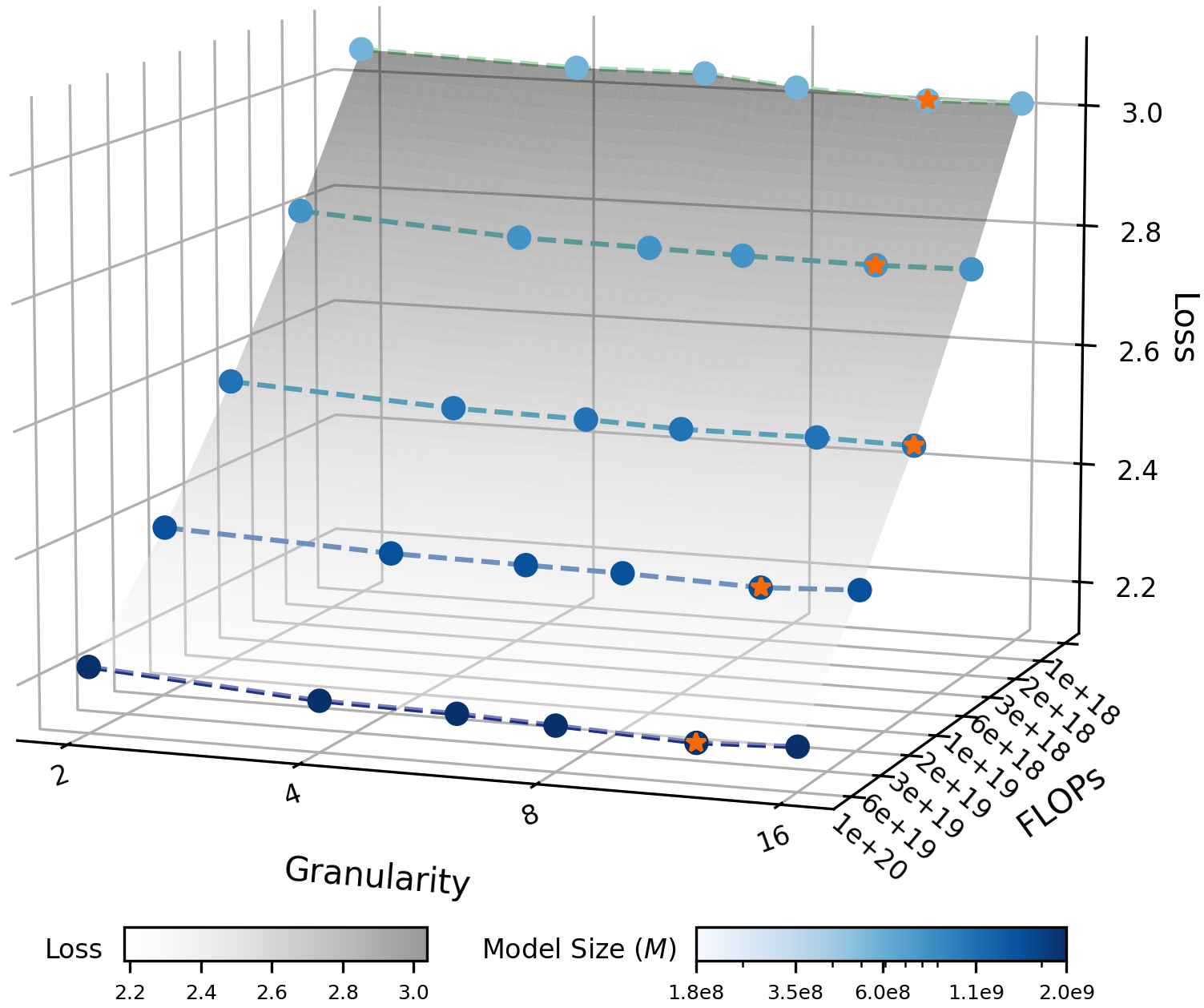}
   \caption{IsoFLOPs curves over different granularity.}
   \label{fig:router}
    \end{subfigure}
    \hfill 
    \begin{subfigure}[b]{0.61\textwidth} 
   \includegraphics[width=\textwidth]{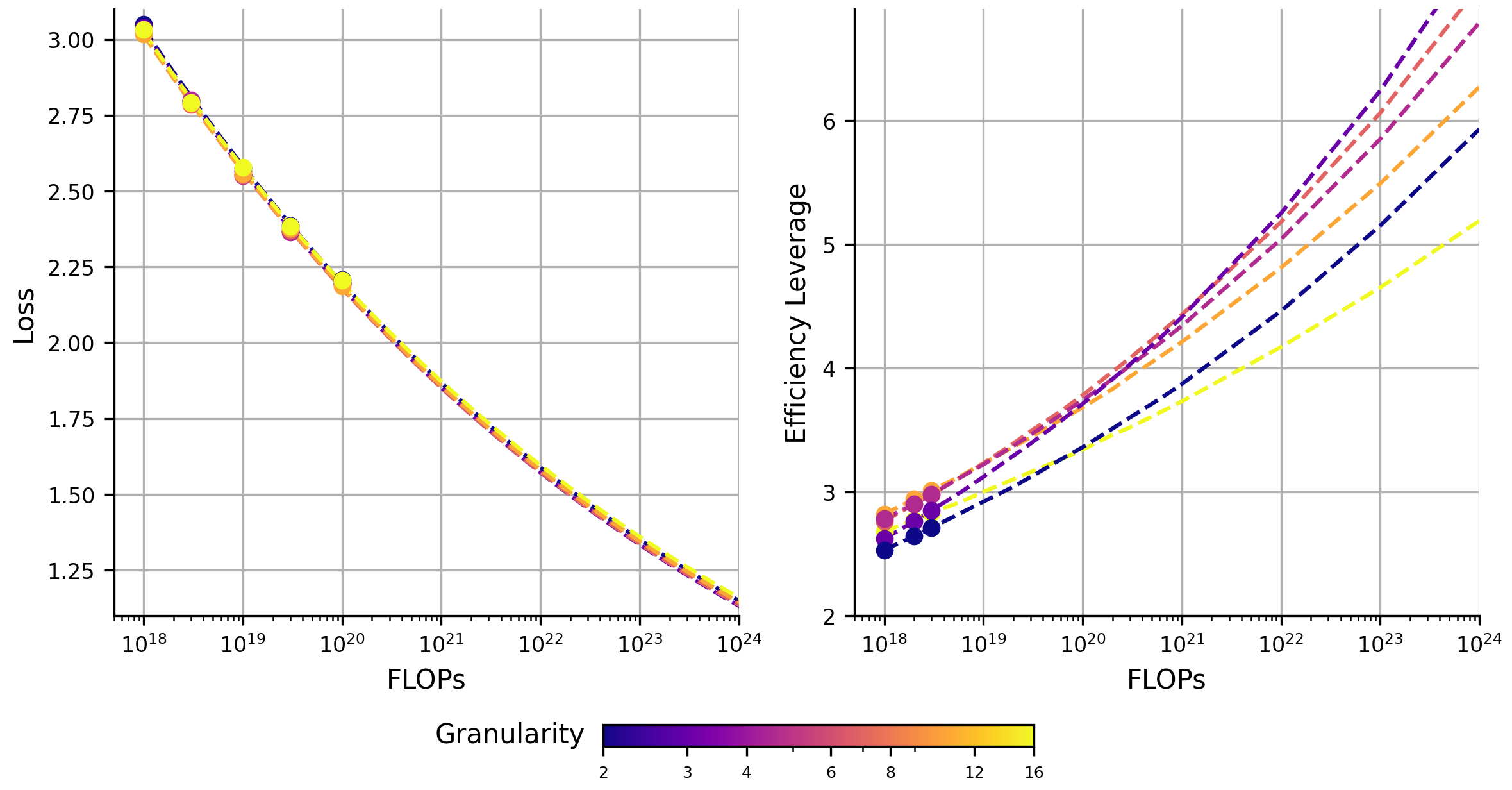}
   \caption{Loss scaling curves under over different granularity.}
   \label{fig:router-loss}
    \end{subfigure}
    \caption{\textbf{Impact of Expert Granularity on Loss Under Weakened Routing Balance. } 
    }
    \label{fig:router-all}
\end{figure}

\paragraph{The Impact of Routing Balance on the Optimal Expert Granularity.}
To investigate how routing quality influences the optimal expert granularity, we induce a state of routing imbalance. This is achieved by setting the coefficient of load balancing loss to 0.001, a setup known to cause load imbalance. In this setting, we train MoE models with a varying expert granularity while maintaining a constant total parameter count. 
As shown in Figure~\ref{fig:router-all}, our results reveal that a coarser expert granularity becomes optimal under such imbalanced routing. Specifically, compared with the results in Section~\ref{sec:granularity}, the IsoFLOPs curves (Figure~\ref{fig:router}) demonstrate that models with coarser granularity ($G=6, 8$) achieve lower loss for a given computational budget. This trend is consistently observed in the loss scaling curves (Figure~\ref{fig:router-loss}). 
This phenomenon indicates that when the routing mechanism becomes a performance bottleneck, a fine-grained architecture with numerous specialized experts is counterproductive. The weakened router cannot distribute tokens effectively, nullifying the benefits of specialization. Consequently, the model benefits more from a coarser-grained design with fewer, more generalized experts, as this simplifies the routing task and mitigates the detrimental effects of the load imbalance.

\begin{figure}[htbp]
    \centering
    \begin{subfigure}[b]{0.382\textwidth} 
        \includegraphics[width=\textwidth]{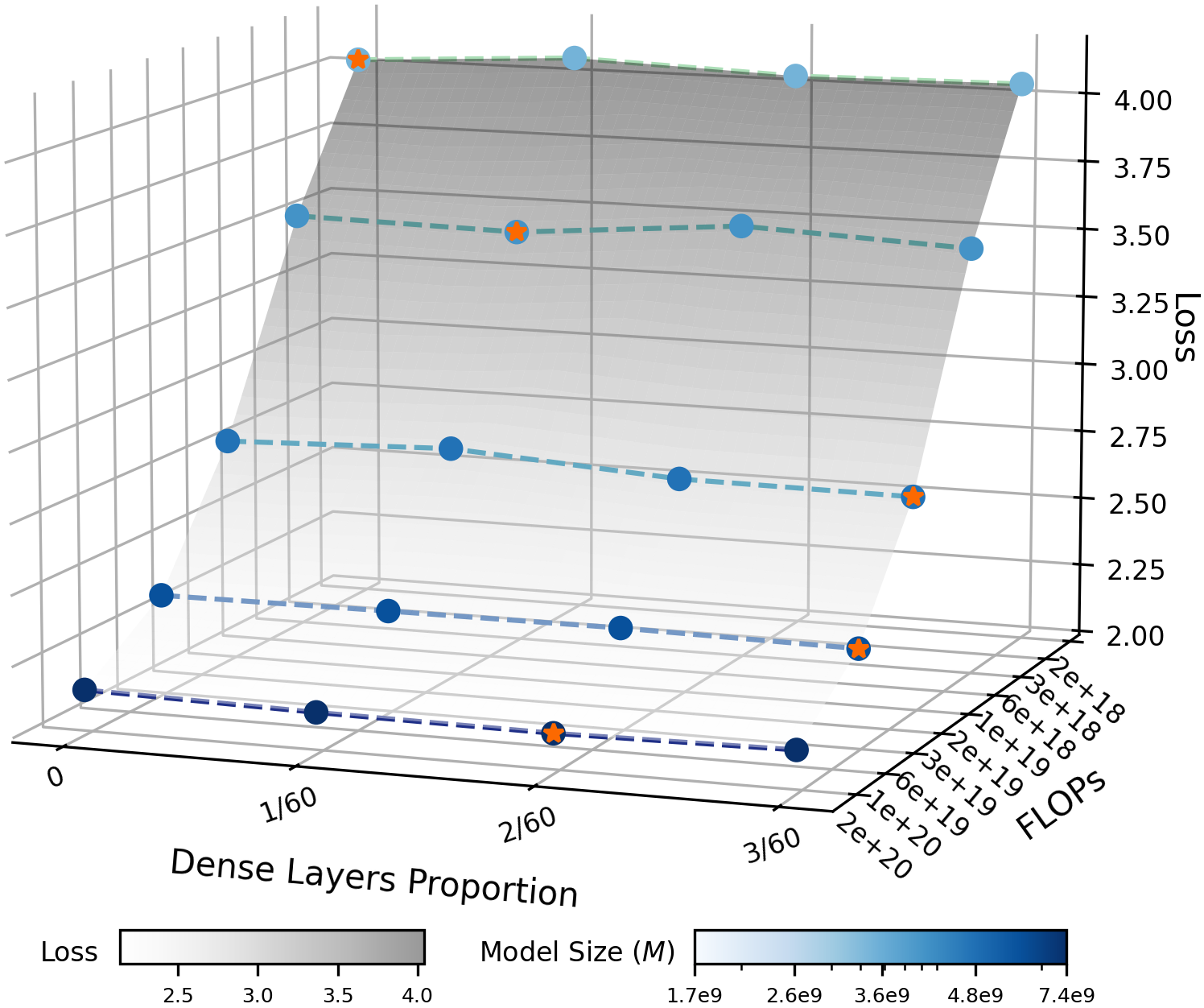}
        \caption{}
        \label{fig:dense}
    \end{subfigure}
    \hfill 
    \begin{subfigure}[b]{0.61\textwidth} 
        \includegraphics[width=\textwidth]{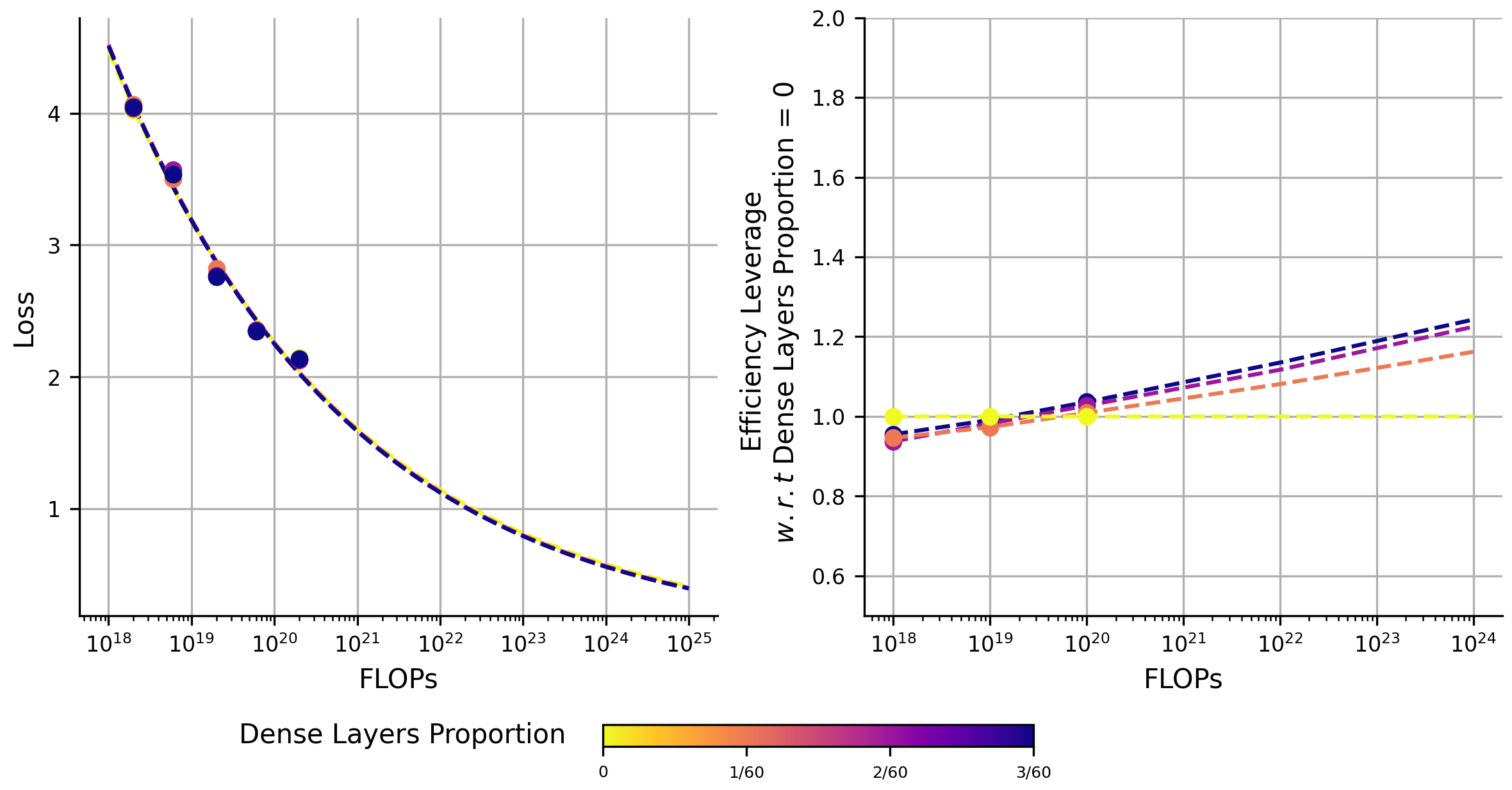}
        \caption{}
        \label{fig:dense-loss}
    \end{subfigure}
    \hfill 
    \begin{subfigure}[b]{0.382\textwidth} 
        \includegraphics[width=\textwidth]{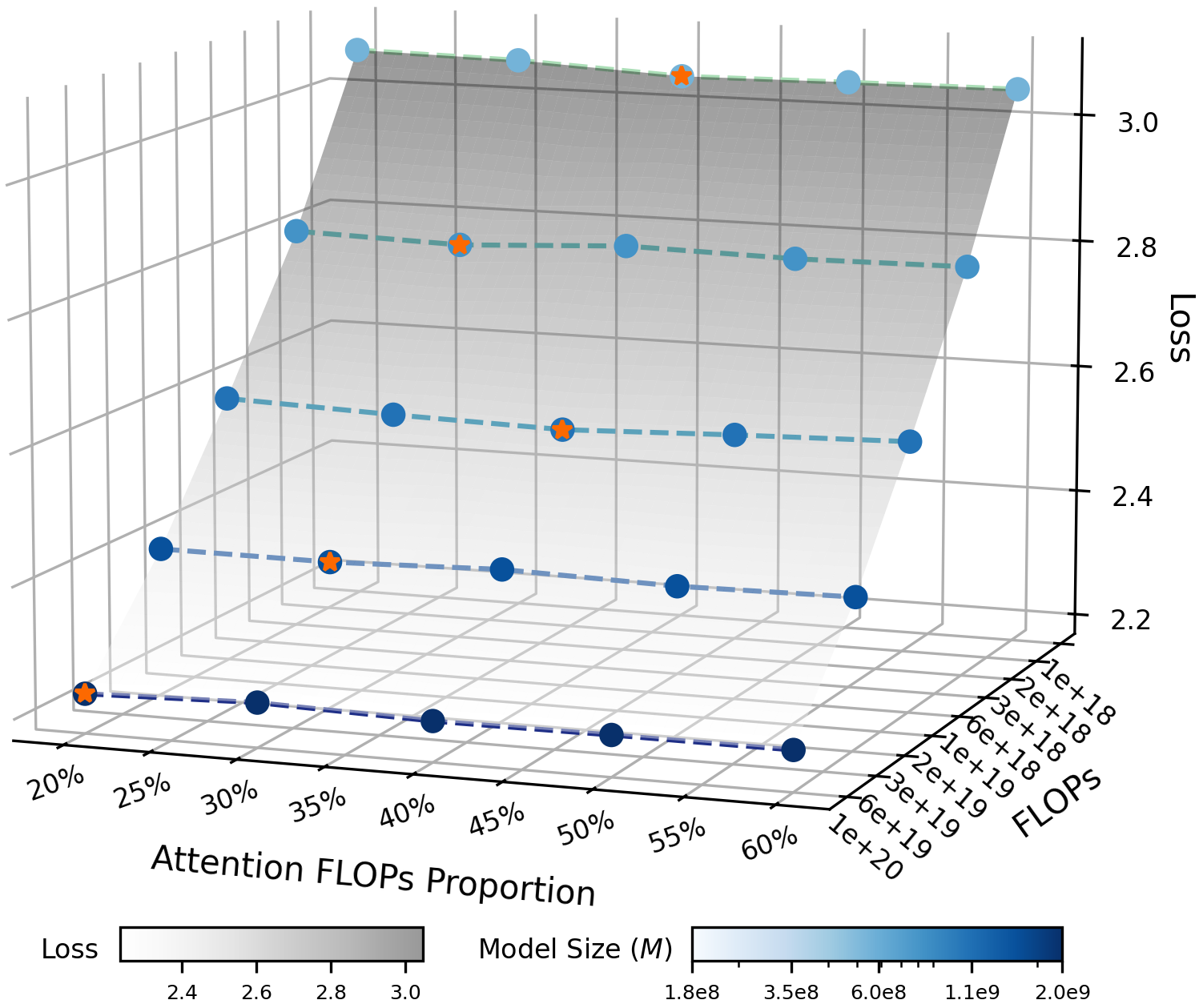}
        \caption{}
        \label{fig:att}
    \end{subfigure}
    \hfill 
    \begin{subfigure}[b]{0.61\textwidth} 
        \includegraphics[width=\textwidth]{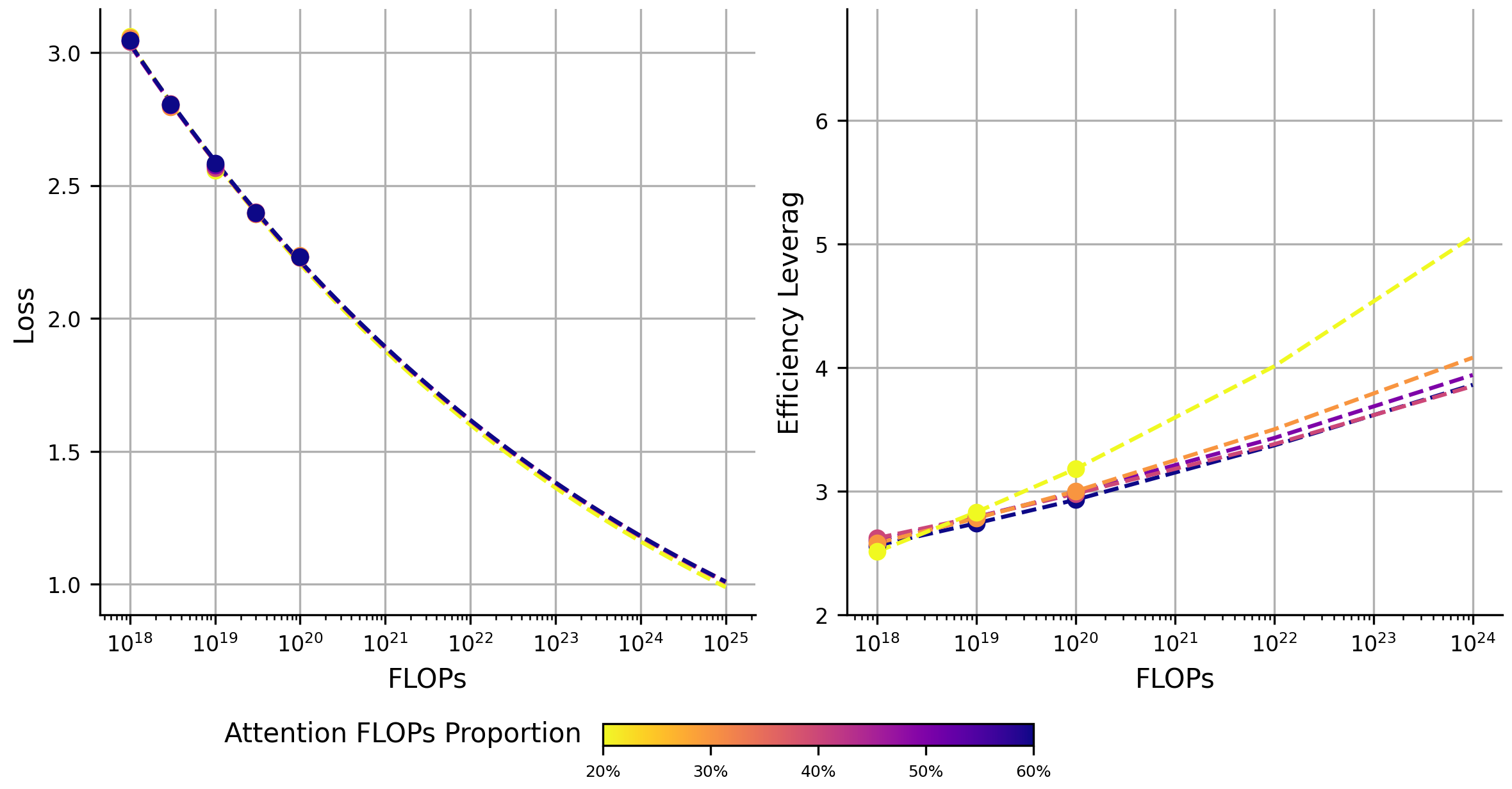}
        \caption{}
        \label{fig:att-loss}
    \end{subfigure}
    \caption{\textbf{Impact of Dense Layers Proportion and Compute Budget Allocation between Attention and FFN.} 
    (a,b) Replacing the first few layers with dense layers shows minor impact on model performance. As computational budgets increase, the optimal proportion of dense layers also gradually rises. 
    (c,d) Modifying the attention FLOPs ratio within a broad range (20\%-50\%) has a negligible influence on model performance, demonstrating the robustness of this configuration.
    }
    \label{fig:dense-att}
\end{figure}

\paragraph{Arrangement of MoE and Dense Layers}
To ensure balanced routing in the early layers, mainstream MoE models typically replace all FFNs except for the first few layers with MoE layers. We investigate the impact of this design decision on the efficiency of MoE models. To ensure a meaningful exploration space, we extend all models in our experiments to 60 layers and set the first 1, 2, or 3 layers as dense layers sequentially. The dimension of these dense layers is set to match the total dimension of the activated experts in the corresponding MoE layers, ensuring the overall computational cost (FLOPs/token) remains constant. This design allows us to isolate and study the effect of the proportion of dense layers on MoE efficiency. The experimental results, presented in Figure~\ref{fig:dense} and \ref{fig:dense-loss}, reveal the following key findings:
1) From a model performance perspective, replacing the first few layers with dense layers has a minor impact. Using a dense proportion of zero as the baseline, we estimated the efficiency leverage for each configuration. Within a FLOPs budget of up to $1 \times 10^{24}$ FLOPs, the efficiency leverage remains close to 1. This indicates that configuring the initial layers as dense offers negligible efficiency improvement. However, this adjustment effectively reduces the total number of parameters in the model and mitigates routing imbalances in the early layers. Thus, despite its limited efficiency gains, this remains a valuable design optimization.
2) Further investigation into the optimal proportion of dense layers under varying computational budgets reveals a trend: as FLOPs budgets increase, the optimal dense proportion also grows. For example, in our experiments, when the compute budget is $1 \times 10^{18}$ FLOPs, the optimal dense proportion is zero. As the compute budget increases to $3 \times 10^{20}$ FLOPs, the optimal dense layer proportion shifts to approximately $2/60$ or $3/60$.

\paragraph{Compute Resource Allocation between Attention and FFN}
As two core components of the Transformer model, the attention mechanism (Attention) and FFN account for the majority of the model's computational load. To this end, we explore the impact of computational allocation between the attention mechanism and the FFN on the efficiency of the MoE model. Specifically, we construct a series of models with fixed model scale $M$ but varying compute budgets by increasing the hidden layer size of the attention module while reducing the hidden layer size of each expert in the MoE. We then observe the performance changes of these models under different computational allocations and evaluate their scaling trends. The experimental results are illustrated in Figure~\ref{fig:att} and \ref{fig:att-loss}, revealing the following key findings:
1)  When the attention FLOPs ratio is between 30\% and 40\%, it represents a relatively stable and reliable configuration. Models tend to achieve optimal or near-optimal performance within this range. This configuration is consistent with the default settings of mainstream open-source MoE models.
2) Adjusting the attention FLOPs ratio within a broader range (20\%-50\%) has minor impact on model performance. As shown in Figure~\ref{fig:att-loss}, the loss scaling curves and efficiency leverage of these models are nearly identical. Since the attention mechanism generally has a higher computational density (\ie FLOPs-per-parameter) compared to the FFN, increasing the attention FLOPs ratio while keeping the overall model size constant reduces the total number of model parameters, resulting in higher knowledge density. However, this also implies potentially higher downstream inference costs.

\section{Additional Evaluation Results of Ling-mini-beta}
\begin{figure*}[htbp]
    \centering
    \includegraphics[width=\textwidth]{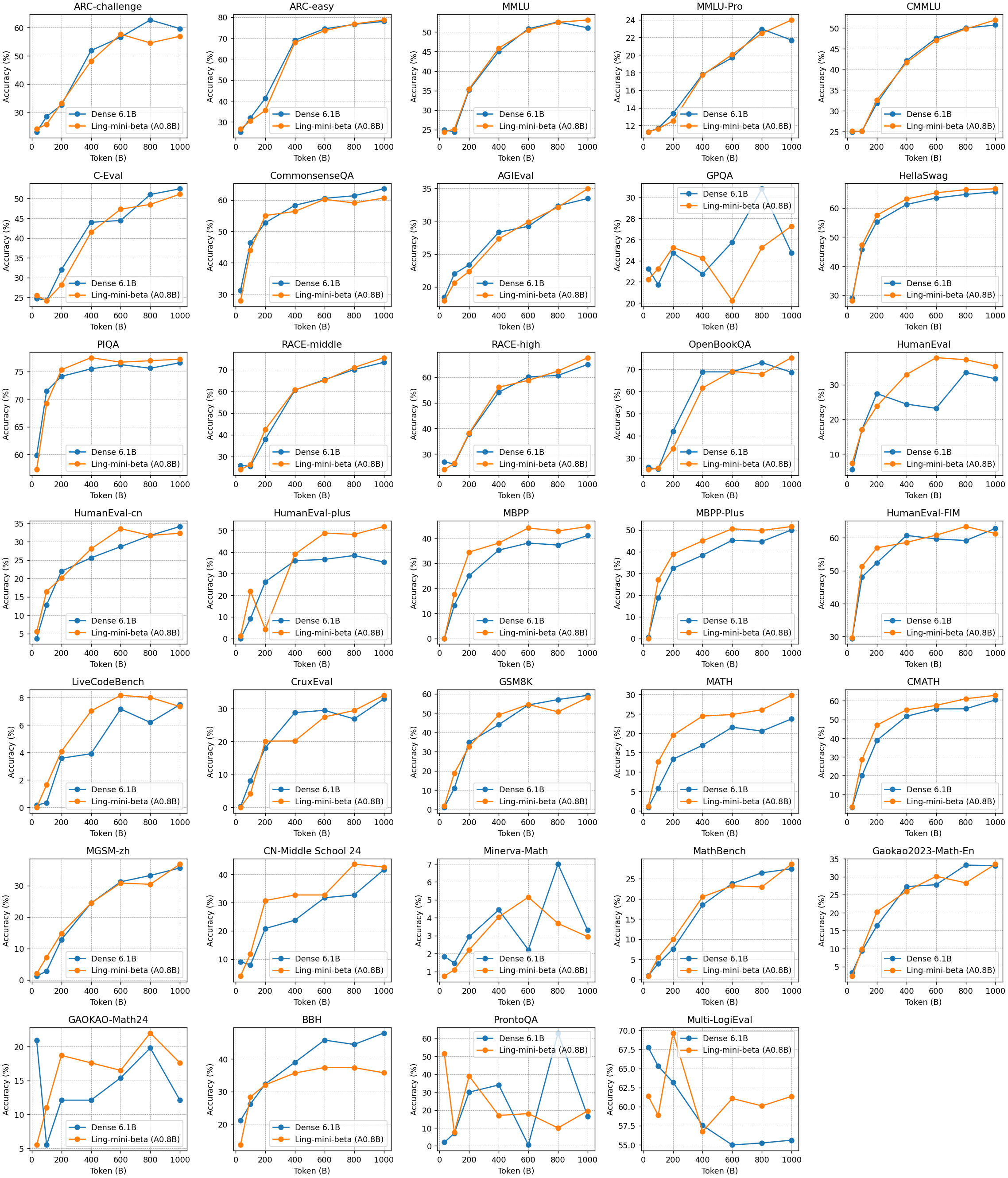}
    \caption{Overall and category-wise performance comparison between \mini (17B-A0.8B) and \dense. }
    \label{fig:category-eval-mini}
\end{figure*}

We present a detailed evaluation of \mini's training process. Figure \ref{fig:category-eval-mini} provides a comprehensive comparison across datasets and categories, as outlined in the main experiments in Section \ref{sec:mini_eval}. The results show that \mini achieves comparable performance to \dense on the majority of datasets.

\section{List of Experimental Models}
\label{app:model_list}

The detailed configurations for all experiments conducted in this study are presented in Tables~\ref{table:activation_ratio_configs} (activation ratio), Tables~\ref{table:granularity_configs} (expert granularity), Tables~\ref{table:shared_expert_configs} (shared experts), Tables~\ref{table:layer_arrangement_configs} (layer arrangement), and Tables~\ref{table:attention_ffn_configs} (compute allocation between attention and FFNs).

\begin{table}[h]
\small
  \caption{Experimental configurations for the expert activation ratio analysis.  Within each group, the number of activated experts ($E_a=2$) is fixed, while the total number of experts ($E$) is varied to study the effect of the activation ratio.}
  \label{table:activation_ratio_configs}
  \centering
  \setlength{\tabcolsep}{5pt} %
  \begin{tabular}{ccccccccccc}
    \toprule
    $n_{layers}$ & $d_{model}$ & $d_{expert}$ & $n_{heads}$ & $n_{kv\_head}$ & $E$                          & $E_a$ & $E_s$ & $\eta$ & $B$ & Max training FLOPs \\
    \midrule
    8           & 384         & 320          & 8           & 2              & {[}2,4,8,16,32,64,128,256{]} & 2     & 1     & 1.52e-3 & 98     & 2e18               \\
8           & 512         & 512          & 8           & 2              & {[}2,4,8,16,32,64,128,256{]} & 2     & 1     & 1.31e-3 & 147    & 6e18               \\
10          & 640         & 640          & 10          & 2              & {[}2,4,8,16,32,64,128,256{]} & 2     & 1     & 1.11e-3 & 228    & 2e19               \\
14          & 768         & 768          & 12          & 4              & {[}2,4,8,16,32,64,128,256{]} & 2     & 1     & 9.5e-4 & 342    & 6e19               \\
16          & 1024        & 1024         & 16          & 4              & {[}2,4,8,16,32,64,128,256{]} & 2     & 1     & 8.1e-4 & 531    & 2e20               \\
22          & 1280        & 1280         & 20          & 4              & {[}2,4,8,16,32,64,128,256{]} & 2     & 1     & 7.0e-4 & 795    & 6e20  \\
    \bottomrule
  \end{tabular}
\end{table}

\begin{table}[h!]
  \centering
  \caption{Experimental configurations for the expert granularity analysis. Within each group, the base model architecture is fixed while the MoE configuration (total experts $E$, activated experts $E_a$, shared experts $E_s$, and expert dimension $d_{\text{expert}}$) is varied to study the effect of granularity.}
  \label{table:granularity_configs}
  \small %
  \setlength{\tabcolsep}{4pt} %
  \begin{tabular}{cccccccccc}
    \toprule
    $n_{layers}$ & $d_{model}$ & {$n_{\text{heads}}$} & {$E$} & {$E_a$} & {$E_s$} & {$d_{\text{expert}}$} & {$B$} & {$\eta$} & {Max training FLOPs} \\
    \midrule
    \multirow{6}{*}{8} & \multirow{6}{*}{384} & \multirow{6}{*}{8} & 64 & 2 & 1 & 384 & \multirow{6}{*}{98} & \multirow{6}{*}{1.52e-3} & \multirow{6}{*}{2e18} \\
      &     &         & 128 & 4 & 2 & 192 &    &         &      \\
      &     &         & 192 & 6 & 3 & 128 &    &         &      \\
      &     &         & 256 & 8 & 4 & 96  &    &         &      \\
      &     &         & 384 & 12 & 6 & 64  &    &         &      \\
      &     &         & 512 & 16 & 8 & 48  &    &         &      \\
    \midrule
    \multirow{6}{*}{8} & \multirow{6}{*}{512} & \multirow{6}{*}{8} & 64 & 2 & 1 & 512 & \multirow{6}{*}{147} & \multirow{6}{*}{1.31e-3} & \multirow{6}{*}{6e18} \\
      &     &         & 128 & 4 & 2 & 256 &     &         &      \\
      &     &         & 192 & 6 & 3 & 170 &     &         &      \\
      &     &         & 256 & 8 & 4 & 128 &     &         &      \\
      &     &         & 384 & 12 & 6 & 85  &     &         &      \\
      &     &         & 512 & 16 & 8 & 64  &     &         &      \\
    \midrule
    \multirow{6}{*}{10} & \multirow{6}{*}{640} & \multirow{6}{*}{10} & 64 & 2 & 1 & 640 & \multirow{6}{*}{228} & \multirow{6}{*}{1.11e-3} & \multirow{6}{*}{2e19} \\
       &     &          & 128 & 4 & 2 & 320 &     &         &      \\
       &     &          & 192 & 6 & 3 & 213 &     &         &      \\
       &     &          & 256 & 8 & 4 & 160 &     &         &      \\
       &     &          & 384 & 12 & 6 & 106 &     &         &      \\
       &     &          & 512 & 16 & 8 & 80  &     &         &      \\
    \midrule
    \multirow{6}{*}{14} & \multirow{6}{*}{768} & \multirow{6}{*}{12} & 64 & 2 & 1 & 768 & \multirow{6}{*}{342} & \multirow{6}{*}{9.5e-4} & \multirow{6}{*}{6e19} \\
       &     &          & 128 & 4 & 2 & 384 &     &         &      \\
       &     &          & 192 & 6 & 3 & 256 &     &         &      \\
       &     &          & 256 & 8 & 4 & 192 &     &         &      \\
       &     &          & 384 & 12 & 6 & 128 &     &         &      \\
       &     &          & 512 & 16 & 8 & 96  &     &         &      \\
    \midrule
    \multirow{6}{*}{16} & \multirow{6}{*}{1024} & \multirow{6}{*}{16} & 64 & 2 & 1 & 1024 & \multirow{6}{*}{531} & \multirow{6}{*}{8.1e-4} & \multirow{6}{*}{2e20} \\
       &      &          & 128 & 4 & 2 & 512  &     &         &      \\
       &      &          & 192 & 6 & 3 & 341  &     &         &      \\
       &      &          & 256 & 8 & 4 & 256  &     &         &      \\
       &      &          & 384 & 12 & 6 & 170  &     &         &      \\
       &      &          & 512 & 16 & 8 & 128  &     &         &      \\
    \midrule
    \multirow{6}{*}{22} & \multirow{6}{*}{1280} & \multirow{6}{*}{20} & 64 & 2 & 1 & 1280 & \multirow{6}{*}{795} & \multirow{6}{*}{7.0e-4} & \multirow{6}{*}{6e20} \\
       &      &          & 128 & 4 & 2 & 640  &     &         &      \\
       &      &          & 192 & 6 & 3 & 426  &     &         &      \\
       &      &          & 256 & 8 & 4 & 320  &     &         &      \\
       &      &          & 384 & 12 & 6 & 213  &     &         &      \\
       &      &          & 512 & 16 & 8 & 160  &     &         &      \\
    \bottomrule
  \end{tabular}
\end{table}

\begin{table}[h!]
  \centering
  \caption{Experimental configurations for the shared expert ratio analysis. Within each group, we fix the total number of experts ($E=256$) and the total number of activated pathways ($E_a+E_s=12$), while varying the ratio between specialized experts ($E_a$) and shared experts ($E_s$) to study its impact on performance.}
  \label{table:shared_expert_configs}
  \small %
  \setlength{\tabcolsep}{4pt} %
  \begin{tabular}{cccccccccc}
    \toprule
    $n{layers}$ & $d_{model}$ & {$n_{\text{heads}}$} & {$E$} & {$E_a$} & {$E_s$} & {$d_{\text{expert}}$} & {$B$} & {$\eta$} & {Max training FLOPs} \\
    \midrule
    \multirow{6}{*}{8} & \multirow{6}{*}{384} &  \multirow{6}{*}{8} & 256 & 2 & 10 & 96 & \multirow{6}{*}{98} & \multirow{6}{*}{1.52e-3} & \multirow{6}{*}{2e18} \\
      &     &         & 256 & 4 & 8 & 96 &    &         &      \\
      &     &         & 256 & 6 & 6 & 96 &    &         &      \\
      &     &         & 256 & 8 & 4 & 96  &    &         &      \\
      &     &         & 256 & 11 & 1 & 96  &    &         &      \\
      &     &         & 256 & 12 & 0 & 96  &    &         &      \\
    \midrule
    \multirow{6}{*}{8} & \multirow{6}{*}{512} & \multirow{6}{*}{8} & 256 & 2 & 10 & 128 & \multirow{6}{*}{147} & \multirow{6}{*}{1.31e-3} & \multirow{6}{*}{6e18} \\
      &     &         & 256 & 4 & 8 & 128 &     &         &      \\
      &     &         & 256 & 6 & 6 & 128 &     &         &      \\
      &     &         & 256 & 8 & 4 & 128 &     &         &      \\
      &     &         & 256 & 11 & 1 & 128  &     &         &      \\
      &     &         & 256 & 12 & 0 & 128  &     &         &      \\
    \midrule
    \multirow{6}{*}{10} & \multirow{6}{*}{640} & \multirow{6}{*}{10} & 256 & 2 & 10 & 160 & \multirow{6}{*}{228} & \multirow{6}{*}{1.11e-3} & \multirow{6}{*}{2e19} \\
       &     &          & 256 & 4 & 8 & 160 &     &         &      \\
       &     &          & 256 & 6 & 6 & 160 &     &         &      \\
       &     &          & 256 & 8 & 4 & 160 &     &         &      \\
       &     &          & 256 & 11 & 1 & 160 &     &         &      \\
       &     &          & 256 & 12 & 0 & 160  &     &         &      \\
    \midrule
    \multirow{6}{*}{14} & \multirow{6}{*}{768} & \multirow{6}{*}{12} & 256 & 2 & 10 & 192 & \multirow{6}{*}{342} & \multirow{6}{*}{9.5e-4} & \multirow{6}{*}{6e19} \\
       &     &          & 256 & 4 & 8 & 192 &     &         &      \\
       &     &          & 256 & 6 & 6 & 192 &     &         &      \\
       &     &          & 256 & 8 & 4 & 192 &     &         &      \\
       &     &          & 256 & 11 & 2 & 192 &     &         &      \\
       &     &          & 256 & 12 & 0 & 192  &     &         &      \\
    \midrule
    \multirow{6}{*}{16} & \multirow{6}{*}{1024} & \multirow{6}{*}{16} & 256 & 2 & 10 & 256 & \multirow{6}{*}{531} & \multirow{6}{*}{8.1e-4} & \multirow{6}{*}{2e20} \\
       &      &          & 256 & 4 & 8 & 256  &     &         &      \\
       &      &          & 256 & 6 & 6 & 256  &     &         &      \\
       &      &          & 256 & 8 & 4 & 256  &     &         &      \\
       &      &          & 256 & 11 & 1 & 256  &     &         &      \\
       &      &          & 256 & 12 & 0 & 256  &     &         &      \\
    \midrule
    \multirow{6}{*}{22} & \multirow{6}{*}{1280} & \multirow{6}{*}{20} & 256 & 2 & 10 & 320 & \multirow{6}{*}{795} & \multirow{6}{*}{7.0e-4} & \multirow{6}{*}{6e20} \\
       &      &          & 256 & 4 & 8 & 320  &     &         &      \\
       &      &          & 256 & 6 & 6 & 320  &     &         &      \\
       &      &          & 256 & 8 & 4 & 320  &     &         &      \\
       &      &          & 256 & 11 & 1 & 320  &     &         &      \\
       &      &          & 256 & 12 & 0 & 320  &     &         &      \\
    \bottomrule
  \end{tabular}
\end{table}

\begin{table}[h!]
  \centering
  \caption{Experimental configurations for the arrangement of MoE and dense layers analysis. Within each group, the total number of layers is fixed at 60, while the mix of dense layers ($n_{\text{dense\_layers}}$) and MoE layers ($n_{\text{moe\_layers}}$) is varied to study the impact of their ratio and placement on performance.}
  \label{table:layer_arrangement_configs}
  \small %
  \setlength{\tabcolsep}{4pt} %
  \begin{tabular}{ccccccccccccc}
    \toprule
    $n_{layers}$ & $n_{dense\_layers}$ & $n_{moe\_layers}$ & $d_{model}$ & $d_{ffn}$ & {$n_{\text{heads}}$} & {$E$} & {$E_a$} & {$E_s$} & {$d_{\text{expert}}$} & {$B$} & {$\eta$} & {Max training FLOPs} \\
    \midrule
    \multirow{4}{*}{60} & 0 & 60 & \multirow{4}{*}{384} & \multirow{4}{*}{1280} & \multirow{4}{*}{8} & \multirow{4}{*}{64} & \multirow{4}{*}{2} & \multirow{4}{*}{1} & \multirow{4}{*}{384} & \multirow{4}{*}{98} & \multirow{4}{*}{1.52e-3} & \multirow{4}{*}{2e18} \\
       &      1 & 59 &  &      &    &    &    &    &    &     &         &      \\
       &      2 & 58 &  &      &    &    &    &    &    &     &         &      \\
       &      3 & 57 &  &      &    &    &    &    &    &     &         &      \\
    \midrule
    \multirow{4}{*}{60} & 0 & 60 & \multirow{4}{*}{512} & \multirow{4}{*}{2048} & \multirow{4}{*}{8} & \multirow{4}{*}{64} & \multirow{4}{*}{2} & \multirow{4}{*}{1} & \multirow{4}{*}{512} & \multirow{4}{*}{147} & \multirow{4}{*}{1.31e-3} & \multirow{4}{*}{6e18} \\
       &      1 & 59 &  &      &    &    &    &    &    &     &         &      \\
       &      2 & 58 &  &      &    &    &    &    &    &     &         &      \\
       &      3 & 57 &  &      &    &    &    &    &    &     &         &      \\
    \midrule
    \multirow{4}{*}{60} & 0 & 60 & \multirow{4}{*}{640} & \multirow{4}{*}{2560} & \multirow{4}{*}{10} & \multirow{4}{*}{64} & \multirow{4}{*}{2} & \multirow{4}{*}{1} & \multirow{4}{*}{640} & \multirow{4}{*}{228} & \multirow{4}{*}{1.11e-3} & \multirow{4}{*}{2e19} \\
       &      1 & 59 &  &      &    &    &    &    &    &     &         &      \\
       &      2 & 58 &  &      &    &    &    &    &    &     &         &      \\
       &      3 & 57 &  &      &    &    &    &    &    &     &         &      \\
    \midrule
    \multirow{4}{*}{60} & 0 & 60 & \multirow{4}{*}{768} & \multirow{4}{*}{3072} & \multirow{4}{*}{12} & \multirow{4}{*}{64} & \multirow{4}{*}{2} & \multirow{4}{*}{1} & \multirow{4}{*}{768} & \multirow{4}{*}{342} & \multirow{4}{*}{9.5e-4} & \multirow{4}{*}{6e19} \\
       &      1 & 59 &  &      &    &    &    &    &    &     &         &      \\
       &      2 & 58 &  &      &    &    &    &    &    &     &         &      \\
       &      3 & 57 &  &      &    &    &    &    &    &     &         &      \\
    \midrule
    \multirow{4}{*}{60} & 0 & 60 & \multirow{4}{*}{1024} & \multirow{4}{*}{4096} & \multirow{4}{*}{16} & \multirow{4}{*}{64} & \multirow{4}{*}{2} & \multirow{4}{*}{1} & \multirow{4}{*}{1024} & \multirow{4}{*}{531} & \multirow{4}{*}{8.1e-4} & \multirow{4}{*}{2e20} \\
       &      1 & 59 &  &      &    &    &    &    &    &     &         &      \\
       &      2 & 58 &  &      &    &    &    &    &    &     &         &      \\
       &      3 & 57 &  &      &    &    &    &    &    &     &         &      \\
    \midrule
    \multirow{4}{*}{60} & 0 & 60 & \multirow{4}{*}{1280} & \multirow{4}{*}{5120} & \multirow{4}{*}{20} & \multirow{4}{*}{64} & \multirow{4}{*}{2} & \multirow{4}{*}{1} & \multirow{4}{*}{1280} & \multirow{4}{*}{795} & \multirow{4}{*}{7.0e-4} & \multirow{4}{*}{6e20} \\
       &      1 & 59 &  &      &    &    &    &    &    &     &         &      \\
       &      2 & 58 &  &      &    &    &    &    &    &     &         &      \\
       &      3 & 57 &  &      &    &    &    &    &    &     &         &      \\
    \bottomrule
  \end{tabular}
\end{table}

\begin{table}[h!]
  \centering
  \caption{Experimental configurations for analyzing the compute allocation between attention and FFNs. Within each group, the core MoE structure is held constant, while we systematically vary the model's hidden dimension ($d_{\text{model}}$) and the expert dimension ($d_{\text{expert}}$) to explore the optimal trade-off in compute allocation between the attention mechanism and the FFN experts.}
  \label{table:attention_ffn_configs}
  \small %
  \setlength{\tabcolsep}{4pt} %
  \begin{tabular}{ccccccccccc}
    \toprule
    $_{layers}$ & $d_{model}$ & $d_{expert}$ & $n_{heads}$ & $n_{kv\_head}$ & $E$ & $E_s$ & $E_a$ & $\eta$   & $B$    & Max training FLOPs \\
    \midrule
    8           & 352         & 450          & 8           & 2              & 64   & 1     & 2     & 1.52e-3 & 96     & 2e18               \\
8           & 368         & 380          & 8           & 2              & 64   & 1     & 2     & 1.52e-3 & 96     & 2e18               \\
8           & 384         & 320          & 8           & 2              & 64   & 1     & 2     & 1.52e-3 & 96     & 2e18               \\
8           & 400         & 260          & 8           & 2              & 64   & 1     & 2     & 1.52e-3 & 96     & 2e18               \\
8           & 416         & 208          & 8           & 2              & 64   & 1     & 2     & 1.52e-3 & 96     & 2e18               \\
8           & 480         & 626          & 8           & 2              & 64   & 1     & 2     & 1.31e-3 & 160    & 6e18               \\
8           & 512         & 512          & 8           & 2              & 64   & 1     & 2     & 1.31e-3 & 160    & 6e18               \\
8           & 544         & 410          & 8           & 2              & 64   & 1     & 2     & 1.31e-3 & 160    & 6e18               \\
8           & 560         & 364          & 8           & 2              & 64   & 1     & 2     & 1.31e-3 & 160    & 6e18               \\
8           & 576         & 320          & 8           & 2              & 64   & 1     & 2     & 1.31e-3 & 160    & 6e18               \\
10          & 600         & 766          & 10          & 2              & 64   & 1     & 2     & 1.11e-3 & 224    & 2e19               \\
10          & 640         & 640          & 10          & 2              & 64   & 1     & 2     & 1.11e-3 & 224    & 2e19               \\
10          & 680         & 528          & 10          & 2              & 64   & 1     & 2     & 1.11e-3 & 224    & 2e19               \\
10          & 700         & 476          & 10          & 2              & 64   & 1     & 2     & 1.11e-3 & 224    & 2e19               \\
10          & 740         & 380          & 10          & 2              & 64   & 1     & 2     & 1.11e-3 & 224    & 2e19               \\
14          & 696         & 988          & 12          & 4              & 64   & 1     & 2     & 9.5e-3 & 320    & 6e19               \\
14          & 768         & 768          & 12          & 4              & 64   & 1     & 2     & 9.5e-3 & 320    & 6e19               \\
14          & 816         & 642          & 12          & 4              & 64   & 1     & 2     & 9.5e-3 & 320    & 6e19               \\
14          & 840         & 584          & 12          & 4              & 64   & 1     & 2     & 9.5e-3 & 320    & 6e19               \\
14          & 888         & 474          & 12          & 4              & 64   & 1     & 2     & 9.5e-3 & 320    & 6e19               \\
16          & 896         & 1378         & 16          & 4              & 64   & 1     & 2     & 8.1e-3 & 512    & 2e20               \\
16          & 1024        & 1024         & 16          & 4              & 64   & 1     & 2     & 8.1e-3 & 512    & 2e20               \\
16          & 1088        & 876          & 16          & 4              & 64   & 1     & 2     & 8.1e-3 & 512    & 2e20               \\
16          & 1152        & 742          & 16          & 4              & 64   & 1     & 2     & 8.1e-3 & 512    & 2e20               \\
16          & 1184        & 680          & 16          & 4              & 64   & 1     & 2     & 8.1e-3 & 512    & 2e20               \\
22          & 1120        & 1686         & 20          & 4              & 64   & 1     & 2     & 7.0e-3 & 768    & 6e20               \\
22          & 1280        & 1280         & 20          & 4              & 64   & 1     & 2     & 7.0e-3 & 768    & 6e20               \\
22          & 1360        & 1110         & 20          & 4              & 64   & 1     & 2     & 7.0e-3 & 768    & 6e20               \\
22          & 1440        & 956          & 20          & 4              & 64   & 1     & 2     & 7.0e-3 & 768    & 6e20               \\
22          & 1520        & 816          & 20          & 4              & 64   & 1     & 2     & 7.0e-3 & 768    & 6e20              \\
    \bottomrule
  \end{tabular}
\end{table}